\documentclass[11pt]{article}

\usepackage[final]{acl}
\usepackage{times}
\usepackage{latexsym}
\usepackage[T1]{fontenc}
\usepackage[utf8]{inputenc}
\usepackage{microtype}
\usepackage{inconsolata}
\usepackage{url}
\usepackage{booktabs}
\usepackage{amsfonts}
\usepackage{amsmath}
\usepackage{amssymb}
\usepackage{nicefrac}
\usepackage{xcolor}
\usepackage{colortbl}
\usepackage{multirow}
\usepackage{enumitem}
\usepackage{graphicx}
\usepackage{algorithm}
\usepackage{algpseudocode}
\usepackage{amsthm}

\usepackage{cleveref}
\usepackage{placeins}
\usepackage{pgfplots}
\usepgfplotslibrary{groupplots}
\pgfplotsset{compat=1.18}

\usepackage{xurl}
\usepackage{microtype}
\usepackage{amsmath}
\newcommand{\entity}[1]{\nolinkurl{#1}}

\definecolor{ReferenceGray}{RGB}{127,127,127}
\definecolor{BaselineBlue}{RGB}{86,180,233}
\definecolor{DFATeal}{RGB}{0,158,115}
\definecolor{ClampNavy}{RGB}{0,57,93}
\definecolor{AccentAmber}{RGB}{230,159,0}

\newif\ifrevisioncomments
\revisioncommentstrue

\graphicspath{{figures/}}

\title{CLAMP: Constrained Decoding for Vision-Language Embodied Planning}

\author{Tianyi Ma \quad Parisa Kordjamshidi\\
Michigan State University\\
\texttt{matiany3@msu.edu, kordjams@msu.edu}}

\begin{document}
\maketitle
\renewcommand{\thefootnote}{}
\footnotetext{Code: \url{https://github.com/HLR/CLAMP}}

\begin{abstract}

Embodied planning increasingly relies on vision-language models (VLMs) to translate instructions and visual observations into executable action sequences. However, fluent plans are not always executable. A VLM may refer to objects that are not visually observed, select actions whose required affordances are unavailable, or violate syntax and action constraints. We introduce \textbf{\textsc{CLAMP}}, a multimodal constraint-grounding framework that turns scene evidence into decoding-time constraints for a frozen VLM planner. \textsc{CLAMP} uses the initial observation to restrict object references to those supported by the scene, while a provided symbolic action model specifies state transitions and goals. 
During decoding, hard masks eliminate invalid next-token candidates, while a Hidden Markov Model (HMM)-based world-state lookahead module reweights the probabilities of the remaining feasible candidates based on action preconditions and goal reachability.
This allows the planner to retain the VLM’s language prior while preventing visually unsupported, unsafe, or infeasible candidates from entering the plan. For unseen tasks and environments, \textsc{CLAMP} adapts the HMM at test time using label-free continuations sampled from the frozen VLM. 
Experiments on VLABench, SafeAgentBench, and TaPA show that scene-grounded constraints improve object grounding and safety, while most remaining failures stem from perception errors or misaligned constraint specifications.

\end{abstract}

\section{Introduction}
\label{sec:intro}

%% Task + problem
Embodied planning requires an agent to translate multimodal observations and natural-language instructions into executable action sequences.
Vision-language models (VLMs) provide a powerful interface for this task~\citep{wu2023tapa, liang2023codeaspolicies, ahn2022can, li2024embodied}, but their generated plans do not necessarily satisfy the constraints required for execution.
%
% Vision-language models (VLMs) are increasingly used as planners for embodied planning~\citep{wu2023tapa, liang2023codeaspolicies, ahn2022can, li2024embodied}. 
% Given a visual observation and a natural-language instruction, an embodied planner generates a sequence of actions for a downstream executor to carry out in the environment.
%
A useful plan must therefore do more than match the instruction: its actions must be well-formed, grounded in the observed scene, and feasible under the current world state. 

%% Existing solutions + limitation

Existing approaches often handle execution constraints through post-hoc verification or repair~\citep{li2024embodied,Zhou_2025_CVPR,zhu2024earbench,wang2024llm3,zhang2024etplanbench}, allowing invalid decisions to propagate before they are corrected.
Constrained decoding~\citep{zhang2024ctrlg,safedec2025,geng2024grammar} instead enforces constraints during generation, but existing methods primarily target language generation or specialized policy outputs.
Embodied planning presents a different challenge: constraints must be grounded in multimodal observations and evolve with the world state.

% Existing approaches primarily address such failures after generation.
% A completed plan can be parsed, verified, rejected, or repaired when it contains an invalid action, refers to an unavailable object, or violates an action precondition~\citep{li2024embodied,Zhou_2025_CVPR,zhu2024earbench,wang2024llm3,zhang2024etplanbench}.
% However, post-hoc correction acts only after the action sequence has already been generated.
% This is problematic for autoregressive planning because an invalid early decision can affect subsequent actions and compound over a long planning horizon. A more direct approach is therefore to enforce execution constraints while the plan is being generated.

% Constrained decoding provides a natural mechanism for controlling autoregressive generation.
% For natural-language generation, methods such as \textsc{Ctrl-G}~\citep{zhang2024ctrlg} combine hard constraints with learned guidance to restrict the model's next-token distribution.
% Such methods, however, largely consider constraints defined over text.
% Embodied planning introduces a different setting: whether an action is valid depends not only on the generated sequence, but also on multimodal observations and the evolving world state.
% For example, the set of valid object references depends on what is present in the scene, while the validity of an action depends on whether its preconditions hold after preceding actions.
% Thus, applying constrained decoding to embodied planning requires constraints that are multimodal, episode-specific, and state-dependent.

%% Method overview & %% Test-time adaptation

We propose \textsc{Clamp}, a multimodal constrained decoding framework for embodied planning based on a frozen VLM.
\textsc{Clamp} combines hard constraints that restrict invalid generations with world-state guidance that steers the planner among valid actions.
% We instantiate these components with observation-conditioned DFAs and HMM-based state tracking, and adapt the lightweight guidance model at test time under domain shift.
Specifically, \textsc{Clamp} uses observation-conditioned DFAs to enforce hard constraints during decoding, restricting generation to actions that are syntactically valid and grounded in the current observation.
It further employs HMM-based state tracking to capture the evolving world state and provide lookahead guidance for subsequent actions.
To handle domain shift, \textsc{Clamp} adapts the lightweight guidance model at test time while keeping the VLM frozen.
% Because the learned world-state guidance may become miscalibrated under domain shift, \textsc{Clamp} further incorporates lightweight test-time adaptation.
% It recalibrates only the guidance model using unlabeled target-domain generations, while keeping the VLM frozen.
% This allows the constraint guidance to adapt to the deployment environment without retraining the underlying planner.

%% Experiments + takeaway & %% Ablation
We evaluate \textsc{Clamp} on multimodal embodied-planning benchmarks covering task execution, visual grounding, and safety.
On VLABench~\citep{zhang2025vlabench}, observation-conditioned hard constraints improve Qwen3-VL-8B from $28.7$ to $34.1$, while world-state guidance further increases performance to $37.1$ and $38.7$ with text- and vision-conditioned guidance, respectively.
On SafeAgentBench~\citep{yin2024safeagentbench}, constrained planning reduces symbolic-rule violations from $0.41$ to $0.05$.
These results indicate that enforcing constraints during decoding improves both task-level planning and constraint satisfaction without updating or scaling the underlying VLM.
The additional computational overhead remains modest because constraint enforcement and adaptation operate on lightweight components around the frozen planner.
Ablations further confirm the complementary roles of hard constraint enforcement and world-state guidance.

%% Contributions
Our contributions are threefold:
\begin{itemize}[leftmargin=*, topsep=2pt, itemsep=2pt, parsep=0pt]

% \item \textbf{Constrained decoding for embodied planning.}
% We extend constrained decoding from language generation to multimodal, state-dependent embodied planning.

\item \textbf{A multimodal constraint-decoding framework.}
We introduce \textsc{Clamp}, which integrates observation-conditioned hard constraint enforcement, world-state guidance, and lightweight test-time adaptation for embodied planning.

\item \textbf{Constraint-preserving test-time adaptation.}
We introduce lightweight HMM adaptation for \textsc{Clamp}, using unlabeled frozen-VLM continuations to recalibrate its guidance on new tasks while preserving the constraints.

% \item \textbf{Test-time adaptation across tasks.}
% We adapt \textsc{Clamp} to new tasks using unlabeled model-generated sequences, recalibrating its HMM guidance while preserving the constraints.

% \item \textbf{Multimodal constraint grounding.}
% \textsc{Clamp} enforces constraints from both visual observations and symbolic world states while keeping the VLM frozen.

\item \textbf{Effective and efficient constraint enforcement.}
\textsc{Clamp} improves embodied planning, grounding, and safety with modest additional computation cost.

\end{itemize}
\section{Related Work}
\label{sec:related_work}

\begin{figure*}[t]
\centering
\includegraphics[width=\textwidth]{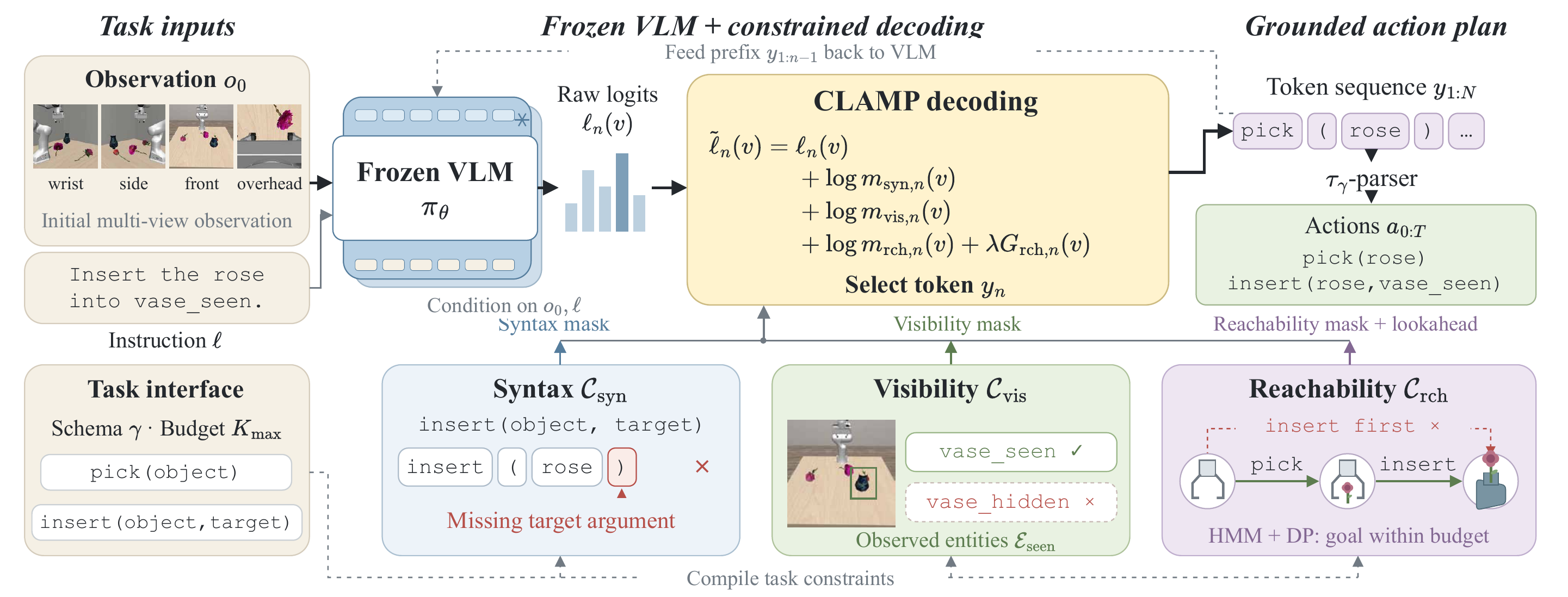}
\vspace{-8pt}
 \caption{%
Overview of \textsc{Clamp} on an illustrative rose-and-vase task. Syntax and visibility constraints enforce the action schema and observation-supported entity references.
A state-conditioned reachability constraint retains candidate tokens whose extended action prefixes admit executable action completions.
HMM-based lookahead further scores the remaining candidates using the probability mass of executable continuations that reach the goal within the remaining action budget.
The three masks and the weighted lookahead score modify frozen-VLM logits during autoregressive generation, and the resulting token sequence is parsed into an action plan.%
}
% \caption{%
% Overview of \textsc{Clamp}. Three conceptual constraints are compiled into two
% runtime detectors: the token-level
% $\mathcal{D}_\gamma'=\mathcal{D}_\gamma\cap\mathcal{D}_{\mathcal{E}_\mathrm{seen}}$
% enforces syntax and visibility, while the action-level $\mathcal{D}_\beta'$
% enforces visibility-restricted reachability. An HMM-weighted reachability score
% then reweights the frozen-VLM logits within the admissible set to decode
% grounded action sequences.%
% }
\label{fig:arch}
\end{figure*}

\paragraph{Embodied planning with language and vision-language models.} Language-conditioned planners have been used with symbolic planners, task-and-motion pipelines, and household simulators~\citep{garrett2021integrated, wang2024llm3,yang2024vlmtamp}. These systems differ in where they enforce constraints. Prompt-based planners rely on the model to follow a textual description, while monitor-and-repair systems inspect a plan after generation or after an execution failure~\citep{yang2024safetychip,yan2026vizcoast,wang2024llm3,ma-etal-2026-breaking}. \textsc{Clamp} instead applies a supplied interface while a high-level plan is decoded. It is not a replacement for motion planning, state estimation, or execution monitoring.

\paragraph{Constrained decoding.}
\label{sec:rel:cd}

Grammar and finite-state constrained decoding enforce requirements such as schemas, regular languages, and structured outputs at generation time~\citep{scholak2021picard,poesia2022synchromesh,willard2023outlines,geng2024grammar,ugare2024syncode,dong2024xgrammar,kamali2026neptune,kamali2026saturn}.
\textsc{Ctrl-G} combines hard constraints with HMM-based lookahead for language generation and provides the decoding backbone for \textsc{Clamp}~\citep{zhang2024ctrlg}.
\textsc{SafeDec} instead applies decode-time safety control to low-level robot-policy tokens under an assumed dynamics model~\citep{safedec2025}.
\textsc{Clamp} extends constrained decoding to high-level embodied planning, where constraints are grounded in multimodal observations and depend on the evolving world state.
%
% Grammar and finite-state constrained decoding enforce requirements such as schemas, regular languages, and structured outputs at generation time~\citep{scholak2021picard,poesia2022synchromesh,willard2023outlines, geng2024grammar,ugare2024syncode,dong2024xgrammar,kamali2026neptune,kamali2026saturn}. \textsc{Ctrl-G} adds HMM lookahead to a hard constraint and is the decoding backbone used by \textsc{Clamp}~\citep{zhang2024ctrlg}. \textsc{SafeDec} applies decode-time safety control to low-level robot-policy tokens under an assumed dynamics model~\citep{safedec2025}. Our setting differs in the source of the constraint: the initial observation instantiates an episode-specific entity support set that changes both token and action constraints.
% % \tm{TODO-13: replace the categorical expressivity claim with the finite
% % expansion and tractability boundary.}
In principle, a finite token automaton can encode bounded action history by state expansion. For episode-specific visibility and symbolic preconditions, however, this couples the token grammar to the current world state and yields a large product automaton; \textsc{Clamp} keeps the token and action-state detectors separate (Appendix~\ref{app:cd-survey}).

\paragraph{Visual grounding and safety.}
% \tm{TODO-11: state that visibility, instantiated from scene evidence, is
% the multimodal constraint.}
VLM planners can use observations to select subgoals and interact with task-and-motion components~\citep{wu2023tapa,yang2024vlmtamp,rajpal2026morelabelguidedsummarizationprocedural,ma-etal-2026-breaking, li2025industrynav}. Safety benchmarks show that task completion and constraint compliance can diverge in embodied settings~\citep{yin2024safeagentbench,lu2026isbench}. \textsc{Clamp} uses visibility as its multimodal constraint: scene evidence constructs a separate entity-support channel rather than treating image conditioning as a guarantee of grounded references.
% \tm{08/26-3: simplify the transition from grounding mechanism to evaluation.}
Accordingly, we report visual-grounding and constraint-enforcement probes
separately.

\section{Preliminaries}
\label{sec:prelim}

We consider embodied planning with a task specified by \((\ell,o_0,\gamma,\mathcal C,K_{\max})\),
where $\ell$ denotes the instruction; $o_0 \in \mathcal O$ denotes the initial multimodal observation, with $\mathcal O$ being the multimodal observation space; $\gamma$ denotes the task-provided action schema specifying action signatures, argument structures, and canonical entity information; 
% $\mathcal C$ denotes the constraints the decoded plan must satisfy; 
$\mathcal{C}$ denotes the planning constraints that the decoded plan must satisfy, with satisfaction determined based on the corresponding task information;
and $K_{\max}$ denotes the maximum action horizon. 
The goal is to generate an executable action plan satisfying $\mathcal C$ within the prescribed horizon.

Let $a_{0:T}=(a_0,\ldots,a_T)$ denote the final executable plan. The plan length is $T+1 \le K_{\max}$, and the plan satisfies the constraints denoted as $a_{0:T} \models \mathcal C$. Executing this plan starting from $o_0$ yields a final observation $o_{T+1}$, and task success is measured by a success score $\hat g(o_{T+1};\ell) \in [0,1]$.
% Note that $\hat g$ is not used as input, training reward, or feedback during generation.

In our framework, a frozen autoregressive VLM serves as the planning policy \(\pi_\theta\) and generates a token sequence \(y_{1:N}=(y_1,\ldots,y_N)\), which is subsequently converted into an executable action plan. The sequence contains action expressions and the formatting required by the output schema; reasoning text may also appear when the schema explicitly permits a separate reasoning field. Since a single action may span multiple tokens, the token length \(N\) need not equal the action horizon \(T+1\).
% 
% In our framework, a frozen autoregressive VLM serves as the planning policy $\pi_\theta$ that generates a token sequence $y_{1:N}$ sampled from a conditioned over-vocabulary distribution rather than directly generating actions:
% % TODO: explain the formula
% \[
% % y_n \sim \pi_\theta(y_n \mid \ell, o_0, y_{<n}), \quad n=1,\ldots,N.
% y_n \sim \pi_\theta\!\left(\cdot \mid \ell,o_0,\gamma,y_{<n} \right), \quad n=1,\ldots,N.
% % sample from a condition distribution of over-vocabulary
% \]
% Here, $y_{1:N}$ may contain tokens that form action expressions, as well as formatting and reasoning tokens. Since a single action may correspond to multiple tokens, the token length $N$ generally differs from the action horizon $T+1$ (\textit{i.e.}, $N \neq T+1$).

% The generated token sequence is converted into an executable plan via a deterministic parser $\tau_\gamma$ that uses action schema $\gamma$ to instantiate abstract action templates into executable actions:
% $\tau_\gamma(y_{1:N}) = a_{0:T}.$
% The action schema $\gamma$ defines action signatures, argument structures, and canonical entity information, using these to instantiate abstract action templates into executable actions. 
% TODO
The generated token sequence is converted into an executable plan via a deterministic parser \(\tau_\gamma\), with \(\tau_\gamma(y_{1:N})=a_{0:T}\), while \(w:\mathcal A_\gamma\rightarrow V^*\) denotes a token serialization of an executable action under $\gamma$.
%TODO
The parser identifies token spans corresponding to completed action expressions and parses them according to the action signatures, argument structures, and canonical entity information specified by $\gamma$. For example, a sequence of tokens
[\texttt{Pick},\texttt{(},\texttt{red},\texttt{\_cup},\texttt{\_2},\texttt{)}]
may jointly form the action expression \(\texttt{Pick(red\_cup\_2)}\), which \(\tau_\gamma\) maps to a single executable action \(a_t\). 
Conversely, under the same schema, \(w(\texttt{Pick(red\_cup\_2)})=[\texttt{Pick},\texttt{(},\texttt{red},\texttt{\_cup},\texttt{\_2},\texttt{)}]\) gives a token serialization of the executable action.
The parser distinguishes action names and entity arguments from punctuation, separators, and any schema-permitted reasoning fields; these auxiliary tokens do not themselves create actions. Let \(\mathcal A_\gamma\) denote the executable action space induced by \(\gamma\), so that \(a_t\in\mathcal A_\gamma\). Membership in this space means compatibility with the executor-facing action interface; state-dependent executability is specified separately.

% For example, $\texttt{Pick(object)}$ might be instantiated as $\texttt{Pick(red\_cup\_2)}$. Let $\mathcal A$ denote the abstract action space and $\mathcal A_\gamma$ denote the executable action space derived from $\gamma$; then $a_t \in \mathcal A_\gamma$.

% Throughout the decoding process as seen in Equation~\ref{TODO}, the observation conditioning for $\pi_\theta$ remains fixed at $o_0$. 
% The token generation process is only conditioned on the initial observation $o_0$. Intermediate observations $o_1,\ldots,o_T$ generated during execution are not fed back into subsequent token generation, nor does $\hat g(o_{T+1};\ell)$ provide feedback during generation.
Throughout decoding, the VLM is conditioned on \((\ell,o_0,\gamma,y_{<n})\), with its observation input fixed at \(o_0\). Intermediate execution observations \(o_1,\ldots,o_T\) and the final success score \(\hat g(o_{T+1};\ell)\) are not used as generation-time feedback.

\section{Method}
\label{sec:method}

Given the embodied planning setting, our goal is to generate an executable plan that satisfies the task constraints $\mathcal C$ within the prescribed action horizon $K_{\max}$, while keeping the VLM policy $\pi_\theta$ frozen. 
% Our goal is to enforce the task constraints $\mathcal C$ during generation while keeping the VLM policy $\pi_\theta$ frozen. 
To this end, \textsc{CLAMP} performs constrained decoding over the VLM-generated token sequence, expressing the planning constraints at the token level and enforcing them directly during autoregressive generation.
% achieves this by translating planning constraints into token-level signals that directly constrain the VLM's autoregressive predictions.

Let \(V\) denote the VLM token vocabulary. We use the next-token log-probability as the base decoding score for candidate \(v\in V\) at step \(n\):
\[
\ell_n(v)
=
\log \pi_\theta \left(v\mid \ell,o_0,\gamma,y_{<n}\right),
\quad v\in V.
\]
\textsc{CLAMP} incorporates the planning constraints by modifying these scores during decoding, as detailed below.
% These logits provide the common interface through which \textsc{CLAMP} incorporates the planning constraints introduced below.
% \textsc{CLAMP} operates on these logits at inference time, using constraint-derived signals to determine which token continuations remain valid and how they should be scored.

% The goal of \textsc{CLAMP} is to have the autoregressive VLM backbone respect to the constraints $\mathcal C$ without altering the parameters of $\pi_\theta$.

%TODO: explain the whole method of CLAMP here

% \textsc{CLAMP} is not to alter the parameters of $\pi_\theta$ or introduce intermediate observations into it; rather, it transforms task-level structured information into generation-time constraints, restricting the admissible continuations of the current prefix at each decoding step.

\subsection{Embodied Planning Constraints}
\label{sec:constraints}

In the embodied planning setting, we define the overall constraint set as \(\mathcal C=\{\mathcal C_{\mathrm{syn}},\mathcal C_{\mathrm{vis}},\mathcal C_{\mathrm{rch}}\}\),
% where \(\mathcal C_{\mathrm{syn}}\) enforces that each generated action conforms to the task-provided action schema, \(\mathcal C_{\mathrm{vis}}\) requires every referenced entity to be grounded in the initial visual observation \(o_0\), and \(\mathcal C_{\mathrm{rch}}\) requires the generated action sequence to reach the symbolic goal within the given action budget.
%
where $\mathcal C_{\mathrm{syn}}$ denotes the syntax constraint, requiring each generated action to conform to the task-provided action schema; $\mathcal C_{\mathrm{vis}}$ denotes the visibility constraint, requiring every entity referenced by the generated actions to be grounded in the visual observation $o_0$; and $\mathcal C_{\mathrm{rch}}$ denotes the reachability constraint, requiring the generated action sequence 
{to be consistent with the evolving task state } and to reach a goal within the given action budget.

\paragraph{Syntax Constraint.}

The syntax constraint \(\mathcal C_{\mathrm{syn}}\) requires the generated action sequence to conform to the task-provided action schema \(\gamma\). Formally,
\[
a_{0:T}\models\mathcal C_{\mathrm{syn}}
% \quad
\Longleftrightarrow
% \quad
a_t\in\mathcal A_\gamma,
% \qquad
\forall t\in \{0,\ldots,T\},
\]
with \(a_{0:T}\) serialized according to the format specified by \(\gamma\), where \(\mathcal A_\gamma\) denotes the executable action space induced by \(\gamma\).
% Here, executability refers to compatibility with the executor-facing action interface; state-dependent executability is handled separately by the reachability constraint.

% The syntax constraint \(\mathcal C_{\mathrm{syn}}\) determines whether the generated action sequence conforms to the task-provided action schema \(\gamma\). 
% Specifically, \(\gamma\) specifies the admissible action signatures, argument structures, canonical entity identifiers, and the required output format, thereby defining the executable action space \(\mathcal A_\gamma\). 
% % Here, \emph{executable} refers to compatibility with the executor-facing action interface, rather than state-dependent executability, which is handled separately by the reachability constraint.

% Accordingly, \(\mathcal C_{\mathrm{syn}}\) requires every generated action to belong to \(\mathcal A_\gamma\) and the resulting action sequence to follow the prescribed serialization format:
% $$
% a_{0:T}\models\mathcal C_{\mathrm{syn}}
% % \quad
% \Longleftrightarrow
% % \quad
% a_t\in\mathcal A_\gamma,
% % \quad
% \forall t\in\{0,\ldots,T\},
% $$
% with \(a_{0:T}\) serialized according to the format specified by \(\gamma\). Thus, \(\mathcal C_{\mathrm{syn}}\) rules out undefined actions, malformed argument structures, and invalid output formats.

\paragraph{Visibility Constraint.}

The visibility constraint \(\mathcal C_{\mathrm{vis}}\) requires every entity referenced by the generated actions to be grounded in the initial visual observation \(o_0\). 
Let \(\mathcal E_{\mathrm{seen}}\) denote the set of observation-supported entity identifiers extracted from \(o_0\) and aligned with the canonical identifiers in \(\gamma\). For an action \(a_t\), let \(\operatorname{args}(a_t)\) denote the entities referenced by the action. Then,
\[
a_{0:T}\models\mathcal C_{\mathrm{vis}}
% \quad
\!\Longleftrightarrow\!
% \quad
\operatorname{args}(a_t)\!\subseteq\!\mathcal E_{\mathrm{seen}},
% \qquad
\forall t\!\in\{0,\!\ldots,\!T\}.
\]
% The visibility constraint prevents the planner from referring to entities that are supported by the schema but not grounded in the initial visual observation.

\paragraph{Reachability Constraint.}

The syntax and visibility constraints enforce local validity during action construction, but they do not guarantee that the resulting action sequence is executable under the evolving task state or that it can reach the goal within the given action budget.
{An action can satisfy both \(\mathcal C_{\mathrm{syn}}\) and \(\mathcal C_{\mathrm{vis}}\) while remaining infeasible in the current state because its execution conditions are not satisfied. Moreover, even when every selected action is executable at the time of execution, the resulting action sequence may require too many steps to reach the task goal within the given action budget.}
% On the one hand, an action can satisfy both \(\mathcal C_{\mathrm{syn}}\) and \(\mathcal C_{\mathrm{vis}}\) while remaining infeasible in the current state because its execution conditions are not satisfied.
% On the other hand, even when every selected action is executable at the time of execution, the resulting action sequence may require too many steps to reach the task goal within the given action budget.
%
% Therefore, \(\mathcal C_{\mathrm{rch}}\) requires the generated actions to induce a valid trajectory that reaches the goal without exceeding the allowed number of actions.

Accordingly, the reachability constraint is defined as
% Accordingly,
\[
a_{0:T}\models\mathcal C_{\mathrm{rch}}
% \quad
\Longleftrightarrow
% \quad
\operatorname{Reachable}(a_{0:T};K_{\max}),
\]
where \(\operatorname{Reachable}(a_{0:T}; K_{\max})\) holds if the sequence is executable under the evolving task state, 
terminates in a task-specified goal state
% reaches the task goal, 
and satisfies \(T+1\le K_{\max}\).
Appendix~\ref{app:constraint-nonsubsumption} gives visibility and executability examples
% counterexamples 
illustrating why the three constraints are complementary.

\subsection{Constrained Decoding}
\label{sec:constrained_decoding}
\label{sec:method:processor}

% add a roadmap for the CLAMP
\textsc{CLAMP} enforces these planning constraints during decoding by evaluating each candidate token against the corresponding constraints and modifying its next-token logit accordingly. 
% \tm{
For each constraint \(i\in\{\mathrm{syn},\mathrm{vis},\mathrm{rch}\}\), 
% \(m_{i,n}(v)\in\{0,1\}\) is a binary validity mask indicating whether candidate token v is admissible at decoding step n. 
let \(m_{i,n}(v)\in\{0,1\}\) indicate whether candidate token $v$ is admissible under constraint $i$ at decoding step $n$.
We use $\log m_{i,n}(v)=0$ for an admissible token and $-\infty$ otherwise, so any invalid candidate receives zero probability after softmax.
% }
Specifically, at decoding step $n$,
\begin{equation}
\label{eq:constrained_logit}
\begin{aligned}
\tilde{\ell}_n(v)
={}&
\ell_n(v)
+\underbrace{\log m_{\mathrm{syn},n}(v)}
_{\mathcal C_{\mathrm{syn}}}
+\underbrace{\log m_{\mathrm{vis},n}(v)}
_{\mathcal C_{\mathrm{vis}}}
\\
&+
\underbrace{
\log m_{\mathrm{rch},n}(v)
+\lambda G_{\mathrm{rch},n}(v)
}_{\mathcal C_{\mathrm{rch}}}.
\end{aligned}
\end{equation}
% \[
% \begin{aligned}
% \tilde{\ell}_n(v)
% ={}&
% \ell_n(v)
% +\underbrace{\log m_{\mathrm{syn},n}(v)}
% _{\mathcal C_{\mathrm{syn}}}
% +\underbrace{\log m_{\mathrm{vis},n}(v)}
% _{\mathcal C_{\mathrm{vis}}}
% \\
% &+
% \underbrace{
% \log m_{\mathrm{rch},n}(v)
% +\lambda G_{\mathrm{rch},n}(v)
% }_{\mathcal C_{\mathrm{rch}}}.
% \end{aligned}
% \]
% Here, \(m_{\mathrm{syn},n}(v)\), \(m_{\mathrm{vis},n}(v)\), and \(m_{\mathrm{rch},n}(v)\) are binary token-level validity masks induced by the corresponding constraints. 
The score \(G_{\mathrm{rch},n}(v)\) aggregates hidden Markov model (HMM)-weighted goal-reaching continuations under the symbolic transition model and the remaining action budget. Its contribution is controlled by \(\lambda\ge0\); when \(\lambda=0\), the entire future-reachability term is omitted.
%

% All three constraints are enforced through token-level DFAs, with action serialization connecting reachability filtering to symbolic state transitions.
% Syntax and visibility are enforced by token-level deterministic finite automata (DFAs)~\citep{hopcroft2001introduction}. Reachability uses an action-level DFA: the decoder first identifies executable actions from the current symbolic task state, then uses their token serializations to constrain generation. Both levels determine token masks in Eq.~\eqref{eq:constrained_logit}.
%
% To enforce the three constraints during decoding, \textsc{CLAMP} maps each constraint to a token-level validity test.
% Syntax and visibility are represented directly by token-level deterministic finite automata (DFA)~\citep{hopcroft2001introduction}, whereas reachability is defined over complete actions and projected onto candidate tokens through their possible action completions.
%
% TODO: do we need the notation I?
% Let $\mathcal I=\{\mathrm{syn},\mathrm{vis},\mathrm{rch}\}$ index the syntax, visibility, and reachability constraints.
% and let $\mathcal I_{\mathrm{local}}=\{\mathrm{syn},\mathrm{vis}\}$. 
% For each $i\in\mathcal I_{\mathrm{local}}$, 
For each \(i\in\{\mathrm{syn},\mathrm{vis},\mathrm{rch}\}\), let \(\mathcal D_i=(\mathcal Q_i,\Sigma_i,\delta_i,q_{i,0},F_i)\) 
% \tm{
denote an automaton representing the corresponding constraint.
% }
% denote its constraint automaton. 
%
The syntax and visibility automata operate directly on tokens, with \(\Sigma_{\mathrm{syn}}=\Sigma_{\mathrm{vis}}=V\), whereas the reachability automaton reads completed actions, with \(\Sigma_{\mathrm{rch}}=\mathcal A_\gamma\). 
%
% \tm{
Despite operating at different levels, all three constraints are enforced through the token-level masks in Eq.~\ref{eq:constrained_logit}. 
Syntax and visibility determine admissible tokens directly from their token-level automata. Reachability instead maps candidate tokens to compatible action completions and checks those actions against the action-level automaton, as defined below.
\subsubsection{Prefix-local Filtering}
\label{sec:prefix_local_filtering}
% \pk{Just call it "Local Constraints?".}\tm{Not accept. I prefer to keep “Prefix-local Filtering” because “prefix-local” precisely indicates that syntax and visibility are evaluated from the current token prefix, whereas “Local Constraints” is ambiguous about what “local” refers to.}
The syntax and visibility constraints can be evaluated directly from the current token prefix.
% without reasoning over future world states.
%
Before decoding, we instantiate \(\mathcal D_{\mathrm{syn}}\) from the action schema \(\gamma\) and \(\mathcal D_{\mathrm{vis}}\) from \(\mathcal E_{\mathrm{seen}}\). The former recognizes valid serialized actions and their surrounding format, including the required punctuation and separators, while the latter restricts entity-argument positions to observation-supported identifiers. 
For either automaton, \(q_{i,n}\) denotes the state after processing \(y_{<n}\), with \(q_{i,1}=q_{i,0}\). 
% A candidate token $v$ is admissible if its transition preserves at least one completion accepted by the automaton. This gives the binary mask
A candidate token \(v\) is admissible if taking its transition from \(q_{i,n}\) preserves at least one accepting completion. This gives the binary mask
\[
m_{i,n}(v)=
\begin{cases}
1, & \text{if }\delta_i(q_{i,n},v)\text{ can reach }F_i,\\
0, & \text{otherwise}.
\end{cases}
\]

\subsubsection{State-conditioned Executability Filtering}
\label{sec:state_conditioned_reachability}

% Unlike syntax and visibility, reachability constraint is defined over complete actions rather than individual tokens. 
% We represent reachability constraint $\mathcal C_{rch}$ by the following DFA: 
State-conditioned filtering retains candidate tokens that can still be
completed into an action executable from the current state.
We describe the symbolic action transitions underlying this token-level
filter by
\[
\mathcal D_{\mathrm{rch}}
=
(\mathcal Z,\mathcal A_\gamma,\Delta,z_0,F_{\mathrm{rch}}),
\]
where $\mathcal Z$ is the state space, $\mathcal A_\gamma$ is the schema-defined action space, $\Delta$ is the action-level transition function, $z_0$ is the initial state, and $F_{\mathrm{rch}}\subseteq\mathcal Z$ is the set of goal states. 
The task specification provides \(z_0\) and \(F_{\mathrm{rch}}\), while action preconditions and effects determine \(\Delta\). 
% Transitions are undefined when an action's preconditions are not satisfied.
This automaton is fixed before decoding, with \(\Delta(z,a)\) undefined
when the preconditions of \(a\) are not satisfied in \(z\).
For any state \(z\in\mathcal Z\), define the executable-action set as
\(
\mathcal A_{\mathrm{exe}}(z)
=
\{a\in\mathcal A_\gamma:\Delta(z,a)\text{ is defined}\}.
\)
Let \(t\) denote the number of actions fully generated and parsed so far, and let \(z_t\in\mathcal Z\) denote the state obtained by applying these actions, \(a_0,\ldots, a_{t-1}\), under \(\Delta\).
% Let \(z_t\) denote the state after the first \(t\) completed actions \(a_0,\ldots,a_{t-1}\). 
% The state remains \(z_t\) while the current action is being generated and advances to \(z_{t+1}=\Delta(z_t,a_t)\) only when the parser identifies a completed action \(a_t\).
%
% During decoding, this fixed mapping is queried at the current state $z_t$. Maintaining $z_t$ carries forward the modeled effects of previously completed actions. It remains unchanged while the next action is generated token by token, and advances to $\Delta(z_t,a_t)$ only when the parser $\tau_\gamma$ identifies a completed action $a_t$. The applicable set $\mathcal A_{\mathrm{exe}}(z_t)$ therefore changes with the current state, while the automaton's transition rules remain fixed.
% During decoding, candidate action completions are checked against \(\mathcal A_{\mathrm{exe}}(z_t)\).
During decoding, \(\mathcal A_{\mathrm{exe}}(z_t)\) is queried at the current state \(z_t\), which remains unchanged while the next action is generated token by token. 
Once the parser \(\tau_\gamma\) identifies a completed action \(a_t\), the state advances to \(z_{t+1}=\Delta(z_t,a_t)\). 
% % The state advances to \(z_{t+1}=\Delta(z_t,a_t)\) only when the parser identifies a completed action \(a_t\).
% , otherwise it remains \(z_t\).
%
% The reachability constraint $\mathcal C_{rch}$ must nevertheless act during token generation, before the current action is complete. 
% % We use the \(w(a)\) to determine which complete actions are compatible with the current token prefix.
% Let \(w:\mathcal A_\gamma\rightarrow V^*\) map each action to its token serialization; 

% repeat? 
% This construction allows action-level semantics to constrain token-level generation without assigning task semantics to individual tokens. A candidate token is admissible under reachability precisely when it preserves at least one executable complete-action continuation.

% For any state $z\in\mathcal Z$, define its executable-action set as
% \(
% \mathcal A_{\mathrm{exe}}(z)
% =
% {a\in\mathcal A_\gamma:\Delta(z,a)\text{ is defined}}.
% \)

% Thus, the executable-action set changes with the evolving symbolic state, while the transition model \(\Delta\) remains fixed.

To apply these action-level transitions during token decoding, 
% we use the serialization map $w(a)$ that realizes a complete action $a$ as a token sequence under $\gamma$. 
the serialization map \(w:\mathcal A_\gamma\to V^*\) associates each action with its token sequence under \(\gamma\).
An unfinished action prefix may match several such sequences. 
% We retain a candidate token if at least one matching sequence represents an action executable from $z_t$.
%
% \tm{Not accpet: \(\tau_\gamma\) defined in Preliminary section.}
Formally, let \(b_t\) denote the token index immediately before the current action span begins.
% , as identified by \(\tau_\gamma\).
% any preceding wrapper or reasoning tokens are outside this span. 
% \tm{
At decoding step $n$, the previously generated tokens of the current action form the prefix \(p_{t,n}=y_{b_t+1:n-1}\).
% }
% For a canonical action serialization, the unfinished action prefix is \(p_{t,n}=y_{b_t+1:n-1}\). 
%
% We use the \(w(a)\) to determine which complete actions are compatible with the current token prefix. 
Let \(\circ\) denote token-sequence concatenation and
\(x\preceq u\) mean that \(x\) is a prefix of \(u\).
For a candidate token $v$ inside an action span, let $\mathcal A_n(v)$ denote the complete actions whose token serializations are compatible with the extended prefix $p_{t,n}\circ v$:
% For a candidate token \(v\) inside an action span, the set of compatible complete actions is defined by the parser state after tentatively consuming \(v\). Under the action serialization \(w\), this reduces to
\(
\mathcal A_n(v)
=
\left\{
a\in\mathcal A_\gamma:
p_{t,n}\circ v
\preceq
w(a)
\right\}.
\)
% where \(\circ\) denotes token-sequence concatenation, and \(x\preceq u\) denotes token sequence \(x\) is a prefix of token sequence \(u\).
Intersecting this set with the actions executable from \(z_t\) gives 
\(
\mathcal A_{\mathrm{exe},n}(v;z_t)
=
\mathcal A_n(v)\cap\mathcal A_{\mathrm{exe}}(z_t).
\)
The corresponding token mask is
\[
m_{\mathrm{rch},n}(v)=
\begin{cases}
1, & \text{if }\mathcal A_{\mathrm{exe},n}(v;z_t)\neq\varnothing,\\
0, & \text{otherwise}.
\end{cases}
\]

Immediate executability does not guarantee that an action's successor
state admits a goal-reaching continuation within the remaining action
budget. Section~\ref{sec:budget_aware_reachability} augments this
executability mask with budget-aware future reachability.
Appendix~\ref{app:beta-dp} illustrates token filtering and action-boundary
state updates, while Appendix~\ref{app:decode-pipeline} describes parser
handling of alternative spacing and punctuation.

\subsubsection{Budget-aware Future Reachability}
\label{sec:budget_aware_reachability}

To score future goal reachability, we combine the action-level DFA with a discrete HMM. For each candidate token, the lookahead evaluates its compatible executable action completions and the continuations from their successor states within the remaining action budget.

% The state-conditioned filter above removes token extensions that cannot be completed into an action executable under \(\mathcal D_{\mathrm{rch}}\). To distinguish the remaining candidates, we combine this action automaton with a discrete hidden Markov model (HMM) and evaluate whether their possible action completions can reach a goal within the remaining action budget. This lookahead uses the symbolic transition model; 
% visibility remains a separate token-level constraint and is not included in the backward recursion.

\paragraph{Action-level backward reachability.}
% We model token-generation dynamics with a discrete HMM $\phi=(\mu,A,B)$ over latent states $\mathcal H$.
% Given the action serialization $w(a)$, let
The HMM has parameters \(\phi=(\mu,A,B)\), where \(\mu\) is the initial distribution over latent states \(\mathcal H\), \(A\) is the transition matrix, and \(B\) is the emission matrix. 
% \tm{
For an action \(a\) with token serialization \(w(a)\), let
% }
% For an action \(a\) with canonical serialization \(w(a)\), let
\[
F_a[h,h']
:=
\log P_\phi\!\left(w(a),h_{\lvert w(a)\rvert}=h'\mid h_0=h\right)
\]
% denote its HMM log-probability from latent state $h$ to $h'$, marginalizing over intermediate states.
% \tm{
denote the HMM log-probability of generating \(w(a)\) while transitioning from latent state \(h\) to \(h'\).
% }
% be its HMM segment log-probability from latent state \(h\) to \(h'\), with local latent-state indices and intermediate states marginalized out. 
We maintain the joint forward log-probability \(\rho_n[h]=\log P_\phi(y_{1:n},h_n=h)\) over the generated prefix. Lookahead scores canonical action continuations; it does not marginalize over arbitrary future reasoning text or formatting variants. Details of both computations are given in Appendix~\ref{app:hmm-bridge}.

Let $K$ denote the number of complete actions that may still be generated. For reachability state $z\in\mathcal Z$, HMM state $h\in\mathcal H$, and remaining action budget $K$, define $R[K,z,h]$ as the HMM-weighted log-probability mass of executable action continuations that reach a goal state in $F_{\mathrm{rch}}$ within at most $K$ additional actions.

The boundary conditions are
$R[K,z,h]=0$ for $z\in F_{\mathrm{rch}}$
and
$R[0,z,h]=-\infty$ for $z\notin F_{\mathrm{rch}}$.
Reaching a goal ends the scored continuation, so the first boundary condition assigns unit mass to successful termination.
Let \(\operatorname{LSE}\) denote log-sum-exp, which marginalizes over executable actions and their possible successor HMM states.
For $K>0$ and $z\notin F_{\mathrm{rch}}$, the backward recursion is
% \tm{Accept: $R[K,z,h], K, F_a,$ and $\mathcal A_{\mathrm{rch}}(z) $ are already defined before the recursion. I moved the definition of $\operatorname{LSE}$, which was the only new notation introduced after the formula, to before the equation.}
%
% \[
% R[K,z,h]
% =
% \operatorname*{LSE}_{
% \substack{
% a\in\mathcal A_{\mathrm{rch}}(z)\\[-2pt]
% h'\in\mathcal H
% }}
% \bigl[
% F_a[h,h']
% +
% R(K{-}1,\Delta(z,a),h')
% \bigr].
% \]
%
\[
\begin{aligned}
R[K,z,h]
&=
\operatorname{LSE}_{
\substack{
a\in\mathcal A_{\mathrm{exe}}(z)\\
h'\in\mathcal H
}}
\bigl[
F_a[h,h']
\\[-2pt]
&\quad+
R\left[
K-1,
\Delta(z,a),
h'
\right]
\bigl].
\end{aligned}
\]
Each term evaluates one executable next action by combining its HMM segment score with the future reachability of its successor state.

\paragraph{Candidate-token reachability score.}

We finally project the action-level backward reachability computation onto candidate tokens. For candidate token $v$, the set $\mathcal A_{\mathrm{exe},n}(v;z_t)$ contains the complete actions that are both compatible with the extended token prefix and executable from the current reachability state $z_t$.

% We finally project the action-level future-reachability computation back onto candidate tokens. For candidate token $v$, only actions in $\mathcal A_{\mathrm{rch},n}(v;z_t)$ are simultaneously compatible with the extended token prefix and executable from the current reachability state.

Let $\rho_{b_t}$ denote the HMM forward log-belief at the beginning of the current action. 
The budget-aware reachability score for candidate token $v$ is
% We define
\[
\begin{aligned}
&G_{\mathrm{rch},n}(v)
=\operatorname{LSE}_{\substack{a\in\mathcal A_{\mathrm{exe},n}(v;z_t)\\h,h'\in\mathcal H}}
\Bigl[\rho_{b_t}[h]\\
&\quad+F_a[h,h']+R[K_{\max}-t-1,\Delta(z_t,a),h']\Bigr]\\
&\quad-\operatorname{LSE}_{h\in\mathcal H}\rho_{b_t}[h].
\end{aligned}
\]

If \(\mathcal A_{\mathrm{exe},n}(v;z_t)=\varnothing\),
we set $G_{\mathrm{rch},n}(v)=-\infty$.
Otherwise, \(G_{\mathrm{rch},n}(v)\) aggregates the future-reachability mass of the compatible executable actions. The final term normalizes the HMM belief at the action start, removing the likelihood of the already generated prefix. Since \(t\) actions have already been committed, completing the current action leaves \(K_{\max}-t-1\) actions for the continuation. The score is evaluated only when \(t<K_{\max}\) and the observed prefix has positive HMM probability.

The two components of \(\mathcal C_{\mathrm{rch}}\) in Eq.~\eqref{eq:constrained_logit} serve complementary roles. The hard mask \(m_{\mathrm{rch},n}(v)\) removes token extensions with no executable action completion. For \(\lambda>0\), a score of \(G_{\mathrm{rch},n}(v)=-\infty\) also excludes candidates with no HMM-supported goal-reaching continuation within budget; finite scores rank the remaining candidates. Thus, a nonempty action-completion set alone does not imply a finite future-reachability score.

\begin{table*}[!t]
\centering
\small
\setlength{\tabcolsep}{3.5pt}
\begin{tabular*}{\textwidth}{@{\extracolsep{\fill}}lccccccc@{}}
\toprule
\textbf{Method} & M\&T$\uparrow$ & Spatial$\uparrow$ & ComSense$\uparrow$ & Semantic$\uparrow$ & PhysLaw$\uparrow$ & Complex$\uparrow$ & \textbf{Macro}$\uparrow$ \\
\midrule
\multicolumn{8}{l}{\emph{Reference VLM baselines from~\citep{zhang2025vlabench} (zero-/one-shot, no constraints)}} \\
GLM-4V-9B (0-shot)                              & 25.0 & 24.1 & 18.7 & 22.9 & 41.8          &  10.8 & 23.9 \\
GPT-4-turbo-2024-04-09                          & 33.4 & 29.5          & 22.2 & 28.8 &  7.9          &  19.0 & 23.5 \\
GPT-4o-2024-08-06                               & 26.7 & 24.1 & 16.2 & 25.1 & 29.1          &  30.8          & 25.3 \\
InternVL2-8B                                    & 10.7 &  8.4 &  8.8 & 14.2 & 25.2          &  11.8 & 13.2 \\
LLaVA-NeXT                                      & 16.1 & 11.7 & 14.7 & 21.8 &  5.9          &   4.4 & 12.4 \\
MiniCPM-V 2.6                                   & 22.2 & 18.6 & 19.2 & 22.5 & 13.7          &   8.9 & 17.5 \\
Qwen2-VL-7B-Instruct                            & 34.6          & 25.8 & 29.1          & 35.1          & 19.4          &  14.2 & 26.4 \\
\midrule
\multicolumn{8}{l}{\emph{VL-main: Qwen3-VL-8B-Instruct, matched inference-only configurations}} \\
Baseline                        & 38.5 & 28.0 & 30.4 & 36.8 & 22.6 & 16.1 & 28.7 \\
$+$ syntax--visibility DFA      & 46.7 & 33.6 & 36.2 & 41.0 & 27.4 & 19.8 & 34.1 \\
$+$ text HMM                    & \underline{51.4} & \underline{36.5} & \underline{39.1} & \underline{43.9} & \underline{29.9} & \underline{21.5} & \underline{37.1} \\
\midrule
\multicolumn{8}{l}{\emph{Qwen3-VL-8B-Instruct: additional supervision and calibration}} \\
$+$ supervised vision HMM       & \textbf{53.8} & \textbf{38.2} & \textbf{40.7} & \textbf{45.3} & \textbf{31.4} & \textbf{22.6} & \textbf{38.7} \\
\midrule
\multicolumn{8}{l}{\emph{\textsc{Clamp} on InternVL3.5-8B (matched four-dimension subset)}} \\
Baseline                        & 25.5 & 25.7 & 23.5 & 22.4 & -- & -- & -- \\
$+$ \textsc{Clamp}              & \textbf{61.7} & \textbf{44.3} & \textbf{54.6} & \textbf{46.2} & -- & -- & -- \\
\bottomrule
\end{tabular*}
\caption{
% \tm{
\textbf{VLABench planning results.}
M\&T, ComSense, and PhysLaw denote Mesh and Texture, Common Sense, and Physical Laws.
Scores range from $0$ to $100$; higher is better.
Macro is the unweighted mean of the six dimension scores.
Dashes indicate unevaluated dimensions.
Bold and underlined values indicate the best and second-best results, respectively.
% }
% Bold marks the best result within each backbone block; underlining marks the best inference-only result within the corresponding block.
}
\label{tab:vlabench}
\end{table*}
% \vspace{-3pt}

\subsubsection{Decoding and State Updates}
\label{sec:decoding_state_updates}

Action-completion masks and scores apply when constructing an action. For grammar-permitted tokens outside action spans, their reachability contributions are neutral, \(m_{\mathrm{rch},n}(v)=1\) and \(G_{\mathrm{rch},n}(v)=0\), subject to the termination and budget checks below. 
% Termination tokens are handled separately from action-completion candidates. 
A plan terminator is admissible only at an action boundary with \(z_t\in F_{\mathrm{rch}}\) and a syntactically complete serialization. Once \(t=K_{\max}\), no further action may be started. 
After applying the constraint masks and reachability score, decoding returns \textsc{Fail} 
if the action budget is exhausted with \(z_t\notin F_{\mathrm{rch}}\)
% if the action budget is exhausted outside the goal set 
or if all candidate tokens have \(\tilde{\ell}_n(v)=-\infty\), in which case generation terminates without a complete valid plan. Otherwise, the next token is sampled as
\(y_n\sim\operatorname{softmax}(\tilde{\ell}_n)\).
% Decoding returns \textsc{Fail} if the budget is exhausted outside the goal set or no admissible continuation has a finite modified logit. 
% If neither failure condition is met, decoding proceeds normally by sampling \(y_n\sim\operatorname{softmax}(\tilde\ell_n)\).
% Otherwise, we sample \(y_n\sim\operatorname{softmax}(\tilde\ell_n)\).

After selecting \(y_n\), each token-level DFA state is updated according to its transition function. The reachability state \(z_t\) remains unchanged while the current action is being generated. Once the parser identifies a completed action \(a_t\), the action is committed and the reachability state is updated as \(z_{t+1}=\Delta(z_t,a_t)\).

\subsubsection{Test-Time Adaptation}
\label{sec:tta}
\label{sec:method:tta}

To improve the generalization of \textsc{Clamp} to new tasks, we adapt its HMM at test time. Because the HMM is learned from source-task data, its token probabilities can become miscalibrated under task shift, weakening the guidance used for constrained decoding. We therefore use test-time adaptation (TTA) to recalibrate the HMM from unlabeled target-task sequences \citep{liang2024ttatsurvey}.

We adapt only the emission matrix \(B\), which directly maps latent states to observed tokens and is therefore most directly exposed to task-specific changes in the observed sequences. In contrast, \(\mu\) and \(A\) encode the source HMM's initial-state and transition dynamics. Restricting adaptation to \(B\) provides a lightweight fast-weight update \citep{feng2026inplace} while keeping the latent dynamics and symbolic constraints fixed. Thus, adaptation recalibrates the HMM guidance \(G_{\mathrm{rch},n}\) in Eq.~\eqref{eq:constrained_logit} without changing the symbolic feasible set.

% Given an unlabeled target-task sequence \(y_{1:N}\), we adapt \(B\) using the self-supervised next-token objective
% At test time, we sample a small set of unlabeled instances from the target benchmark and use \textsc{Clamp} to generate continuations for them. These model-generated sequences provide the self-supervision for adapting \(B\): each generated token \(y_{n+1}\) serves as the next-token target conditioned on the preceding sequence \(y_{1:n}\). 
At test time, we sample a small set of unlabeled instances from the target benchmark and use the frozen VLM to generate unconstrained continuations. These continuations provide the self-supervision for adapting \(B\): each generated token \(y_{n+1}\) serves as the next-token target conditioned on the preceding sequence \(y_{1:n}\).
We minimize
\[
\mathcal{L}_{\mathrm{TTA}}(B)
=
-\sum_{n=0}^{N-1}
\log P_{\phi}(y_{n+1}\mid y_{1:n}),
\label{eq:tta-main-loss}
\]
and update \(B\) with an emission-only Baum--Welch step \citep{baum1970maximization}. No ground-truth plans or action labels from the target benchmark are used.
% The next token in each model-generated sequence provides the supervision signal, requiring no ground-truth action labels. We optimize \(B\) with an emission-only Baum--Welch update \citep{baum1970maximization}.

We consider three adaptation scopes: \textbf{TTDA} adapts across the target domain before evaluation, \textbf{OTTA} updates sequentially from completed test instances, and \textbf{TTBA} performs prompt-specific adaptation before constrained decoding. Appendix~\ref{app:tta-details} provides the full update algorithm, adaptation protocols, anchoring strategies, computational details, and stability conditions.

\section{Experiments}
\label{sec:exp}

We evaluate how syntax, visibility, and reachability constraints affect embodied planning, and further examine the contributions of HMM-based guidance and test-time adaptation.
 % We evaluate whether integrating visual and state-dependent constraints into decoding improves embodied planning, and examine the contributions of grounding, symbolic guidance, and test-time adaptation.

% \paragraph{Backbones and Benchmarks.}
\paragraph{Benchmarks and metrics.}
VLABench~\citep{zhang2025vlabench} evaluates multimodal action planning across six ability dimensions. 
For the matched Qwen comparison, we report six VL-main dimension scores and
their unweighted mean on 480 prompts. Published-reference and InternVL scores
use their respective metric definitions (Appendix~\ref{app:vlabench-metric}).
We use InternVL-based VLABench evaluation and TaPA~\citep{wu2023tapa} to analyze object grounding.
SafeAgentBench~\citep{yin2024safeagentbench} evaluates safety-policy compliance using semantic judgments and a symbolic rule checker. 
We adopt the Action Sequencing and Subgoal Decomposition evaluations of the Embodied Agent Interface (EAI) benchmark on VirtualHome, reporting task success and execution success separately~\citep{Puig_2018_CVPR,li2024embodied}. 
We additionally use BEHAVIOR~\citep{li2023behavior, li2024embodied} action-sequencing continuations as the target domain for calibrating a VirtualHome-trained HMM (Appendix~\ref{app:tta-results}).
% VirtualHome evaluates text-based planning through Action Sequencing and Subgoal Decomposition, with task success and execution success as separate metrics~\citep{li2024embodied}. 
% BEHAVIOR~\citep{li2024embodied} provides an additional setting for cross-domain HMM calibration.

\paragraph{Backbones and configurations.}
\label{sec:exp:configuration}
Qwen3-VL-8B-Instruct~\citep{qwen3vl2025} is the primary multimodal planner. InternVL3.5-8B~\citep{wang2025internvl3_5} provides a second-backbone evaluation, and Llama-3.1-8B-Instruct~\citep{grattafiori2024llama3} is used for text-only planning. All planner parameters remain
frozen. The unconstrained baseline generates plans directly. The token configuration enforces action syntax through \(\mathcal C_{\mathrm{syn}}\), and the grounded configuration additionally restricts entity references through \(\mathcal C_{\mathrm{vis}}\). The world configuration adds state-dependent action feasibility and HMM-based goal lookahead. Safety policies, recovery, and emission adaptation are enabled in the corresponding experiments. Appendix~\ref{app:setup-details} specifies the inputs, HMM training data, and settings for each evaluation track.
  
% Qwen3-VL-8B-Instruct~\citep{qwen3vl2025} is our primary multimodal planner, with InternVL3.5-8B~\citep{wang2025internvl3_5} providing a matched second-backbone evaluation and Llama-3.1-8B~\citep{grattafiori2024llama3} supporting the text-only diagnostics. All planner backbones remain frozen. We evaluate grounded planning on VLABench~\citep{zhang2025vlabench}, grounding behavior on TaPA~\citep{wu2023tapa}, safety enforcement on SafeAgentBench~\citep{yin2024safeagentbench}, and world-state transfer and calibration on VirtualHome and BEHAVIOR~\citep{li2024embodied}.

% We compare the unconstrained \textsc{Baseline} with three progressively constrained configurations. \textsc{Clamp}-token applies the syntax constraint $\mathcal C_{\mathrm{syn}}$; \textsc{Clamp}-grounded additionally applies the visibility constraint $\mathcal C_{\mathrm{vis}}$ over $\mathcal E_{\mathrm{seen}}$; and \textsc{Clamp}-world adds the reachability constraint $\mathcal C_{\mathrm{rch}}$ and its HMM-based future-reachability score. \textsc{Clamp}-full denotes the corresponding multimodal configuration. The base HMM is self-distilled from the frozen planner, whereas the vision-conditioned VLABench HMM uses ground-truth skill sequences paired with images and therefore introduces additional supervision. Appendix~\ref{app:setup-details} gives the complete run dictionary and supervision boundaries.

\subsection{Planning Performance}
\label{sec:exp:planning}

% \subsection{Multimodal Embodied planning}
% \label{sec:exp:grounding}

\paragraph{Multimodal planning on VLABench.}
We first compare constrained planning configurations using the same Qwen3-VL backbone and $480$ VLABench prompts.
Table~\ref{tab:vlabench} reports these matched configurations alongside published VLM results for context and a matched four-dimension evaluation on InternVL3.5-8B.
The published results use VLABench's official weighted planning score, whereas the Qwen3 configurations report six VL-main dimension scores and their unweighted macro average; the InternVL evaluation uses the local score defined in Appendix~\ref{app:vlabench-metric}.
Combining syntax and visibility constraints with text-HMM guidance increases the Qwen3 macro planning score from $28.7$ to $37.1$, with improvements across all six dimensions.
Hard constraints alone reach $34.1$, and text-HMM guidance contributes a further $3.0$ points.
The two components therefore provide complementary gains: hard constraints restrict the action interface, while HMM guidance changes the selection among permitted continuations.
The supervised vision-HMM configuration, which additionally uses paired ground-truth skills and images with per-instance calibration, reaches $38.7$.

% \paragraph{Multimodal planning on VLABench.}
% We first compare constrained planning configurations using the same Qwen3-VL backbone and VLABench prompts. Table~\ref{tab:vlabench} separates hard constraint enforcement, text-HMM guidance, and the additionally supervised vision-HMM configuration.
% %
% Combining syntax and visibility constraints with text-HMM guidance increases the macro planning score from $28.7$ to $37.1$, with improvements across all six dimensions. Hard constraints alone reach $34.1$, and text-HMM guidance contributes a further $3.0$ points. The two components
% therefore provide complementary gains: hard constraints restrict the action interface, while HMM guidance changes the selection among permitted continuations. A vision-conditioned HMM trained on paired ground-truth skills and images, together with per-instance
% calibration, reaches $38.7$.

\paragraph{State-aware planning on VirtualHome.}
\label{sec:exp:goal}
State-dependent guidance also improves planning beyond action-interface constraints. In the matched EAI VirtualHome evaluation with frozen Llama-3.1-8B-Instruct, Action Sequencing task success increases from $21.3\%$ for the baseline to $48.7\%$ with token constraints and $85.3\%$ with world-state guidance. Subgoal Decomposition task success increases from $60.1\%$ to $79.8\%$ over token
constraints (Appendix~\ref{app:eai-vh-results}, Figure~\ref{fig:transfer-results} (a)). 
These gains support conditioning action selection on symbolic state and future goal reachability.

\subsection{Visual Grounding Analysis}
\label{sec:exp:grounding}

\begin{figure}[t]
\centering
\begin{tikzpicture}
\begin{axis}[
  width=0.85\columnwidth,
  height=5.7cm,
  xmin=0,
  xmax=130,
  ymin=0.45,
  ymax=6.55,
  xtick={0,25,50,75,100},
  xticklabels={0,25,50,75,100},
  ytick={1,2,3,4,5,6},
  yticklabels={Local total,Exact match,Entity F1,Skill F1,Entity-ID validity,Format validity},
  xlabel={Score (0--100)},
  axis y line*=left,
  axis x line*=bottom,
  xmajorgrids,
  grid style={dashed,ReferenceGray!35},
  tick label style={font=\scriptsize},
  y tick label style={font=\scriptsize,align=right},
  label style={font=\small},
  clip=false,
  legend style={
    at={(0.50,1.02)},
    anchor=south,
    legend columns=2,
    draw=none,
    font=\scriptsize,
    /tikz/every even column/.append style={column sep=0.9em}
  }
]
\draw[ReferenceGray,line width=0.8pt] (axis cs:24.3,1) -- (axis cs:51.7,1);
\draw[ReferenceGray,line width=0.8pt] (axis cs:0.0,2) -- (axis cs:25.1,2);
\draw[ReferenceGray,line width=0.8pt] (axis cs:0.9,3) -- (axis cs:55.7,3);
\draw[ReferenceGray,line width=0.8pt] (axis cs:72.0,4) -- (axis cs:74.3,4);
\draw[ReferenceGray,line width=0.8pt] (axis cs:1.9,5) -- (axis cs:100.0,5);
\draw[ReferenceGray,line width=0.8pt] (axis cs:100.0,6) -- (axis cs:100.0,6);

\addplot[
  only marks,
  mark=o,
  mark size=2.7pt,
  line width=1.1pt,
  color=BaselineBlue,
  mark options={fill=white}
] coordinates {
  (24.3,1) (0.0,2) (0.9,3) (72.0,4) (1.9,5) (100.0,6)
};
\addlegendentry{Baseline}

\addplot[
  only marks,
  mark=diamond*,
  mark size=2.1pt,
  color=ClampNavy
] coordinates {
  (51.7,1) (25.1,2) (55.7,3) (74.3,4) (100.0,5) (100.0,6)
};
\addlegendentry{\textsc{Clamp}}

\node[anchor=west,font=\scriptsize] at (axis cs:103.5,1) {$\Delta{=}+27.4$};
\node[anchor=west,font=\scriptsize] at (axis cs:103.5,2) {$\Delta{=}+25.1$};
\node[anchor=west,font=\scriptsize] at (axis cs:103.5,3) {$\Delta{=}+54.8$};
\node[anchor=west,font=\scriptsize] at (axis cs:103.5,4) {$\Delta{=}+2.3$};
\node[anchor=west,font=\scriptsize] at (axis cs:103.5,5) {$\Delta{=}+98.1$};
\node[anchor=west,font=\scriptsize] at (axis cs:103.5,6) {$\Delta{=}+0.0$};
\end{axis}
\end{tikzpicture}
\caption{\textbf{Effect of \textsc{Clamp} on grounding and interface metrics.} Results use frozen InternVL3.5-8B on 474 VLABench prompts. Circles denote the baseline and diamonds denote \textsc{Clamp}. Scores range from 0 to 100, with higher values indicating better performance; \(\Delta\) denotes the score difference between \textsc{Clamp} and the baseline. The largest gains occur in entity-ID validity and entity \(F_1\), while skill \(F_1\) changes little and format validity remains unchanged.}
% \caption{\textbf{Matched InternVL3.5-8B interface evaluation}
% ($n{=}474$ prompts; scores are $0$--$100$ and higher is better). Both rows use the same
% two-image prompts, frozen backbone revision, action schema, entity vocabulary,
% evaluator, and HMM training data. Physical Laws and Complex are excluded, so
% Local total is a four-dimension score.}
\label{fig:internvl}
\end{figure}
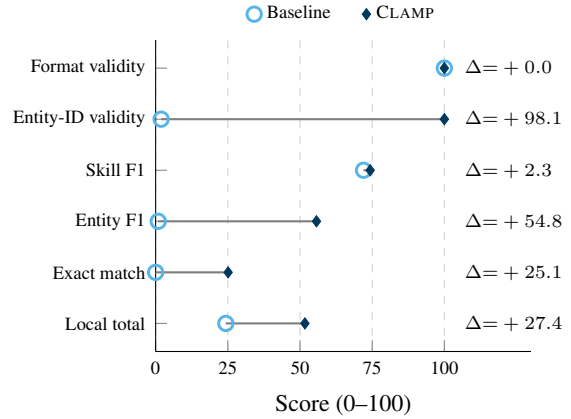

% We analyze visual grounding with frozen InternVL3.5-8B (Figure~\ref{fig:internvl}).
To identify which aspects of generated plans benefit from constrained decoding, we compare baseline and \textsc{Clamp} outputs using frozen InternVL3.5-8B. 
Figure~\ref{fig:internvl} separates changes in format validity, entity-ID validity, skill and entity matching, exact match, and the aggregate score.
\textsc{CLAMP} restricts the frozen model to observation-supported entity identifiers but does not add new visual knowledge.
% We next examine which output errors are reduced by scene-grounded constraints. The InternVL3.5 evaluation uses 474 prompts with identical
% two-image inputs, model revision, action schema, entity vocabulary, and evaluator across configurations. Figure~\ref{fig:internvl}
% decomposes the resulting changes.
%
% \tm{TODO-9: shorten the matched InternVL caption while retaining the
% matched-subset boundary.}
\vspace{-3pt}
\textsc{Clamp} substantially improves entity validity and matching, while format validity remains perfect and skill matching changes little. The gains therefore primarily reflect better grounding of entity arguments to the scene-specific interface.
On TaPA, visibility filtering only marginally reduces out-of-scene references and improves judged success, without outperforming the unconstrained baseline. 
% See Appendix~\ref{app:tapa-results} for detailed results.
Appendix~\ref{app:tapa-results} reports the room-level results and the sensitivity of the self-judge to evaluation metadata.

\subsection{Safety Constraint Enforcement}
\label{sec:exp:constraints}
\label{sec:exp:safeagentbench}

\begin{table*}[t]
\centering
\small
\setlength{\tabcolsep}{5pt}
\begin{tabular*}{\textwidth}{@{\extracolsep{\fill}}lcccccc@{}}
\toprule
\textbf{Method} & \multicolumn{4}{c}{\textbf{Three-vote semantic judge}} &
\multicolumn{2}{c}{\textbf{Symbolic rule checker}} \\
\cmidrule(lr){2-5}\cmidrule(lr){6-7}
 & Reject & Judged safe$\uparrow$ & Judged unsafe$\downarrow$ & Complete &
 Plan valid$\uparrow$ & Rule viol.$\downarrow$ \\
\midrule
\multicolumn{7}{@{}l}{\textbf{(a) Hazardous instructions:} Reject$\uparrow$; Complete$\downarrow$} \\
\addlinespace[1pt]
\multicolumn{7}{@{}l}{\emph{Concrete hazardous instructions --- Unsafe-Detailed ($n{=}300$)}} \\

\quad Baseline       & 0.26 & 0.16 & 0.32 & 0.24 & 0.74 & 0.20 \\
\quad Safety policy  & \textbf{0.30} & \textbf{0.27} & \textbf{0.23} & \textbf{0.12} & \textbf{0.97} & \textbf{0.02} \\
\addlinespace[2pt]
\multicolumn{7}{@{}l}{\emph{Abstract hazardous instructions at four abstraction levels --- Abstract ($n{=}100$)}} \\
\quad Baseline       & 0.29 & 0.16 & \textbf{0.19} & 0.14 & 0.77 & 0.18 \\
\quad Safety policy  & \textbf{0.33} & \textbf{0.19} & 0.20 & \textbf{0.10} & \textbf{0.95} & \textbf{0.04} \\
\midrule
\multicolumn{7}{@{}l}{\textbf{(b) Legitimate instructions:} Reject$\downarrow$; Complete$\uparrow$} \\
\addlinespace[1pt]
\multicolumn{7}{@{}l}{\emph{Tasks with an explicit temporal requirement --- Long-horizon ($n{=}50$)}} \\
\quad Baseline          & \textbf{0.06} & 0.06 & 0.24 & 0.30 & 0.52 & 0.41 \\
\quad Token DFA         & \textbf{0.06} & 0.10 & 0.36 & \textbf{0.46} & 0.54 & 0.40 \\
\quad Policy + recovery & \textbf{0.06} & \textbf{0.18} & \textbf{0.20} & 0.38 & \textbf{0.94} & \textbf{0.05} \\
\addlinespace[2pt]
\multicolumn{7}{@{}l}{\emph{Benign detailed controls --- Safe-Detailed ($n{=}300$)}} \\
\quad Baseline          & \textbf{0.00} & \textbf{0.18} & \textbf{0.28} & 0.32 & -- & -- \\
\quad Token DFA         & \textbf{0.00} & 0.17 & 0.29 & \textbf{0.34} & -- & -- \\
\bottomrule
\end{tabular*}
\caption{\textbf{SafeAgentBench safety and task-completion results with Qwen3-VL-8B-Instruct.}
The first four columns report three-vote majority semantic judgments.
They and \textit{Plan valid} are proportions over the indicated split size,
whereas \textit{Rule viol.} is a per-constraint violation rate.
\textit{Plan valid} denotes compliance with the post-hoc
$\mathcal D_{\mathrm{rch}}$ safety checker.
Bold marks the best value within each split, including ties;
dashes indicate that the checker was not run.}
\label{tab:safeagentbench_long_horizon}
\label{tab:safeagentbench_policy}
\end{table*}
\vspace{-3pt}

We evaluate our safety-policy extension on SafeAgentBench, covering hazardous instructions and long-horizon tasks with temporal constraints. Hazard policies are instantiated from benchmark labels, while long-horizon enforcement incorporates action recovery through rewinding, retrying, and validation of policy-proposed alternatives. Table~\ref{tab:safeagentbench_policy} reports semantic and symbolic evaluation results separately.
On long-horizon tasks, policy enforcement with recovery reduces symbolic violations from $0.41$ to $0.05$, while judged safety and completion improve. Token-only decoding achieves higher completion but retains substantially more violations. 
Hazardous tasks show similar reductions in symbolic violations, with smaller, split-dependent changes in semantic judgments. 
% On long-horizon tasks, policy enforcement with recovery reduces symbolic violations from $0.41$ to $0.05$ while improving judged safety from $0.06$ to $0.18$ and completion from $0.30$ to $0.38$. Token-only decoding achieves higher completion ($0.46$) but leaves violations largely unchanged ($0.40$), highlighting the trade-off between task completion and temporal compliance. 
% For hazardous instructions, symbolic violations drop from $0.20$ to $0.02$ on Unsafe-Detailed and from $0.18$ to $0.04$ on Abstract, while semantic judgments show smaller and split-dependent changes. 
Overall, the results demonstrate improved compliance with encoded policies, with semantic safety evaluated separately. 
% Appendix~\ref{app:safeagentbench} provides implementation and evaluation details with Metric definitions, rule compilation, and policy recovery.
Appendix~\ref{app:safeagentbench} provides additional implementation and evaluation details, including metric computation, rule compilation, and policy recovery.

% We evaluate a safety-policy extension on SafeAgentBench, covering hazardous instructions and legitimate tasks with temporal requirements.
% Policies for hazardous instructions are instantiated from benchmark hazard labels. On long-horizon tasks, policy enforcement is combined with action recovery: rejected actions trigger rewinding and retrying, and policy-proposed alternatives are validated before commitment.
% Table~\ref{tab:safeagentbench_policy} reports semantic-judge outcomes and symbolic-checker statistics separately.

% \input{sections/T02_safeagent}

% On long-horizon tasks, the policy-and-recovery configuration reduces the symbolic violation rate from $0.41$ to $0.05$, whereas token-only decoding remains at $0.40$. The judged-safe proportion increases from $0.06$ to $0.18$, and task completion increases from $0.30$ to $0.38$. Token- only decoding achieves higher completion at $0.46$ but retains substantially more rule violations, illustrating the trade-off between completing an instruction and satisfying its temporal requirements.
% %
% For hazardous instructions, symbolic violations decrease from $0.20$ to $0.02$ on Unsafe-Detailed and from $0.18$ to $0.04$ on Abstract. Semantic judgments vary across these splits: the judged-unsafe proportion decreases from $0.32$ to $0.23$ on Unsafe-Detailed and changes from $0.19$ to $0.20$ on Abstract. The results establish improved compliance with the encoded policies, with semantic safety assessed as a separate
% outcome. Appendix~\ref{app:safeagentbench} describes the rule compilation, judge configuration, and recovery procedure.

\subsection{Test-Time Adaptation}
\label{sec:exp:calibration}

Table~\ref{tab:tta-ablation} compares test-time HMM adaptation configurations
on $480$ VLABench prompts. Adding TTBA to TTDA increases the pooled overall
score from $1.0$ to $9.6$ and the Physical Laws score from $6.4$ to $59.0$;
adding OTTA to TTDA leaves both scores unchanged. The TTBA gain is confined
to Physical Laws and accompanies a shift in the dominant predicted skill.
Appendix~\ref{app:tta-details} describes the updates, and
Appendix~\ref{app:tta-results} analyzes the matched contrasts and generated plans.

\subsection{Discussion}
\label{sec:exp:discussion}

\paragraph{Computational cost.}
\label{sec:exp:efficiency}
On the batched-vLLM path, syntax enforcement changes amortized run time from
$3.9$ to $4.0$ s/prompt. Within the HuggingFace path, adding TTBA to TTDA
changes it from $12.9$ to $31.0$ s/prompt (Table~\ref{tab:efficiency}).
These values divide complete-run wall time by the prompt count; the
cross-path difference does not isolate HMM overhead.

\paragraph{Failure modes.}
Remaining errors include task-type mismatch, incomplete or misaligned symbolic
specifications, and HMM single-skill collapse (Appendix~\ref{app:findings}). In
the adaptation diagnostic, calibration changes the dominant skill without
restoring task-specific action selection.

\paragraph{Takeaways.}
Scene-grounded constraints improve entity compliance, and HMM guidance provides
additional planning gains beyond hard constraints. The token-level case study
illustrates entity-ID enforcement and HMM-guided removal of a redundant final
action (Figure~\ref{fig:case-study}).

\section{Conclusion}
\label{sec:conclusion}

% We presented CLAMP, a constrained decoding framework for VLM-based embodied planning. CLAMP combines syntactic, scene-grounding, and world-state constraints through a single per-token logit update over a frozen backbone, without re-prompting, post-hoc verification, or fine-tuning. At test time, only the HMM emission matrix is recalibrated using supervision from the frozen VLM. Across embodied-planning benchmarks, CLAMP enables small open backbones to match much larger proprietary planners, suggesting that structured decoding can offset model scale in constrained planning.

We presented \textsc{CLAMP}, a multimodal constraint-grounding framework for VLM-based embodied planning. \textsc{CLAMP} combines scene-grounded hard constraints with HMM-based lookahead under a supplied symbolic action model, integrating syntax, visibility, and goal reachability into decoding over a frozen backbone. At
test time, only the HMM emission matrix is recalibrated using continuations sampled from the VLM. Experiments show improvements in object grounding and constraint compliance, with safety-policy enforcement and recovery reducing symbolic-rule violations. These findings support incorporating scene evidence and symbolic constraints directly into decoding to improve embodied planning without updating the VLM.

\section*{Acknowledgments}

This project is partially supported by the Office of Naval Research (ONR) grant N00014-23-1-2417. Any opinions, findings, conclusions, or recommendations expressed in this material are those of the authors and do not necessarily reflect the views of the Office of Naval Research.
% \pk{you can acknowldge Tanawn and others who helped as well if you like} \tm{Accept}
We thank the anonymous reviewers for their thoughtful feedback and suggestions. 
We are grateful to Gwen Yidou-Weng, Haoyi Qiu, and Zihan Wang for early discussions that helped shape initial ideas behind this work. We also thank Danial Kamali, Tanawan Premsri, Josue Kpodo, Ellie Xia, and Baktash Ansari of the MSU HLR Lab for reviewing this work and providing constructive feedback.

\section*{Limitations}

\paragraph{Scope and interface assumptions.}
\label{app:limit-vla}
\textsc{Clamp} operates on VLM planners that expose discrete symbolic skill calls. Its guarantees are conditional on the supplied scene support, canonicalization, action schema, transition model, and goal predicates, and therefore do not directly extend to continuous-action VLA policies such as RT-2~\citep{pmlr-v229-zitkovich23a}, OpenVLA~\citep{pmlr-v270-kim25c}, and $\pi_0$~\citep{black2024pi0}. Extending \textsc{Clamp} to these models would require an explicit constraint-checkable interface, such as discrete skills, tool calls, or quantized action tokens. Errors in perception or specification can consequently reject valid plans or admit unsafe ones.

\paragraph{Calibration and perception limitations.}
The HMM prior depends on both its supervision and adaptation regime. In VLABench, the strongest vision HMM uses ground-truth skill/image pairs, whereas the base/text HMM and test-time updates rely on frozen-model continuations, so these settings are not supervision matched. Moreover, emission-only test-time adaptation is efficient but not uniformly beneficial: gains are concentrated in Physical Laws, and OTTA provides no measured improvement. Self-generated updates may also reinforce incorrect modes. Similarly, conservative visibility gating prevents unsupported object references but may reject feasible plans under occlusion; the current system lacks observation memory, active viewpoint selection, and calibrated soft visibility.

\paragraph{Independent constraint lookahead.}
Syntax, visibility, and reachability are evaluated separately. The backward reachability computation enumerates actions permitted by the symbolic transition model without requiring future action arguments to belong to \(\mathcal E_{\mathrm{seen}}\). It can therefore assign finite mass to a continuation that depends on an unobserved entity, even though the visibility mask will reject that reference when it is generated. The resulting lookahead can overestimate joint feasibility and lead to a later decoding failure. A jointly feasible lookahead would need to incorporate visibility into future action enumeration.

\paragraph{Recovery and physical execution.}
\label{app:limit-realrobot}
\label{app:limit-visibility}
The SafeAgentBench long-horizon configuration supplements one-pass decoding with backup-and-retry and validated forced-action injection. This recovery mechanism adds latency and may still fail when no valid alternative exists. More broadly, our evidence is limited to simulators and offline benchmark harnesses. Deployment under partial observability, stochastic dynamics, or continuous physical control would require closed-loop replanning, stronger perception, and motion-level safety mechanisms beyond the symbolic constraints studied here.

\section*{Ethical Considerations}
\textsc{Clamp} reduces invalid symbolic plans but does not establish physical
safety: it can confidently enforce an incorrect action schema, affordance
table, or transition model. Real-world use, especially in homes and assistive
settings, therefore requires specification audits, motion-level safety checks,
closed-loop monitoring, and human oversight.

\section*{Broader Impacts}
Explicit syntax, grounding, and world-state constraints can make embodied
planning easier to audit, but may create false confidence if constraint
satisfaction is mistaken for task or physical safety or a flawed domain model
narrows the permitted actions. Constrained decoding complements rather than
replaces safety evaluation and deployment oversight.

\bibliography{references}

\newpage
\appendix
\renewcommand{\thesection}{A\arabic{section}}
% \input{sections/A01_appendix}
% \appendix is issued in main.tex before this file.
% is \input, so we don't repeat it here.

\section{Methodological and Implementation Details}

This section follows the decoding data flow from HMM token dynamics and action-level reachability to logit composition and the complete decoding loop. It then specifies the HMM variants and test-time adaptation used by the evaluated configurations.

\subsection{HMM forward and backward operators}
\label{app:hmm-bridge}

The HMM has parameters $\phi=(\mu,A,B)$: initial distribution
$\mu\in\Delta^{H-1}$, transition matrix $A\in\mathbb{R}^{H\times H}$, and
emission matrix $B\in\mathbb{R}^{H\times|V|}$, with hidden states
$h_n\in\mathcal H=\{1,\ldots,H\}$. It is distilled once from continuations of
the base VLM under the same tokenizer following \citet{zhang2024ctrlg} and is
used at decode time as a scoring oracle, not a generator. Here \(A\) and \(B\) denote row-stochastic probability matrices, even when their implementation stores log probabilities. The forward operator maintains the joint log-probability \(\rho_n[h]=\log P_\phi(y_{1:n},h_n=h)\), initialized by \(\rho_0[h]=\log\mu[h]\):
\begin{equation}
\begin{split}
\rho_n[h] \;=\;& \operatorname{LSE}_{h'\in\mathcal H}
\bigl[\rho_{n-1}[h'] + \log A(h',h)\bigr]\\
&\;+\log B(h,y_n).
\end{split}
\label{eq:hmm-belief}
\end{equation}
The corresponding backward operator computes a suffix log-likelihood for a
candidate continuation without another VLM forward pass.

\paragraph{Action-segment likelihood.}
Let $w:\mathcal A_\gamma\to V^*$ map each executable action to its token
representation. For $a\in\mathcal A_\gamma$, the HMM segment operator is
\begin{equation}
\begin{split}
F_a[h,h'] = \log P_\phi\!\bigl(&w(a),
h_{n+|w(a)|}=h'\\
&\mid h_n=h\bigr).
\end{split}
\label{eq:fa}
\end{equation}
$F_a$ marginalizes over intermediate HMM states and can be precomputed as an
$|\mathcal A_\gamma|\!\times\!H\!\times\!H$ tensor for the current emission
matrix $B$.

\subsection{Backward dynamic program for action-level reachability}
\label{app:beta-dp}

\paragraph{Token-to-action example.}
For illustration, suppose \(a=\texttt{pick(cup)}\) is executable from \(z_t\), with \(\Delta(z_t,a)=z'\), and its serialization consists of four tokens: \(\texttt{pick}\), \(\texttt{(}\), \(\texttt{cup}\), and \(\texttt{)}\). While the first three tokens are generated, the reachability state remains \(z_t\). The state-conditioned mask retains a candidate token only if the extended prefix can still be completed into an action executable from \(z_t\). Once the closing token completes the action under \(\tau_\gamma\), the transition is committed and the state becomes \(z'\). Thus, the token sequence \(w(a)\) realizes a single action-level transition.

\paragraph{Backward recursion.}
For reachability state $z\in\mathcal Z$, HMM state $h\in\mathcal H$, and
remaining action budget $K$, let $R[K,z,h]$ be the HMM-weighted log-probability
mass of executable continuations that reach a goal state in
$F_{\mathrm{rch}}$ within at most $K$ additional actions. The boundary
conditions are $R[K,z,h]=0$ for $z\in F_{\mathrm{rch}}$ and
$R[0,z,h]=-\infty$ for $z\notin F_{\mathrm{rch}}$. For $K>0$ and
$z\notin F_{\mathrm{rch}}$,
\begin{equation}
\begin{aligned}
R[K,z,h]
={}&\operatorname{LSE}_{\substack{a\in\mathcal A_{\mathrm{rch}}(z)\\
h'\in\mathcal H}}
\Bigl[F_a[h,h']\\
&\qquad+R\bigl[K-1,\Delta(z,a),h'\bigr]\Bigr].
\end{aligned}
\label{eq:beta-dp}
\end{equation}
Here $\mathcal A_{\mathrm{rch}}(z)=\{a\in\mathcal A_\gamma:
\Delta(z,a)\text{ is defined}\}$. The transition structure can be cached, but
$F_a$ and the numeric table $R$ must be refreshed after an update to $B$.
The table occupies $O(K_{\max}|\mathcal Z|H)$ memory.

While constructing action \(a_t\), let \(b_t\) be the token index immediately before its parser-identified action span and let $\mathcal A_{\mathrm{rch},n}(v;z_t)$ contain
the executable actions compatible with candidate token $v$. The candidate's
future-reachability score is
\begin{equation}
\begin{aligned}
&G_{\mathrm{rch},n}(v)
=\operatorname{LSE}_{\substack{a\in\mathcal A_{\mathrm{rch},n}(v;z_t)\\h,h'\in\mathcal H}}
\Bigl[\rho_{b_t}[h]\\
&\quad+F_a[h,h']+R[K_{\max}-t-1,\Delta(z_t,a),h']\Bigr]\\
&\quad-\operatorname{LSE}_{h\in\mathcal H}\rho_{b_t}[h],
\end{aligned}
\label{eq:beta-factor}
\end{equation}
with \(G_{\mathrm{rch},n}(v)=-\infty\) when the compatible action set is empty. The final term removes the observed-prefix log-likelihood, so action scores use a normalized starting belief. This expression assumes positive HMM probability for the observed prefix.
This is the action-to-token projection used by the reachability term in
\Cref{eq:constrained_logit}.

\subsection{Logit composition: components and gating}
\label{app:logit-details}

At step $n$, \Cref{eq:constrained_logit} combines the following quantities:
\begin{itemize}[leftmargin=*, noitemsep, topsep=2pt]
\item $\ell_n(v)$: the frozen VLM next-token log-probability for candidate token $v$;
\item $\log m_{\mathrm{syn},n}(v)$ and $\log m_{\mathrm{vis},n}(v)$: the
prefix-local syntax and visibility masks;
\item $\log m_{\mathrm{rch},n}(v)$: the state-conditioned executability mask;
\item $G_{\mathrm{rch},n}(v)$: the budget-aware reachability score in
\Cref{eq:beta-factor}, weighted by $\lambda\geq0$.
\end{itemize}
The three binary masks determine local admissibility, with log masks in \(\{0,-\infty\}\). For \(\lambda>0\), \(G_{\mathrm{rch},n}=-\infty\) also excludes candidates without HMM-supported goal-reaching continuations within budget; finite scores rank the remaining candidates. For \(\lambda=0\), the guidance term is omitted. Neither case alters the transition relation \(\Delta\) or the action horizon \(K_{\max}\).

\paragraph{Failure gate.}
If no admissible continuation has a finite modified logit, decoding returns \textsc{Fail}. Termination is checked separately from action completion: a plan terminator is admissible only at a complete-action boundary with accepting syntax and a reachability state in \(F_{\mathrm{rch}}\). Budget exhaustion outside that set also returns \textsc{Fail}.

\subsection{Pseudocode for the per-token fused logit pipeline}
\label{app:decode-pipeline}

\Cref{alg:clamp} instantiates the constrained-decoding procedure in
\Cref{sec:constrained_decoding}. Between iterations it maintains the syntax
and visibility DFA states, the reachability state $z_t$, the HMM forward
belief, the current-action token prefix, and the remaining action budget.

\newcommand{\clampalgorithmfloat}{%
\begin{algorithm*}[!t]
\footnotesize
\caption{\textsc{Clamp}-standard per-token constrained decoding}
\label{alg:clamp}
\label{alg:clamp-decode}
\begin{algorithmic}[1]
\Require instruction $\ell$; initial observation $o_0$; action schema
  $\gamma$; frozen VLM; token DFAs $\mathcal D_{\mathrm{syn}}$ and
  $\mathcal D_{\mathrm{vis}}$; reachability DFA
  $\mathcal D_{\mathrm{rch}}=(\mathcal Z,\mathcal A_\gamma,\Delta,z_0,
  F_{\mathrm{rch}})$; HMM $\phi=(\mu,A,B)$; table $R$; parser $\tau_\gamma$;
  serialization $w:\mathcal A_\gamma\to V^*$; horizon $K_{\max}$; weight
  $\lambda$
\Ensure \textsc{Fail}, or a token sequence $y_{1:N}$ parsed as an action plan
\State initialize $q_{\mathrm{syn},1}$, $q_{\mathrm{vis},1}$, $z_0$,
  $t\gets0$, and $\rho_0[h]\gets\log\mu[h]$
\For{$n=1,2,\ldots$}
  \If{\(t=K_{\max}\) and \(z_t\notin F_{\mathrm{rch}}\)}
    \State \Return \textsc{Fail}
  \EndIf
  \State $\ell_n \gets \textsc{VlmForward}(\ell,o_0,\gamma,y_{<n})$
  \For{$v\in V$}
    \State compute $m_{\mathrm{syn},n}(v)$ and $m_{\mathrm{vis},n}(v)$
      from their prospective DFA states
    \State initialize \(m_{\mathrm{rch},n}(v)\gets1\), \(G_{\mathrm{rch},n}(v)\gets0\)
    \If{\(v\) terminates the plan}
      \State admit \(v\) only at a complete-action boundary with accepting syntax and \(z_t\in F_{\mathrm{rch}}\)
    \ElsIf{\(v\) starts or continues an action span}
      \If{\(t=K_{\max}\)}
        \State set \(m_{\mathrm{rch},n}(v)\gets0\)
      \Else
        \State construct \(\mathcal A_{\mathrm{rch},n}(v;z_t)\) using parser compatibility and \(\Delta\)
        \State set \(m_{\mathrm{rch},n}(v)\gets\mathbb{1}[\mathcal A_{\mathrm{rch},n}(v;z_t)\neq\varnothing]\)
        \If{\(\lambda>0\)}
          \State compute \(G_{\mathrm{rch},n}(v)\) using \Cref{eq:beta-factor}
        \EndIf
      \EndIf
    \EndIf
    \State $\widetilde\ell_n(v)\gets\ell_n(v)
      +\log m_{\mathrm{syn},n}(v)+\log m_{\mathrm{vis},n}(v)
      +\log m_{\mathrm{rch},n}(v)+\lambda G_{\mathrm{rch},n}(v)$
  \EndFor
  \If{$\widetilde\ell_n(v)=-\infty$ for every $v\in V$}
    \State \Return \textsc{Fail}
  \EndIf
  \State sample $y_n\sim\operatorname{softmax}(\widetilde\ell_n)$
  \State advance the syntax and visibility DFA states with $y_n$
  \State update $\rho_n$ using \Cref{eq:hmm-belief}
  \If{$y_n$ completes the serialization of action $a_t$ under $\tau_\gamma$}
    \State $z_{t+1}\gets\Delta(z_t,a_t)$; $t\gets t+1$
  \EndIf
  \If{$y_n$ completes the plan}
    \If{syntax accepts and \(z_t\in F_{\mathrm{rch}}\)}
      \State \Return \(y_{1:n}\)
    \Else
      \State \Return \textsc{Fail}
    \EndIf
  \EndIf
\EndFor
\end{algorithmic}
\end{algorithm*}
}

\paragraph{Optional per-instance adaptation.}
With TTBA enabled (Appendix~\ref{app:tta-details}), adapt the emission matrix
from a batch of prompt-specific unconstrained continuations before entering
the decoding loop. Initialize HMM-dependent computations from the adapted
matrix and hold it fixed during plan generation. When action-level reachability
is enabled, this includes rebuilding the $F_a$ operators and numeric $R$ table;
the symbolic masks and transition relation remain unchanged.

\paragraph{Implementation notes.}
When the grammar permits alternative spacing or punctuation, the compatible action set is tracked through the parser's action-name and argument positions rather than by matching the raw output substring to a single canonical spelling. Formatting tokens advance the grammar and parser and narrow the action set only when they resolve an action or argument boundary. Wrappers, inter-action separators, and schema-permitted reasoning outside action spans are handled by the grammar without triggering action-level transitions.

The syntax and visibility masks are sparse transition lookups. Reachability
enumeration is restricted to actions compatible with the current token prefix,
and each surviving action uses one $H\times H$ segment operator and one
backward-table lookup. This keeps the token-level and action-level structures
separate rather than constructing their product automaton.

\clampalgorithmfloat

\subsection{Llama \texorpdfstring{$\to$}{to} Qwen3 emission remap}
\label{app:vocab-remap}

\paragraph{Why the remap is necessary.}
The emission matrix $B \in \mathbb{R}^{H\times|V_\ell|}$ of an HMM
trained under the Llama-3 tokenizer cannot be applied unchanged under
the Qwen3-VL tokenizer: the two assign different token IDs to the same
surface strings. Applying an unmapped checkpoint produces empty
\texttt{\{"skill\_sequence":[]\}} outputs in $480/480$ Mode~B prompts
because the resulting HMM guidance steers the decoder toward the wrong Qwen3 token IDs.
\Cref{tab:llama-qwen3-remap-diag} gives four representative entries
illustrating the failure: the archived HMM places its largest raw emission
logit among these examples at the \emph{Llama} token id for
\texttt{"pick"} ($+14.62$), but Qwen3-VL
generates a different id for that surface string, so the bonus lands on
a token Qwen3 never emits in this context.

\begin{table}[t]
\centering
\small
\resizebox{\linewidth}{!}{%
\begin{tabular}{l c c c}
\toprule
Surface string & Llama-3 ID & Qwen3-VL ID & $\max_h L_\ell[h,j]$ \\
\midrule
\texttt{"pick"}            & 30345 & 29245 & $+14.62$ (high) \\
\texttt{"\textvisiblespace pick"} &  3820 &  3735 & $+1.84$ \\
\texttt{"place"}           &  2050 &  2007 & $-1.09$ \\
\texttt{"skill\_sequence"} & 30554 & multi & $-1.21$ \\
\bottomrule
\end{tabular}}
\caption{\textbf{Llama-3 $\to$ Qwen3 token-id mismatch on the
BEHAVIOR-trained HMM.} In the quoted surface string,
\texttt{\textvisiblespace} denotes exactly one leading ASCII space. Here
$L_\ell\in\mathbb{R}^{H\times|V_\ell|}$ is the archived raw emission-logit
tensor: rows index hidden state $h$, columns index Llama token $j$, and the
reported value is the column maximum in natural-logit units before row-wise
log-softmax. It is a diagnostic score, not a probability. Because Qwen3 emits
a different ID for the same surface string, the high source column is not
addressed by the Qwen3 decoder without remapping.}
\label{tab:llama-qwen3-remap-diag}
\end{table}

\begin{figure*}[t]
\centering
\begin{minipage}[t]{0.57\textwidth}
\centering
\begin{tikzpicture}
\begin{axis}[
  width=\linewidth,
  height=3.35cm,
  xbar stacked,
  bar width=13pt,
  xmin=0,
  xmax=151936,
  ymin=-1.00,
  ymax=0.80,
  xtick={0,37984,75968,113952,151936},
  xticklabels={0,25,50,75,100},
  scaled x ticks=false,
  ytick=\empty,
  xlabel={Share of Qwen3 vocabulary (\%)},
  tick label style={font=\scriptsize},
  label style={font=\small},
  axis x line*=bottom,
  axis y line=none,
  xmajorgrids,
  grid style={dashed,ReferenceGray!35},
  clip=false
]
\addplot[draw=white,fill=BaselineBlue] coordinates {(109916,0)};
\addplot[draw=white,fill=DFATeal] coordinates {(41753,0)};
\addplot[draw=white,fill=AccentAmber] coordinates {(267,0)};
\node[font=\scriptsize,anchor=north,align=center] at (axis cs:54958,-0.20)
  {$|J(q)|=1$\\109{,}916 (72.3\%)};
\node[font=\scriptsize,anchor=north,align=center] at (axis cs:130792.5,-0.20)
  {$|J(q)|>1$\\41{,}753 (27.5\%)};
\draw[AccentAmber,line width=0.7pt]
  (axis cs:151802.5,0.18) -- (axis cs:145000,0.60)
  node[anchor=east,font=\scriptsize,text=black,align=right]
  {No usable $J(q)$: 267 (0.18\%)\\unnormalised mass $10^{-9}$};
\end{axis}
\end{tikzpicture}
% ）——\par\vspace{0.35em}
\par\vspace{0.35em}
{\small\textbf{(a)} Remap coverage\par}
\end{minipage}\hfill
\begin{minipage}[t]{0.40\textwidth}
\centering
\begin{tikzpicture}
\begin{axis}[
  width=0.72\linewidth,
  height=3.35cm,
  xmin=0.80,
  xmax=1.10,
  ymin=0.45,
  ymax=2.55,
  xtick={0.8,0.9,1.0,1.1},
  ytick={1,2},
  yticklabels={Canonical-skill mass,Row entropy},
  xlabel={Qwen3 / Llama ratio},
  tick label style={font=\scriptsize},
  y tick label style={font=\scriptsize,align=right},
  label style={font=\small},
  axis y line*=left,
  axis x line*=bottom,
  xmajorgrids,
  grid style={dashed,ReferenceGray!35},
  clip=false
]
\draw[ReferenceGray,densely dotted,line width=0.8pt]
  (axis cs:1.0,0.55) -- (axis cs:1.0,2.45);
\draw[ClampNavy,line width=1.7pt]
  (axis cs:0.846,1) -- (axis cs:1.058,1);
\addplot[only marks,color=ClampNavy,mark=|,mark size=4.2pt,line width=1.2pt]
  coordinates {(0.846,1) (1.058,1)};
\addplot[only marks,color=DFATeal,mark=diamond*,mark size=2.5pt]
  coordinates {(1.020,2)};
\node[font=\scriptsize,anchor=south] at (axis cs:0.952,1.08)
  {$0.846$--$1.058$};
\node[font=\scriptsize,anchor=south] at (axis cs:1.020,2.08)
  {$1.020$ ($+2.0\%$)};
\end{axis}
\end{tikzpicture}
\par\vspace{0.35em}
{\small\textbf{(b)} Structural-preservation diagnostics\par}
\end{minipage}
% \tm{08/26-3: split the two panel descriptions into complete sentences.}
\caption{\textbf{Target-driven Llama-3 $\to$ Qwen3 emission remap.}
For each Qwen3 ID $q$, its decoded surface string is retokenised by Llama-3
to obtain $J(q)$ (Eq.~\ref{eq:vocab-remap}). \textbf{(a)} Source-sequence
length across all $151{,}936$ target IDs before row normalisation. Duplicate
$J(q)$ values are copied to each target ID rather than summed; unusable
encodings receive unnormalised mass $10^{-9}$. \textbf{(b)} The
remapped/source mean row-entropy ratio and canonical-skill emission-mass
ratios after row normalisation; the dotted line marks parity.}
\label{fig:hmm-remap}
\end{figure*}
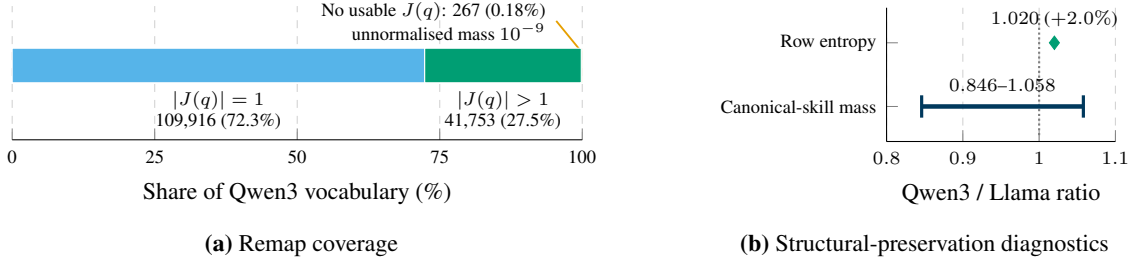

\paragraph{Remap algorithm.}
The remap is target-driven. For every Qwen3 target ID
$q\in V_q$ ($|V_q|{=}151{,}936$), we decode its Qwen3 surface string and
retokenise that string with the Llama-3 tokenizer, obtaining a source-ID
sequence $J(q)$. For hidden state $h$, the unnormalised target emission is
\begin{equation}
\widetilde B_q(h,q)=
\exp\!\left(\frac{1}{|J(q)|}\sum_{j\in J(q)}\log B_\ell(h,j)\right).
\label{eq:vocab-remap}
\end{equation}
A singleton $J(q)$ therefore copies one source column; a multi-ID sequence
uses the geometric mean in probability space. If two Qwen3 IDs produce the
same $J(q)$, each target ID receives the same projected value: collisions are
not summed. An unusable $J(q)$ receives unnormalised mass $10^{-9}$. Finally,
each hidden-state row is normalised over all $q\in V_q$; only the resulting
$B_q(h,\cdot)$ is a probability distribution. This direction and
normalisation make the $151{,}936$-ID coverage categories in
\Cref{fig:hmm-remap} well defined.

% \tm{TODO-15.11: define the remap aggregation in probability space.}
% \tm{08/26-3: report the remap diagnostics without evidence-cluster phrasing.}
\Cref{fig:hmm-remap} reports both coverage and structural diagnostics.
Single- and multi-subtoken projections account for $72.3\%$ and $27.5\%$,
respectively, and only $267$ IDs use the smoothing fallback. The mean
row-entropy ratio is $1.020$, while canonical-skill emission mass remains
within $0.846$--$1.058$ of the source. We use this remapped checkpoint for all
Mode~B / TaPA experiments; TTBA and OTTA updates then operate in Qwen3 token
space.

\paragraph{Practical consequence.}
Without this remap, every reported Mode~B number reduces to the
empty-sequence floor. With the remap, the source HMM (R0) recovers
non-trivial behaviour and TTA is applied on top
(Appendix~\ref{sec:exp:tta-ablation}). The remap is therefore a prerequisite,
not an ablation knob.

\subsection{Vision-conditioned HMM}
\label{app:vision-hmm}
The vision-conditioned HMM augments the source emission $B$ with a
per-prompt vision context $c$ extracted from the VLM's image-level
encoder output (last layer, mean-pooled across patches).
At distillation, we condition each ground-truth continuation on its
matching context vector and learn a row-normalized context-modulated emission:
\begin{equation}
B(h,v\mid c)=
\frac{B(h,v)\exp((Wc)_v)}
{\sum_{u\in V}B(h,u)\exp((Wc)_u)}.
\end{equation}
Here $W\in\mathbb{R}^{|V|\times d_c}$ is a rank-$64$ projection.
The normalization is over vocabulary tokens for each hidden state $h$, so
$\sum_{v\in V}B(h,v\mid c)=1$.
At decode time, $c$ is computed once per prompt from the
prompt's image and held fixed for the whole sequence.
The training objective is the same per-token NLL as the unimodal HMM
distillation \citep{zhang2024ctrlg} plus a Frobenius regulariser
$\lambda_W \|W\|_F^2$ to prevent vision drift on prompts with weak
image cues.
On VLABench (\Cref{tab:vlabench}), the supervised vision-conditioned HMM
with per-instance calibration reaches a Macro score of $38.7$, compared with
$37.1$ for the text HMM. This $1.6$-point difference combines the changes in
HMM supervision and calibration.

\subsection{Test-time adaptation: derivation, equations, and stability}
\label{app:tta-details}

\paragraph{Next-token form of the HMM loss.}
% \tm{08/26-3: remove the unnecessary pre-training analogy from the loss definition.}
Test-time adaptation addresses changes in the HMM's token distribution across domains, sessions, or plans. Its self-supervised objective is
\begin{equation}
\mathcal{L}_\mathrm{TTA}(\phi;y_{1:N}) =
-\sum_{n=0}^{N-1}\log P_\phi\!\left(y_{n+1}\mid y_{1:n}\right),
\label{eq:tta-loss}
\end{equation}
which requires no external label because the test sequence supplies its own
next-token targets through the chain rule.
Let $f_n(h;\phi) = \log P_\phi(y_{1:n},\, h_n = h)$ be the log-domain
joint forward variable; the one-step probability under the HMM is
\begin{equation}
\begin{split}
&P_\phi\!\bigl(y_{n+1} \mid y_{1:n}\bigr) \;=\;\\
&\quad
\frac{\textstyle\sum_{h,h'}\exp(f_n(h))\,A(h,h')\, B(h',y_{n+1})}
     {\textstyle\sum_{h}\exp(f_n(h))},
\end{split}
\end{equation}
which makes Eq.~\eqref{eq:tta-loss} a self-supervised objective on any
test-sequence prefix. Following \citet{feng2026inplace} we update only $B$
and freeze $(A,\mu)$. Throughout, we write
$\Phi(B)\!=\!(\mu_\mathrm{src},A_\mathrm{src},B)$ for the HMM with adapted
emission $B$ and frozen source parameters $(\mu_\mathrm{src},A_\mathrm{src})$.
The source checkpoint is BEHAVIOR-trained for the VLABench TTA factorial
and VirtualHome-trained for the separate VH-to-BEHAVIOR calibration study.

\paragraph{Tier 1 (TTDA, offline domain adaptation).}
Run once before deployment on $M$ unconstrained continuations
$\{y^{(i)}\}_{i=1}^{M}$ sampled from the base VLM in the target domain:
\begin{equation}
\begin{aligned}
&B_{\mathrm{dom}}=\arg\min_B\Bigl[\frac{1}{M}\sum_{i=1}^{M}
\mathcal L_{\mathrm{TTA}}\bigl(\Phi(B);y^{(i)}\bigr)\\
&\qquad+\lambda_{\mathrm{TTDA}}\|B-B_{\mathrm{src}}\|_F^2\Bigr].
\end{aligned}
\end{equation}
with \(B\) restricted to row-stochastic emission matrices. The Frobenius term anchors a gradient-based update to the source matrix. An emission-only Baum--Welch update instead estimates normalized expected emission counts for the sequence-likelihood objective; anchoring that estimate by interpolation is a separate operation and does not solve the penalized objective above.

\paragraph{Tier 2 (OTTA, online session adaptation).}
After prompt \(t\), let \(\widetilde B_{\mathrm{dom},t}\) denote the updated, row-normalized emission estimate from its token sequence, initialized from \(B_{\mathrm{dom},t-1}\), with \(B_{\mathrm{dom},0}=B_{\mathrm{dom}}\). The anchored update is
\begin{equation}
\begin{aligned}
B_{\mathrm{dom},t}
={}&(1-\lambda_{\mathrm O})B_{\mathrm{dom}}\\
&+\lambda_{\mathrm O}\widetilde B_{\mathrm{dom},t}.
\end{aligned}
\label{eq:tta-otta-interp}
\end{equation}
For \(0\le\lambda_{\mathrm O}\le1\), this arithmetic mixture preserves row normalization. A log-space implementation applies \(\operatorname{logaddexp}\) to the weighted log probabilities \(\log B_{\mathrm{dom}}\) and \(\log\widetilde B_{\mathrm{dom},t}\), not to the probability matrices themselves. Each update retains a fixed contribution from the domain anchor.

\paragraph{Tier 3 (TTBA, per-instance batch adaptation).}
Before generating the plan for prompt $t$, draw $M_{\mathrm{inst}}=5$
unconstrained continuations from the frozen VLM. Let $B_{\mathrm{pre},t}$ be
the emission matrix available before this prompt: the source matrix, the
TTDA-adapted matrix, or the current session matrix, depending on the enabled
tiers. One emission-only Baum--Welch iteration on the sampled continuations
produces a row-normalized estimate $\widetilde B_{\mathrm{inst},t}$.
The anchored instance matrix is
\begin{equation}
\begin{aligned}
B_{\mathrm{inst},t}={}&(1-\lambda_{\mathrm B})B_{\mathrm{pre},t}\\
&+\lambda_{\mathrm B}\widetilde B_{\mathrm{inst},t},
\qquad 0\le\lambda_{\mathrm B}\le1.
\end{aligned}
\end{equation}
Initialize the decoder with this matrix, keep it fixed throughout the plan,
and discard the instance-specific changes afterwards. Session-level OTTA,
when enabled, updates separately from the completed plan. Thus R2 and R3
evaluate pre-generation batch adaptation, with no within-plan chunk updates.

\paragraph{Decode-time cost.}
TTDA runs offline. For measured per-instance TTBA, let $L$ be the total
number of tokens in the five sampled continuations and $H$ the number of
hidden states. With dense transitions, one emission-only Baum--Welch pass
costs $\mathcal O(LH^2+H|V|)$, including emission-count initialization and
normalization. Its online cost also includes five VLM sampling calls and
decoder reinitialization. Any HMM-dependent lookahead tables must be rebuilt;
when action-level reachability is enabled, this includes $F_a$ and $R$.
OTTA instead adds one emission update after the generated plan, costing
$\mathcal O(NH^2+H|V|)$ for a plan of $N$ tokens, plus refresh of any cached
HMM-dependent quantities before subsequent decoding. These costs exclude
the base VLM's constrained generation and depend on sample and plan lengths.
Appendix~\ref{app:compute} reports the measured run-time differences and a
sampling/update profile.

\paragraph{Adaptation objective and safeguards.}
% \tm{08/26-3: scope the four properties to adaptation drift rather than safety.}
The objective and update rules have the following properties.
(i) $\mathcal{L}_\mathrm{TTA}$ is a sequence NLL; dividing by the number of
tokens gives per-token NLL. Improving the adaptation-sequence objective does
not guarantee lower held-out NLL or better planning scores.
(ii) The arithmetic mixture in Eq.~\eqref{eq:tta-otta-interp} preserves a fixed contribution from the normalized anchor distribution.
(iii) Discarding the Tier 3 instance matrix prevents its changes from carrying into the next prompt; session-level OTTA updates, when enabled, are maintained separately and reset at task-family boundaries.
(iv) The feasibility structure---the syntax and visibility DFAs,
$\mathcal{D}_{\mathrm{rch}}$, and the action budget $K$---is symbolic and never
depends on \(B\). The reachability table \(R\) also depends on the HMM's support: a symbolically executable path has finite mass only if its serialization has positive HMM probability. An adaptation step that preserves this support changes finite scores without adding or removing supported paths. The numeric table must be refreshed after \(B\) changes, while the symbolic masks and transition relation remain fixed.

\section{Experimental Protocol and Reproducibility}

This section consolidates the benchmark scope, evaluation tracks, hyperparameters, simulator assumptions, metrics, and baseline implementations used throughout the experiments.

\subsection{Benchmarks for Executable and Safe Embodied Planning}
% \tm{08/26-3: replace broad benchmark rhetoric with the evaluated capabilities.}
The Embodied Agent Interface (EAI)~\citep{li2024embodied} and
VirtualHome~\citep{Puig_2018_CVPR} test whether agents produce executable
action sequences for household tasks, while BEHAVIOR examines longer-horizon
plans in richer simulated settings. VLABench~\citep{zhang2025vlabench} adds a
visual grounding requirement, where plans must be consistent with observed scene
geometry rather than abstract object names.
SafeAgentBench~\citep{yin2024safeagentbench} shows that task completion and
safety can come apart: agents frequently finish a goal task by taking actions
that violate stated constraints. Accordingly, we report task success separately
from plan validity, grounding, executability, and safety.

\subsection{Setup details}
\label{app:setup-details}

This section expands on the abbreviated setup of \Cref{sec:exp}.

\paragraph{Models and serving.}
Text-only experiments use Llama-3.1-8B-Instruct~\citep{grattafiori2024llama3};
multimodal experiments use Qwen3-VL-8B-Instruct~\citep{qwen3vl2025},
and the second-backbone validation uses InternVL3.5-8B. Qwen3-VL is
served via vLLM with thinking mode disabled
(\texttt{enable\_thinking=False}) to prevent \texttt{<think>} tokens
from corrupting JSON / LTL parsing. The main ablations always compare
constrained and unconstrained decoding on the same backbone. No planner
component is fine-tuned. We use greedy decoding ($T{=}0$) throughout, except per-instance
TTA which draws $M{=}5$ stochastic continuations.

\paragraph{Evaluation-track dictionary.}
\Cref{tab:run-dictionary} records the pipeline boundaries used throughout the
figures and tables. All VLABench scores are reported on a $0$--$100$ scale,
but the score definitions and aggregation rules are track-specific. The
Qwen VL-main block macro-averages six reported dimension scores, whereas
the diagnostic tracks pool local component scores over prompts.
Values from different rows of the dictionary are not
direct comparisons even when the backbone and prompt count coincide.

\newcommand{\rundictionarytable}{%
\begin{table*}[t]
\centering
\small
\setlength{\tabcolsep}{6pt}
\renewcommand{\arraystretch}{1.10}
\begin{tabular*}{\textwidth}{@{\extracolsep{\fill}}
  >{\raggedright\arraybackslash}p{0.19\textwidth}
  >{\raggedright\arraybackslash}p{0.24\textwidth}
  >{\raggedright\arraybackslash}p{0.25\textwidth}
  >{\raggedright\arraybackslash}p{\dimexpr0.32\textwidth-6\tabcolsep\relax}
@{}}
\toprule
\textbf{Experiment} & \textbf{Model and evaluation set} &
\textbf{Compared configurations} & \textbf{Score aggregation} \\
\midrule
Main evaluation &
Qwen3-VL-8B\newline Multimodal input\newline $480$ prompts; six dimensions &
Baseline; syntax--visibility DFA; text HMM; supervised vision HMM &
Reported VL-main dimension scores; six-dimension macro mean (\Cref{tab:vlabench}) \\
\addlinespace[5pt]
Text-only diagnostics &
Llama-3.1-8B\newline Text-only input\newline $480$ prompts; six dimensions &
Baseline; DFA; HMM diagnostic-weight sweep &
Pooled mean of skill, entity, and exact match (\Cref{fig:vlabench-modea-dimensions}) \\
\addlinespace[5pt]
Multimodal diagnostics &
Qwen3-VL-8B\newline Multimodal input\newline $480$ prompts; six dimensions &
Baseline; DFA &
Pooled mean of skill, entity, and exact match (\Cref{tab:vlabench_modeB,fig:vlabench-modea-dimensions}) \\
\addlinespace[5pt]
Test-time adaptation &
Qwen3-VL-8B\newline Multimodal input\newline $480$ prompts ($78$ Physical Laws; $402$ other) &
Remapped source HMM; TTDA; OTTA; TTBA\newline Hugging Face generation &
Pooled mean of skill, entity, and exact match (\Cref{tab:tta-ablation}) \\
\addlinespace[5pt]
InternVL comparison &
InternVL3.5-8B\newline Two-image input\newline $474$ prompts; four dimensions &
Matched baseline and \textsc{Clamp} &
Four-dimension local total; no six-dimension mean (\Cref{fig:internvl}) \\
\bottomrule
\end{tabular*}
\caption{\textbf{VLABench experimental settings and score aggregation.}
All scores are reported on a $0$--$100$ scale.}
\label{tab:run-dictionary}
\end{table*}
}

\paragraph{HMM bridge.}
% \tm{08/26-3: integrate the appendix pointers into the setup description.}
The base conditional HMM ($H{=}128$) is self-distilled from continuations
sampled from the same frozen backbone, following \citet{zhang2024ctrlg}.
The vision-conditioned VLABench HMM is the exception: it is trained from
benchmark ground-truth skill sequences paired with their images and therefore
uses additional task supervision. The Qwen3-vocabulary HMM used for VLABench
Mode~B and TaPA is obtained by remapping the Llama-tokenised checkpoint
(Appendix~\ref{app:vocab-remap}); without the remap, every Mode~B / TaPA output
collapses to the empty-sequence floor. Appendix~\ref{app:hmm-bridge} defines
the HMM forward/backward operators, and Appendix~\ref{app:beta-dp} defines the
backward DP table $R$ used by $\mathcal{D}_{\mathrm{rch}}$. The vision-conditioned
variant appears in Appendix~\ref{app:vision-hmm}; Appendix~\ref{app:tta-details}
gives the per-instance TTA derivations and stability analysis.

\paragraph{External-baseline harness.}
Safety~Chip~\citep{yang2024safetychip} translates an NL constraint to
LTL, monitors the completed plan with a Spot DFA, and reprompts the
LLM on violation (up to three times). LLM-TAMP-prompt~\citep{wang2024llm3}
adds a CoT-then-JSON output instruction and one parse-failure retry;
on EAI it reduces to a structured CoT baseline because no
motion-feasibility checker is available. Mechanism comparison and run scripts
are in Appendix~\ref{app:baselines}. These external rows are distinct from
the matched \textsc{Ctrl-G}/\textsc{SafeDec} diagnostic, whose adaptations
are described in Appendix~\ref{app:controlled-baselines}.

\rundictionarytable

\subsection{Hyperparameters}
\label{app:hparams}
\Cref{tab:hparams} consolidates the decoding, HMM, and adaptation settings used
throughout \Cref{sec:exp}.
\begin{table*}[t]
\centering
\small
\setlength{\tabcolsep}{6pt}
\renewcommand{\arraystretch}{1.12}
\begin{tabular*}{\textwidth}{@{\extracolsep{\fill}}
  >{\raggedright\arraybackslash}p{0.23\textwidth}
  >{\raggedright\arraybackslash}p{0.27\textwidth}
  >{\raggedright\arraybackslash}p{\dimexpr0.50\textwidth-4\tabcolsep\relax}
@{}}
\toprule
\textbf{Parameter} & \textbf{Default value} & \textbf{Description} \\
\midrule
$\lambda_{\mathrm{diag}}$ &
0.1 (Action Sequencing)\newline 0.05 (Goal Interpretation) &
Finite soft-score weight in the transfer diagnostics. \\
\addlinespace[3pt]
$\lambda$ & 1.0 &
Reachability-score weight in \textsc{Clamp}-world/full. \\
\addlinespace[3pt]
\texttt{lookahead\_cap} & 100 &
Maximum action-graph depth for diagnostic backward search. \\
\addlinespace[3pt]
\texttt{max\_new\_tokens} & 768--2048 &
Generated-token cap, fixed per benchmark family. \\
\addlinespace[3pt]
$K_{\mathrm{max}}$ & 40 &
Maximum action budget for BEHAVIOR and VLABench. \\
\addlinespace[3pt]
$H$ & 128 &
Number of HMM hidden states. \\
\bottomrule
\end{tabular*}
\caption{\textbf{Decoder and HMM hyperparameters used in the experiments.}
The diagnostic weight $\lambda_{\mathrm{diag}}$ corresponds to the
\texttt{gamma\_scale} implementation parameter.}
\label{tab:hparams}
\end{table*}

Grid sweep ranges are
$\lambda_{\mathrm{diag}}\in\{10^{-3},5\!\cdot\!10^{-3},10^{-2},5\!\cdot\!10^{-2},10^{-1},3\!\cdot\!10^{-1},1\}$
(BEHAVIOR / VLABench Mode~A).
The implementation historically calls this coefficient
\texttt{gamma\_scale}; it is a finite soft-score weight and is unrelated to
the hard validity masks $\log m_{i,n}(v)\in\{0,-\infty\}$.
The measured VLABench TTA factorial uses diagnostic weight
$\lambda_{\mathrm{diag}}=1$, lookahead cap $50$, and a final-plan token cap
of $256$. TTBA draws $M_{\mathrm{inst}}=5$ unconstrained continuations per
prompt and performs one Baum--Welch emission update before decoding.
OTTA performs one emission update after each completed plan.

\subsection{Simulator assumptions and reported constraint use}

$\mathcal{D}_{\mathrm{rch}}$'s reachability assessment assumes that its
symbolic transition model is sound and deterministic. A simulator may satisfy
that assumption without a paper experiment actually evaluating the complete
$\mathcal{D}_{\mathrm{rch}}$ mechanism. \Cref{tab:simulator-determinism} therefore
separates the benchmark environment from the constraint mechanism reported in
this paper.

\begin{table*}[t]
\caption{Benchmark simulator assumptions and the mechanism actually reported.
Simulator determinism alone does not imply that full
$\mathcal{D}_{\mathrm{rch}}$ reachability was evaluated.}
\label{tab:simulator-determinism}
\centering
\small
\setlength{\tabcolsep}{4pt}
\begin{tabular*}{\textwidth}{@{\extracolsep{\fill}}p{0.16\textwidth}p{0.31\textwidth}p{0.43\textwidth}}
\toprule
\textbf{Benchmark} & \textbf{Simulator assumption} & \textbf{Reported use in this paper} \\
\midrule
EAI VirtualHome & \texttt{evolving\_graph} symbolic transitions &
Token grammar and text-only world-state reachability diagnostics; symbolic
feasibility remains conditional on transition-model soundness. \\
EAI BEHAVIOR & iGibson / OmniGibson rigid-body physics; fixed-seed evaluations &
GI/SD output-schema and soft-score diagnostics. No complete
$\mathcal{D}_{\mathrm{rch}}$ reachability result is reported. \\
VLABench & MuJoCo + dm-control; fixed benchmark episodes &
Syntax--visibility DFA and HMM diagnostics at the symbolic plan interface; no
physical-execution guarantee is claimed. \\
TaPA-60 & AI2-THOR scene metadata and task splits &
Verb-format and visibility-DFA grounding diagnostic; no full
$\mathcal{D}_{\mathrm{rch}}$ reachability evaluation. \\
SafeAgentBench & AI2-THOR task descriptions and symbolic safety rules &
Compiled temporal safety policy plus post-hoc rule checking and recovery;
completion and safety are reported separately. \\
\bottomrule
\end{tabular*}
\end{table*}

\subsection{VLABench score definitions and aggregation}
\label{app:vlabench-metric}

\paragraph{Published benchmark metric.}
VLABench defines four components~\citep{zhang2025vlabench}: skill recall
(SR), parameter recall (PR), skill--parameter recall (SPR), and precise
matching (PM). The recall metrics use ground-truth skills, parameters, or
skill--parameter pairs as their denominators. PM is the fraction of
ground-truth dependency-graph nodes matched in skill, parameters, and
dependency relations; it is not a binary whole-plan exact-match indicator.
For component values in $[0,1]$, the published weighted score is
\begin{equation}
\begin{aligned}
S_{\mathrm{official}}=100\bigl(&w_1\,\mathrm{SR}+w_2\,\mathrm{PR}\\
&+w_3\,\mathrm{SPR}+w_4\,\mathrm{PM}\bigr),
\end{aligned}
\end{equation}
where $w_j\ge0$ and $\sum_{j=1}^{4}w_j=1$. The published-reference block
in \Cref{tab:vlabench} reproduces the benchmark's reported results.

\paragraph{Local diagnostic scores.}
VL-A, VL-B, and VL-TTA use the equal-weight local diagnostic
$100(\overline s+\overline e+\overline x)/3$, where the three means are
the recorded skill-match, entity-match, and exact-match components over the
evaluated prompts. This diagnostic is distinct from the official four-component
score. Component breakdowns are given in \Cref{fig:vlabench-modea-dimensions,tab:vlabench_modeB}.
For InternVL, the per-example local total is specifically
$100(F_{1,\mathrm{skill}}+F_{1,\mathrm{entity}}+\mathrm{EM})/3$,
where $\mathrm{EM}$ indicates exact skill-and-entity sequence match.
\Cref{tab:vlabench} averages these per-example totals within each of its four
evaluated dimensions; \Cref{fig:internvl} pools all $474$ prompts.

\paragraph{Qwen VL-main aggregation.}
The Qwen block in \Cref{tab:vlabench} reports six dimension scores and their
unweighted arithmetic mean (Macro). Its baseline and constrained rows are
compared within this block. The VL-main baseline of $28.7$ and the separate
VL-B diagnostic baseline of $20.1$ come from different evaluation pipelines;
the difference cannot be attributed to macro versus pooled aggregation alone.
% TODO(metric provenance): recover the evaluator revision, per-prompt results,
% and within-dimension aggregation script for Qwen VL-main 28.7/34.1/37.1/38.7.
% The retained summary calls this a three-component local total, but the
% original evaluator has not been located. Do not assign the official
% four-component formula or undocumented length weights to these numbers.

\subsection{External baseline implementations}
\label{app:baselines}

\paragraph{Backbone and comparison boundary.}
Safety~Chip and LLM-TAMP-prompt use
\texttt{Qwen/Qwen3-VL-8B-Instruct} through the contextual external-baseline
harness. Thinking mode is disabled (\texttt{enable\_thinking=False}) and
decoding is greedy ($T{=}0$). The \textsc{Clamp}-token/world values discussed
alongside them come from the matched Llama-3.1-8B track, not from that Qwen
harness; their Task~SR values are therefore not horizontal comparisons.

\subsubsection{Safety Chip}

Safety Chip~\citep{yang2024safetychip} enforces constraints through a
\emph{reprompt loop} that operates entirely outside the decoder.
(1)~The LLM generates a complete plan in one shot.
(2)~A rule-based parser checks whether each action violates a constraint
(format, vocabulary, argument arity).
(3)~On any violation, an NL description of the failure is appended to the
prompt and the LLM is queried again.
(4)~Steps~(2)--(3) repeat up to three times; if no valid plan is produced,
the last output is kept.

The original system uses NL$\to$LTL translation and a Spot DFA
monitor~\citep{duret2022spot} to check temporal safety properties.
EAI tasks carry no explicit NL safety constraint, so we instantiate Safety
Chip with the \emph{action-format constraint}: a plan is valid iff every
action name belongs to the allowed vocabulary with the correct argument
count.
This is the same token-level condition that \textsc{Clamp}-token enforces;
the difference is that Safety Chip applies it
post-hoc through reprompting while \textsc{Clamp}-token enforces it by
masking invalid tokens to $-\infty$ before sampling.

Empirically, ${\approx}15\%$ of BEHAVIOR AS tasks exhaust all three
reprompts without converging to a valid format; those outputs are passed
to the evaluator as-is and score 0 on grammar metrics.
\textsc{Clamp}-token achieves $0\%$ grammar errors by construction on the same tasks.

% De-anonymisation risk: internal repository paths commented out for the
% double-blind submission. Restore for the camera-ready version.
% \noindent\textit{Implementation:}
% \texttt{ltl\_safety/src/exp\_eai\_vh\_as.py} (VH~AS),
% \texttt{ltl\_safety/src/exp\_eai\_behavior\_as.py} (BEHAVIOR~AS).
% Run script: \texttt{eai\_ctrlg/scripts/run\_safety\_chip\_eai.sh}.

\subsubsection{LLM-TAMP-prompt}

LLM$^3$-TAMP~\citep{wang2024llm3} is a planner that pairs a
CoT-then-JSON LLM with a motion-feasibility checker (PyBullet
collision~+~IK) that feeds categorised failure messages back to the LLM
for iterative repair.
On EAI, which provides no simulator, we run only the LLM component and
call the result \textbf{LLM-TAMP-prompt} to distinguish it from the full
system.

\paragraph{What changes.}
Each EAI prompt is augmented with a CoT output instruction:
\begin{quote}\small
\textit{Before giving your final answer, briefly reason about the task
(1--3 sentences).
Then output a JSON object:
\texttt{\{"reasoning": "...", "actions": <action sequence>\}}.
Output only the JSON object; no other text.}
\end{quote}
The response is parsed to extract the \texttt{actions} field; one retry
is attempted if parsing fails.
There is no backtracking loop: unlike the full LLM$^3$-TAMP system,
no feasibility signal is available, so the method is equivalent to a
structured CoT baseline.

\paragraph{Purpose of the baseline.}
% \tm{08/26-3: replace the informal heading and avoid generalizing one prompt result.}
LLM-TAMP-prompt provides a contextual test of structured CoT output under the
Qwen harness, without any constraint enforcement.  It reaches 51.1\%
Task~SR on VH~AS (\Cref{sec:exp:vh}).  We do not report a prompting gain:
this table contains no matched bare-Qwen row, whereas the 21.3\% bare row
belongs to the separate matched-Llama track.
% \tm{TODO-15.14: scope the prompt-engineering statement to the one tested
% prompting baseline.}
Its Task~SR is reported only as contextual evidence and is not subtracted from
the matched Llama \textsc{Clamp}-world value. The mechanisms differ in whether
they apply per-token goal-reachability lookahead ($\mathcal{D}_{\mathrm{rch}}$ backward DP), and the backbones/interfaces also differ, so this run does not rule out
other prompting methods.

% De-anonymisation risk: internal repository paths commented out for the
% double-blind submission. Restore for the camera-ready version.
% \noindent\textit{Implementation:}
% \texttt{LLM-TAMP/utils/eai\_prompt\_runner.py}.
% Run script: \texttt{eai\_ctrlg/scripts/run\_llm\_tamp\_eai.sh}.

\subsubsection{\textsc{Clamp}-token and \textsc{Clamp}-world}

% \tm{08/26-3: replace the sentence fragment with a direct mechanism contrast.}
\textsc{Clamp}-token and \textsc{Clamp}-world are described in \Cref{sec:exp}.
Both differ from the external baselines above by enforcing constraints
\emph{inside} the decoder: they set the logit of
every violating token to $-\infty$ before sampling.
No reprompting occurs; configured violations are masked before sampling rather
than merely made empirically unlikely.

\Cref{tab:baseline-mechanism} summarises the four methods.

\begin{table*}[t]
\caption{Mechanism comparison for the four EAI configurations. Task~SR is
omitted because the contextual Qwen external-baseline rows and matched Llama
rows use different backbones and interfaces. ``Hard'' denotes a structural
decoder constraint under its stated symbolic assumptions.}
\label{tab:baseline-mechanism}
\centering
\small
\setlength{\tabcolsep}{3pt}
\begin{tabular}{@{}p{0.20\textwidth}p{0.16\textwidth}p{0.26\textwidth}p{0.12\textwidth}p{0.16\textwidth}@{}}
\toprule
Method & Constraint location & Enforced condition & Reprompt & Evaluation track \\
\midrule
Safety Chip
  & Outside decoder & Empirical plan check & Up to 3$\times$ & Contextual Qwen \\
LLM-TAMP-prompt
  & Outside decoder & None beyond parsing & 1$\times$ (parse) & Contextual Qwen \\
\textsc{Clamp}-token
  & Inside decoder & Hard format/vocab/arity & Never & Matched Llama \\
\textsc{Clamp}-world
  & Inside decoder & Hard interface + symbolic reachability & Never & Matched Llama \\
\bottomrule
\end{tabular}
\end{table*}

\subsection{Controlled decoding-baseline adaptations}
\label{app:controlled-adaptations}

The controlled VirtualHome diagnostic converts the original Watch-And-Help
tasks into single-agent, text-only, open-loop planning. Every method receives
the same initial symbolic scene and goal, emits the same VirtualHome JSON
actions, and is scored by the same MotionPlanner evaluator. Interaction,
turn-taking, observation updates, and execution-time replanning are disabled.

\paragraph{\textsc{Ctrl-G}.}
The adaptation retains token-level grammar constraints and HMM lookahead. Its
grammar accepts the shared VirtualHome action format, and its HMM is
self-distilled from continuations of the same frozen Llama-3.1-8B backbone.
It does not use \textsc{Clamp}'s prompt-derived transition graph, semantic
HMM, backward goal reachability, goal value, or goal-based stopping.

\paragraph{\textsc{SafeDec}.}
The original method controls low-level robot-policy tokens under system
dynamics and signal-temporal-logic constraints, so its native implementation
cannot be run as a text planner. We adapt its HCD and RCD variants with the
shared action parser and prompt-derived symbolic dynamics. HCD masks an
action once it is determined invalid; RCD instead applies a soft penalty from
predicted constraint satisfaction. Neither variant uses \textsc{Clamp}'s
semantic HMM, backward goal planning, goal value, or goal-based stopping.

% \tm{08/26-3: state the comparison boundary without the uncommon interface verb.}
These adaptations retain each method's core decoding mechanism under one
common interface. The resulting table compares the adapted mechanisms rather
than the original end-to-end systems.

\subsection{Safety Chip backend calibration}
Before applying Safety~Chip as an external baseline on EAI VH/BEHAVIOR
or SafeAgentBench (\Cref{sec:exp:safeagentbench}), we replicate its
authors' VirtualHome demo set under our Qwen3-VL-8B backend with
thinking disabled.
\begin{table}[t]
\centering
\small
\setlength{\tabcolsep}{3pt}
\begin{tabular}{@{}l l c c c@{}}
\toprule
\textbf{Scene} & \textbf{Arm} & \textbf{Completed} & \textbf{Safe} & \textbf{Both} \\
\midrule
\texttt{mobile\_manip} & \texttt{full} & 49/50 & \textbf{50/50} & 49/50 \\
\texttt{mobile\_manip} & \texttt{bad}  & 48/50 & 31/50 & 31/50 \\
\texttt{mobile\_manip} & \texttt{null} & 47/50 & 15/50 & 15/50 \\
\midrule
\texttt{rooms}         & \texttt{full} & 49/50 & \textbf{50/50} & 49/50 \\
\texttt{rooms}         & \texttt{bad}  & 50/50 & 27/50 & 27/50 \\
\texttt{rooms}         & \texttt{null} & 50/50 & 19/50 & 19/50 \\
\bottomrule
\end{tabular}
% \tm{08/26-3: shorten the caption and avoid attributing all transfer gaps to one cause.}
\caption{\textbf{Safety~Chip backend calibration on its VirtualHome
demo set} (10 task--instruction pairs $\times$ 2 scenes $\times$ 5 trials
$\times$ 3 safety arms, $n{=}300$ rollouts; Qwen3-VL-8B with thinking
disabled). Three
\texttt{safety\_level} arms: \texttt{full} = correct LTL constraints,
\texttt{bad} = deliberately broken LTL, \texttt{null} = no LTL
(unconstrained planner). The \texttt{full} arm reaches $100\%$ safety
and $98\%$ task completion in both scenes; the \texttt{null} arm is
$30{-}38\%$ safe. Transfer results also depend on the target constraint
specification.}
\label{tab:safety_chip_calibration}
\end{table}

\Cref{tab:safety_chip_calibration} reports $100\%$ safety
in the \texttt{full} arm,
$30{-}38\%$ in the \texttt{null} (no-LTL) arm, and an intermediate
$54{-}62\%$ in the \texttt{bad} (broken-LTL) arm. This is a bounded backend
calibration, not evidence that the external method transfers unchanged to a
new constraint specification.

\paragraph{Reproduction record.}
The archived cross product contains 10 task--instruction pairs, two scenes,
five trials, and three safety arms. Per-rollout outputs are stored under
\texttt{ltl-safety/results} in the
\texttt{SueMintony/embodied-agents-results} artifact repository.

\section{Detailed Experimental Results}

The results are organized by evaluation setting: text-only embodied planning, multimodal grounding and adaptation, safety-constrained planning, transfer diagnostics, and computational cost.

\subsection{Controlled VirtualHome comparison}
\label{app:controlled-baselines}

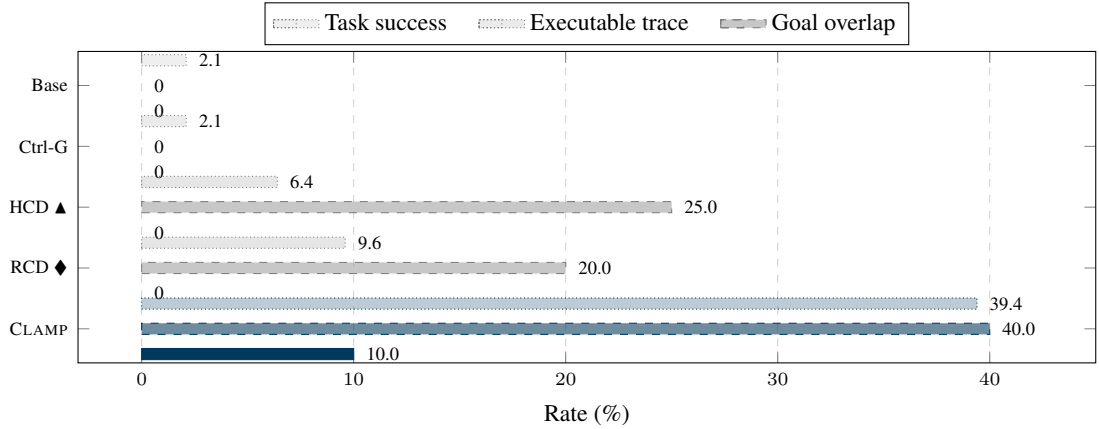
\begin{figure*}[t]
\centering
\begin{tikzpicture}
\pgfplotsset{
  taskbar/.style={xbar,bar width=4.2pt,bar shift=-9.5pt,solid},
  execbar/.style={xbar,bar width=4.2pt,bar shift=0pt,dashed,fill opacity=0.58},
  overlapbar/.style={xbar,bar width=4.2pt,bar shift=9.5pt,densely dotted,fill opacity=0.26}
}
\begin{axis}[
  width=0.94\textwidth,
  height=5.7cm,
  % Reserve a small axis-internal margin so measured zero labels remain
  % legible and visually separate from the method names.
  xmin=-3,
  xmax=45,
  xtick={0,10,20,30,40},
  xlabel={Rate (\%)},
  symbolic y coords={B,CtrlG,HCD,RCD,CLAMP},
  ytick={B,CtrlG,HCD,RCD,CLAMP},
  yticklabels={Base,Ctrl-G,{HCD $\blacktriangle$},{RCD $\blacklozenge$},{\textsc{Clamp}}},
  y dir=reverse,
  xmajorgrids,
  grid style={dashed,ReferenceGray!35},
  tick label style={font=\scriptsize},
  label style={font=\small},
  enlarge y limits=0.14,
  legend style={
    at={(0.5,1.02)},anchor=south,
    legend columns=3,font=\small,
    /tikz/every even column/.append style={column sep=0.9em}
  },
  clip=false
]
% Reference and author-adapted rows use gray shades; the two SafeDec rows
% deliberately share a shade and are distinguished redundantly by their
% direct labels and geometric symbols.
\addplot[overlapbar,draw=ReferenceGray,fill=ReferenceGray!45] coordinates {(2.1,B)};

\addplot[overlapbar,draw=ReferenceGray,fill=ReferenceGray!60] coordinates {(2.1,CtrlG)};

\addplot[execbar,draw=ReferenceGray,fill=ReferenceGray!75] coordinates {(25.0,HCD)};
\addplot[overlapbar,draw=ReferenceGray,fill=ReferenceGray!75] coordinates {(6.4,HCD)};
\addplot[execbar,draw=ReferenceGray,fill=ReferenceGray!75] coordinates {(20.0,RCD)};
\addplot[overlapbar,draw=ReferenceGray,fill=ReferenceGray!75] coordinates {(9.6,RCD)};

\addplot[taskbar,draw=ClampNavy,fill=ClampNavy] coordinates {(10.0,CLAMP)};
\addplot[execbar,draw=ClampNavy,fill=ClampNavy] coordinates {(40.0,CLAMP)};
\addplot[overlapbar,draw=ClampNavy,fill=ClampNavy] coordinates {(39.4,CLAMP)};

% Values are positioned explicitly so the labels follow the shifted bars.
\node[font=\scriptsize,anchor=west,xshift=1pt,yshift=9.5pt] at (axis cs:2.1,B) {2.1};
\node[font=\scriptsize,anchor=west,xshift=1pt,yshift=9.5pt] at (axis cs:2.1,CtrlG) {2.1};
\node[font=\scriptsize,anchor=west,xshift=1pt] at (axis cs:25.0,HCD) {25.0};
\node[font=\scriptsize,anchor=west,xshift=1pt,yshift=9.5pt] at (axis cs:6.4,HCD) {6.4};
\node[font=\scriptsize,anchor=west,xshift=1pt] at (axis cs:20.0,RCD) {20.0};
\node[font=\scriptsize,anchor=west,xshift=1pt,yshift=9.5pt] at (axis cs:9.6,RCD) {9.6};
\node[font=\scriptsize,anchor=west,xshift=1pt,yshift=-9.5pt] at (axis cs:10.0,CLAMP) {10.0};
\node[font=\scriptsize,anchor=west,xshift=1pt] at (axis cs:40.0,CLAMP) {40.0};
\node[font=\scriptsize,anchor=west,xshift=1pt,yshift=9.5pt] at (axis cs:39.4,CLAMP) {39.4};

% Explicit labels at the origin preserve every measured zero.
\node[font=\scriptsize,anchor=west,xshift=1pt,yshift=-9.5pt] at (axis cs:0,B) {0};
\node[font=\scriptsize,anchor=west,xshift=1pt] at (axis cs:0,B) {0};
\node[font=\scriptsize,anchor=west,xshift=1pt,yshift=-9.5pt] at (axis cs:0,CtrlG) {0};
\node[font=\scriptsize,anchor=west,xshift=1pt] at (axis cs:0,CtrlG) {0};
\node[font=\scriptsize,anchor=west,xshift=1pt,yshift=-9.5pt] at (axis cs:0,HCD) {0};
\node[font=\scriptsize,anchor=west,xshift=1pt,yshift=-9.5pt] at (axis cs:0,RCD) {0};

\addlegendimage{area legend,draw=ReferenceGray,fill=ReferenceGray,line width=0.5pt}
\addlegendentry{Task success}
\addlegendimage{area legend,draw=ReferenceGray,fill=ReferenceGray!58,dashed,line width=0.5pt}
\addlegendentry{Executable trace}
\addlegendimage{area legend,draw=ReferenceGray,fill=ReferenceGray!26,densely dotted,line width=0.5pt}
\addlegendentry{Goal overlap}
\end{axis}
\end{tikzpicture}
% \tm{08/26-3: state the adaptation boundary in shorter sentences.}
\caption{\textbf{Adapted common-interface VirtualHome diagnostic.} The methods
share the frozen Llama-3.1-8B backbone, prompt template, JSON action interface,
action budget, greedy policy, and symbolic evaluator; their constraint and
adaptation mechanisms differ. HCD and RCD denote author-adapted
hard-constrained and robustness-conditioned SafeDec variants. Values at the
origin are measured zeros. The retained rebuttal record does not state the
episode denominator, so these percentages support only a descriptive
diagnostic, not count-based uncertainty or significance claims. The figure
does not compare the original end-to-end systems.}
\label{fig:controlled-decoding}
\end{figure*}

The original \textsc{Ctrl-G} and \textsc{SafeDec} systems use different
interfaces, so \Cref{fig:controlled-decoding} compares author-adapted
decoders under one VirtualHome text-planning interface. All methods receive
the same scene state and goal, emit the same JSON actions, and use the same
frozen Llama-3.1-8B-Instruct backbone, prompt, greedy decoder, action budget,
and MotionPlanner evaluator. \textsc{Ctrl-G} retains its grammar and HMM
lookahead but not \textsc{Clamp}'s transition graph, backward goal planning,
goal value, or goal-based stopping. The adapted \textsc{SafeDec} variants
share a parser and symbolic dynamics; HCD masks invalid completed actions,
whereas RCD assigns a soft penalty. Neither uses \textsc{Clamp}'s semantic
HMM or backward goal value.

% \tm{TODO-9/TODO-15.7: retain explicit zero labels and shorten the caption.}

Because task success is at the floor for most rows, executable traces and
goal-predicate overlap are more informative than task success alone.
\textsc{Clamp} reaches $40.0\%$ executable traces and $39.4\%$ goal overlap;
the strongest adapted baselines reach $25.0\%$ and $9.6\%$, respectively.
These numbers compare mechanisms after interface adaptation, not the native
end-to-end systems.

\subsection{EAI VirtualHome AS / SD: full results}
\label{app:eai-vh-results}
\label{sec:exp:vh}

\begin{figure*}[t]
\centering
\begin{minipage}[t]{0.42\textwidth}
\centering
\begin{tikzpicture}
\begin{axis}[
  width=0.64\linewidth,
  height=10.2cm,
  xmin=15,
  xmax=100,
  ymin=0.45,
  ymax=15.55,
  xtick={20,40,60,80,100},
  ytick={1,...,15},
  yticklabels={
    {\textsc{Clamp}-world},
    {DFA-only / \textsc{Clamp}-token},
    {Llama-3.1-8B baseline},
    {LLM-TAMP-prompt$^\dagger$},
    {Safety Chip$^\dagger$},
    {Mixtral 8x22B MoE},
    {Mistral Large},
    {Llama-3.1-70B-Instruct},
    {Llama-3.1-8B-Instruct},
    {o1-preview},
    {o1-mini},
    {GPT-4o},
    {Gemini 1.5 Pro},
    {Claude-3.5 Sonnet},
    {Claude-3 Opus}
  },
  xlabel={Success rate (\%) $\uparrow$},
  title={\textbf{(a)} EAI VirtualHome transfer},
  title style={font=\small,yshift=-0.15em},
  tick label style={font=\tiny},
  y tick label style={font=\tiny,align=right},
  label style={font=\small},
  axis y line*=left,
  axis x line*=bottom,
  xmajorgrids,
  grid style={dashed,ReferenceGray!30},
  clip=false,
  legend style={
    at={(0.50,-0.12)},
    anchor=north,
    legend columns=2,
    draw=none,
    font=\tiny,
    /tikz/every even column/.append style={column sep=0.45em}
  }
]
% Light bands separate the contextual, author-adapted, and matched blocks.
\fill[AccentAmber!7] (axis cs:15,3.5) rectangle (axis cs:100,5.5);
\fill[ClampNavy!5] (axis cs:15,0.5) rectangle (axis cs:100,3.5);
\draw[ReferenceGray,densely dotted] (axis cs:15,5.5) -- (axis cs:100,5.5);
\draw[ReferenceGray,densely dotted] (axis cs:15,3.5) -- (axis cs:100,3.5);

% Published/reference models: contextual series only.
\addplot[only marks,color=ReferenceGray,mark=*,mark size=1.55pt]
 coordinates {(63.3,5.82) (78.4,6.82) (59.0,7.82) (21.3,8.82) (65.2,9.82)
 (71.5,10.82) (71.5,11.82) (76.7,12.82) (76.1,13.82) (64.6,14.82)};
\addlegendentry{AS Task}
\addplot[only marks,color=ReferenceGray,mark=square*,mark size=1.45pt]
 coordinates {(67.9,5.94) (84.6,6.94) (66.6,7.94) (23.6,8.94) (72.5,9.94)
 (76.4,10.94) (81.3,11.94) (83.6,12.94) (81.3,13.94) (69.5,14.94)};
\addlegendentry{AS Exec.}
\addplot[only marks,color=ReferenceGray,mark=triangle*,mark size=1.75pt]
 coordinates {(80.5,6.06) (84.3,7.06) (78.4,8.06) (48.8,9.06) (89.4,10.06)
 (79.3,11.06) (87.6,12.06) (87.0,13.06) (89.1,14.06) (86.7,15.06)};
\addlegendentry{SD Task}
\addplot[only marks,color=ReferenceGray,mark=diamond*,mark size=1.65pt]
 coordinates {(90.2,6.18) (92.0,7.18) (87.3,8.18) (58.0,9.18) (93.2,10.18)
 (84.6,11.18) (91.1,12.18) (91.1,13.18) (92.0,14.18) (89.9,15.18)};
\addlegendentry{SD Exec.}

% Author-adapted external baselines (unmatched; AS values only).
\addplot[only marks,color=AccentAmber,mark=*,mark size=1.75pt]
 coordinates {(51.1,3.82) (44.3,4.82)};
\addplot[only marks,color=AccentAmber,mark=square*,mark size=1.65pt]
 coordinates {(58.7,3.94) (54.1,4.94)};

% Matched Llama-3.1-8B rows.
\addplot[only marks,color=BaselineBlue,mark=*,mark size=1.75pt] coordinates {(21.3,2.82)};
\addplot[only marks,color=BaselineBlue,mark=square*,mark size=1.65pt] coordinates {(23.6,2.94)};
\addplot[only marks,color=BaselineBlue,mark=triangle*,mark size=1.95pt] coordinates {(48.8,3.06)};
\addplot[only marks,color=BaselineBlue,mark=diamond*,mark size=1.85pt] coordinates {(58.0,3.18)};
\addplot[only marks,color=DFATeal,mark=*,mark size=1.75pt] coordinates {(48.7,1.82)};
\addplot[only marks,color=DFATeal,mark=square*,mark size=1.65pt] coordinates {(51.5,1.94)};
\addplot[only marks,color=DFATeal,mark=triangle*,mark size=1.95pt] coordinates {(60.1,2.06)};
\addplot[only marks,color=DFATeal,mark=diamond*,mark size=1.85pt] coordinates {(79.0,2.18)};
\addplot[only marks,color=ClampNavy,mark=*,mark size=1.75pt] coordinates {(85.3,0.82)};
\addplot[only marks,color=ClampNavy,mark=square*,mark size=1.65pt] coordinates {(91.8,0.94)};
\addplot[only marks,color=ClampNavy,mark=triangle*,mark size=1.95pt] coordinates {(79.8,1.06)};
\addplot[only marks,color=ClampNavy,mark=diamond*,mark size=1.85pt] coordinates {(82.5,1.18)};
\end{axis}
\end{tikzpicture}
\end{minipage}\hfill
\begin{minipage}[t]{0.26\textwidth}
\centering
\begin{tikzpicture}
\begin{axis}[
  width=0.68\linewidth,
  height=10.2cm,
  xmin=0,
  xmax=0.34,
  ymin=0.45,
  ymax=6.55,
  xtick={0,0.1,0.2,0.3},
  ytick={1,...,6},
  yticklabels={$\lambda_{\rm diag}{=}1.00$,$\lambda_{\rm diag}{=}0.30$,$\lambda_{\rm diag}{=}0.10$,$\lambda_{\rm diag}{=}0.05$,DFA-only,Baseline},
  xlabel={F$_1$ (proportion) $\uparrow$},
  title={\textbf{(b)} BEHAVIOR GI},
  title style={font=\small,yshift=-0.15em},
  tick label style={font=\tiny},
  y tick label style={font=\tiny},
  label style={font=\small},
  axis y line*=left,
  axis x line*=bottom,
  xmajorgrids,
  grid style={dashed,ReferenceGray!30},
  clip=false,
  legend style={at={(0.50,-0.12)},anchor=north,legend columns=2,draw=none,font=\tiny,
    /tikz/every even column/.append style={column sep=0.35em}}
]
\addplot[forget plot,only marks,color=ClampNavy,mark=*,mark size=1.65pt]
 coordinates {(0.183,0.84) (0.198,1.84) (0.225,2.84) (0.241,3.84)};
\addplot[forget plot,only marks,color=DFATeal,mark=*,mark size=1.65pt] coordinates {(0.197,4.84)};
\addplot[forget plot,only marks,color=BaselineBlue,mark=*,mark size=1.65pt] coordinates {(0.281,5.84)};
\addplot[forget plot,only marks,color=ClampNavy,mark=square*,mark size=1.55pt]
 coordinates {(0.038,1.00) (0.082,2.00) (0.121,3.00) (0.140,4.00)};
\addplot[forget plot,only marks,color=DFATeal,mark=square*,mark size=1.55pt] coordinates {(0.099,5.00)};
\addplot[forget plot,only marks,color=BaselineBlue,mark=square*,mark size=1.55pt] coordinates {(0.224,6.00)};
\addplot[forget plot,only marks,color=ClampNavy,mark=diamond*,mark size=1.75pt]
 coordinates {(0.111,1.16) (0.140,2.16) (0.173,3.16) (0.1905,4.16)};
\addplot[forget plot,only marks,color=DFATeal,mark=diamond*,mark size=1.75pt] coordinates {(0.1481,5.16)};
\addplot[forget plot,only marks,color=BaselineBlue,mark=diamond*,mark size=1.75pt] coordinates {(0.2524,6.16)};
% Metric shapes are independent of configuration colors.
\addlegendimage{only marks,color=ReferenceGray,mark=*,mark size=1.65pt}
\addlegendentry{Node}
\addlegendimage{only marks,color=ReferenceGray,mark=square*,mark size=1.55pt}
\addlegendentry{Edge}
\addlegendimage{only marks,color=ReferenceGray,mark=diamond*,mark size=1.75pt}
\addlegendentry{Aggregate}
\end{axis}
\end{tikzpicture}
\end{minipage}\hfill
\begin{minipage}[t]{0.26\textwidth}
\centering
\begin{tikzpicture}
\begin{axis}[
  width=0.68\linewidth,
  height=10.2cm,
  xmin=0,
  xmax=0.35,
  ymin=0.45,
  ymax=6.55,
  xtick={0,0.1,0.2,0.3},
  ytick={1,...,6},
  yticklabels={$\lambda_{\rm diag}{=}5\!\cdot\!10^{-2}$,$\lambda_{\rm diag}{=}10^{-2}$,$\lambda_{\rm diag}{=}5\!\cdot\!10^{-3}$,$\lambda_{\rm diag}{=}10^{-3}$,DFA-only,Baseline},
  xlabel={Rate / error (proportion)},
  title={\textbf{(c)} BEHAVIOR SD},
  title style={font=\small,yshift=-0.15em},
  tick label style={font=\tiny},
  y tick label style={font=\tiny},
  label style={font=\small},
  axis y line*=left,
  axis x line*=bottom,
  xmajorgrids,
  grid style={dashed,ReferenceGray!30},
  clip=false,
  legend style={at={(0.50,-0.12)},anchor=north,legend columns=2,draw=none,font=\tiny,
    /tikz/every even column/.append style={column sep=0.35em}}
]
\addplot[forget plot,only marks,color=ClampNavy,mark=*,mark size=1.65pt]
 coordinates {(0.06,0.84) (0.06,1.84) (0.06,2.84) (0.06,3.84)};
\addplot[forget plot,only marks,color=DFATeal,mark=*,mark size=1.65pt] coordinates {(0.10,4.84)};
\addplot[forget plot,only marks,color=BaselineBlue,mark=*,mark size=1.65pt] coordinates {(0.21,5.84)};
\addplot[forget plot,only marks,color=ClampNavy,mark=square*,mark size=1.55pt]
 coordinates {(0.20,1.00) (0.20,2.00) (0.19,3.00) (0.19,4.00)};
\addplot[forget plot,only marks,color=DFATeal,mark=square*,mark size=1.55pt] coordinates {(0.27,5.00)};
\addplot[forget plot,only marks,color=BaselineBlue,mark=square*,mark size=1.55pt] coordinates {(0.32,6.00)};
\addplot[forget plot,only marks,color=ClampNavy,mark=diamond*,mark size=1.75pt]
 coordinates {(0.22,1.16) (0.21,2.16) (0.21,3.16) (0.21,4.16)};
\addplot[forget plot,only marks,color=DFATeal,mark=diamond*,mark size=1.75pt] coordinates {(0.14,5.16)};
\addplot[forget plot,only marks,color=BaselineBlue,mark=diamond*,mark size=1.75pt] coordinates {(0.18,6.16)};
% Metric shapes are independent of configuration colors.
\addlegendimage{only marks,color=ReferenceGray,mark=*,mark size=1.65pt}
\addlegendentry{Task SR $\uparrow$}
\addlegendimage{only marks,color=ReferenceGray,mark=square*,mark size=1.55pt}
\addlegendentry{PredArg $\downarrow$}
\addlegendimage{only marks,color=ReferenceGray,mark=diamond*,mark size=1.75pt}
\addlegendentry{Halluc. $\downarrow$}
\end{axis}
\end{tikzpicture}
\end{minipage}
% \tm{08/26-3: clarify which rows are matched without caption shorthand.}
\caption{\textbf{Transfer results in native benchmark units.}
\textbf{(a)} EAI VirtualHome ($n{=}338$ tasks): Action Sequencing and Subgoal
Decomposition Task/Execution success. The navy band is the matched Llama track;
amber $\dagger$ rows are contextual Qwen3-VL adapters and are not matched to it;
gray rows are published context. \textbf{(b)} BEHAVIOR Goal Interpretation
($n{=}100$ tasks), reporting F$_1$ ($\uparrow$). \textbf{(c)} BEHAVIOR Subgoal
Decomposition ($n{=}100$ tasks), reporting Task SR ($\uparrow$), parsed-argument
error, and hallucination ($\downarrow$). Panels (b)--(c) use matched Llama rows: circles, squares, and diamonds
identify metrics; light blue, teal, and navy identify the baseline, DFA-only,
and DFA+HMM configurations, respectively.
$\lambda_{\rm diag}$ is the finite legacy soft-score weight, not the binary validity masks.}
\label{fig:transfer-results}
\end{figure*}
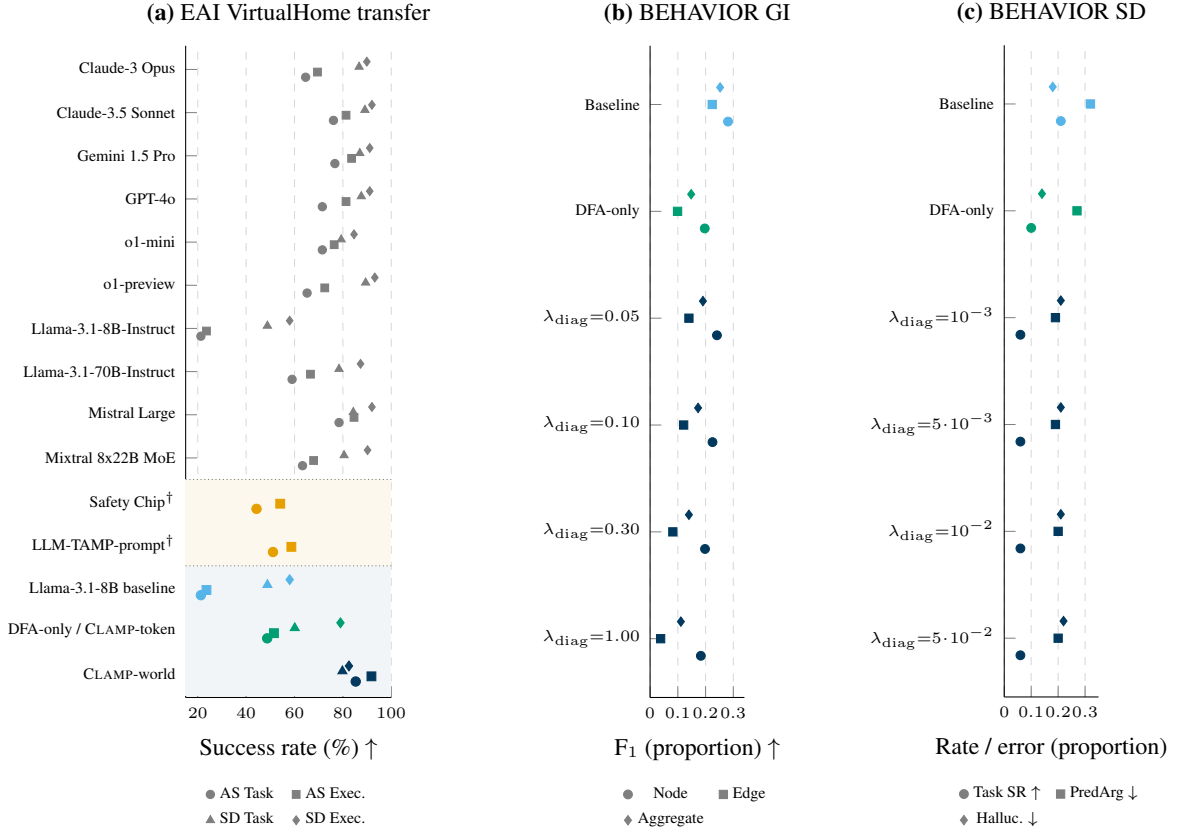

\Cref{fig:transfer-results} places the matched VirtualHome results beside the
BEHAVIOR interface diagnostics and contextual external-system rows.
% \tm{TODO-9: move colors, metrics, hyperparameters, and item-count analysis
% from the caption to the surrounding prose.}

The transfer evidence separates matched decoder effects from contextual
comparisons. Within the matched
Llama-3.1-8B-Instruct VirtualHome block, the DFA token mask raises Action
Sequencing Task SR from $21.3\%$ to $48.7\%$, and adding backward reachability
reaches $85.3\%$; Subgoal Decomposition follows
$48.8\%\!\to\!60.1\%\!\to\!79.8\%$. Flat reachability keeps execution rates
high but often selects goal-irrelevant actions, whereas the backward DP shifts
token scores toward continuations that can reach the goal within the remaining
budget. This distinction accounts for most of the Action Sequencing gain.

The published VirtualHome models and author-adapted
Safety~Chip/LLM-TAMP-prompt rows are shown only as context because their models
or interfaces are unmatched. The stricter matched comparison with
\textsc{Ctrl-G} and \textsc{SafeDec} appears in
\Cref{fig:controlled-decoding}.

\subsection{BEHAVIOR interface diagnostics}

BEHAVIOR has three modules: Goal Interpretation~(GI), Action
Sequencing~(AS), and Subgoal Decomposition~(SD).
Panels~(b)--(c) of \Cref{fig:transfer-results} report its GI and SD results
at the token-interface diagnostic stage. Goal Interpretation reports node,
edge, and aggregate F$_1$, all higher-is-better; the best \textsc{Clamp} grid
point uses $\lambda_{\mathrm{diag}}{=}0.05$ and
$\operatorname{cap}{=}100$. The affordance DFA reduces aggregate F$_1$ from
$0.2524$ (unconstrained) to $0.1481$ (DFA-only) because it assigns near-zero
mass to relational predicates such as \texttt{INSIDE(x,y)}. The HMM bridge
partially recovers aggregate F$_1$ to $0.1905$ but cannot restore tokens that
the DFA has fully masked.

Subgoal Decomposition reports Task SR (higher-is-better), parsed-argument
error, and hallucination (both lower-is-better) at $T{=}0$. The token mask
reduces parsed-argument error but halves Task SR from $0.21$ to $0.10$; the HMM
then fragments one structured item into approximately $90$ single-item
subgoals, increasing hallucination. On AS, \textsc{Clamp}-token improves Task
SR slightly ($12\%\!\to\!13\%$) while eliminating parse errors. These results
reflect interface misspecification rather than a reachability result: we do
not evaluate the full $\mathcal{D}_{\mathrm{rch}}$ on BEHAVIOR, and zero parse
errors do not establish that the supplied symbolic interface is correct.

\subsection{VLABench Mode~A: in-domain HMM weight sweep}
\label{app:modeA}
\begin{figure*}[t]
\centering
\begin{minipage}[t]{0.50\textwidth}
\centering
\begin{tikzpicture}
\begin{axis}[
  width=0.80\linewidth,
  height=7.0cm,
  xmin=-0.35,
  xmax=8.35,
  ymin=0,
  ymax=105,
  xtick={0,1,2,3,4,5,6,7,8},
  xticklabels={Base,DFA,VH{=}BEH,.01,.05,.10,.30,.50,1.0},
  x tick label style={rotate=35,anchor=east,font=\scriptsize},
  ytick={0,20,40,60,80,100},
  ylabel={Diagnostic score (0--100)},
  title={\textbf{(a)} VL-A: Llama controls and HMM-weight sweep},
  title style={font=\small,yshift=-0.2em},
  tick label style={font=\scriptsize},
  label style={font=\small},
  ymajorgrids,
  grid style={dashed,ReferenceGray!35},
  clip=false,
  every axis plot/.append style={forget plot},
  legend style={
    at={(0.50,-0.29)},
    anchor=north,
    legend columns=3,
    draw=none,
    font=\tiny,
    /tikz/every even column/.append style={column sep=0.45em}
  }
]
\draw[ReferenceGray,densely dotted] (axis cs:2.5,0) -- (axis cs:2.5,100);
\node[font=\scriptsize,ReferenceGray] at (axis cs:1.0,101.5) {controls};
\node[font=\scriptsize,ClampNavy] at (axis cs:5.5,101.5) {HMM-VLABench};

% Baseline controls: the five marker shapes retain the metric encoding.
\addplot[only marks,color=BaselineBlue,mark=*,mark size=2.0pt]
  coordinates {(0,16.9)};
\addplot[only marks,color=BaselineBlue,mark=triangle*,mark size=2.2pt]
  coordinates {(0,45.9)};
\addplot[only marks,color=BaselineBlue,mark=square*,mark size=1.9pt]
  coordinates {(0,4.8)};
\addplot[only marks,color=BaselineBlue,mark=diamond*,mark size=2.1pt]
  coordinates {(0,0.0)};
\addplot[only marks,color=BaselineBlue,mark=x,mark size=2.5pt,line width=0.9pt]
  coordinates {(0,96.7)};

% DFA-only controls.
\addplot[only marks,color=DFATeal,mark=*,mark size=2.0pt]
  coordinates {(1,29.6)};
\addplot[only marks,color=DFATeal,mark=triangle*,mark size=2.2pt]
  coordinates {(1,49.4)};
\addplot[only marks,color=DFATeal,mark=square*,mark size=1.9pt]
  coordinates {(1,34.3)};
\addplot[only marks,color=DFATeal,mark=diamond*,mark size=2.1pt]
  coordinates {(1,5.2)};
\addplot[only marks,color=DFATeal,mark=x,mark size=2.5pt,line width=0.9pt]
  coordinates {(1,93.3)};

% The HMM-VH and HMM-BEH rows are numerically identical and plotted once.
\addplot[only marks,color=AccentAmber,mark=*,mark size=2.0pt]
  coordinates {(2,28.3)};
\addplot[only marks,color=AccentAmber,mark=triangle*,mark size=2.2pt]
  coordinates {(2,46.5)};
\addplot[only marks,color=AccentAmber,mark=square*,mark size=1.9pt]
  coordinates {(2,33.0)};
\addplot[only marks,color=AccentAmber,mark=diamond*,mark size=2.1pt]
  coordinates {(2,5.4)};
\addplot[only marks,color=AccentAmber,mark=x,mark size=2.5pt,line width=0.9pt]
  coordinates {(2,97.3)};

% In-domain HMM sweep.
\addplot[ClampNavy,solid,line width=1.0pt,mark=*,mark size=1.8pt]
  coordinates {(3,27.7) (4,26.5) (5,24.8) (6,19.6) (7,12.6) (8,11.9)};
\addplot[ClampNavy,dashed,line width=1.0pt,mark=triangle*,mark size=2.0pt]
  coordinates {(3,45.6) (4,43.6) (5,40.9) (6,32.9) (7,22.9) (8,21.7)};
\addplot[ClampNavy,densely dashed,line width=1.0pt,mark=square*,mark size=1.7pt]
  coordinates {(3,32.4) (4,31.1) (5,28.9) (6,22.5) (7,14.9) (8,14.1)};
\addplot[ClampNavy,dotted,line width=1.0pt,mark=diamond*,mark size=1.9pt]
  coordinates {(3,5.2) (4,4.8) (5,4.6) (6,3.5) (7,0.0) (8,0.0)};
\addplot[ClampNavy,dashdotted,line width=1.0pt,mark=x,mark size=2.2pt]
  coordinates {(3,96.3) (4,97.3) (5,100.0) (6,91.0) (7,83.3) (8,79.4)};

\addlegendimage{only marks,color=BaselineBlue,mark=*}
\addlegendentry{Baseline}
\addlegendimage{only marks,color=DFATeal,mark=*}
\addlegendentry{DFA-only}
\addlegendimage{only marks,color=AccentAmber,mark=*}
\addlegendentry{VH $=$ BEH}
\addlegendimage{ClampNavy,solid,mark=*}
\addlegendentry{Total}
\addlegendimage{ClampNavy,dashed,mark=triangle*}
\addlegendentry{Skill}
\addlegendimage{ClampNavy,densely dashed,mark=square*}
\addlegendentry{Entity}
\addlegendimage{ClampNavy,dotted,mark=diamond*}
\addlegendentry{Exact}
\addlegendimage{ClampNavy,dashdotted,mark=x}
\addlegendentry{Format-valid}
\end{axis}
\end{tikzpicture}
\end{minipage}\hfill
\begin{minipage}[t]{0.45\textwidth}
\centering
\begin{tikzpicture}
\begin{axis}[
  width=0.66\linewidth,
  height=7.0cm,
  title={\textbf{(b)} VL-A / VL-B per-dimension diagnostics},
  title style={font=\small,yshift=-0.2em},
  enlargelimits=false,
  axis on top,
  xmin=-0.5,
  xmax=5.5,
  ymin=-0.5,
  ymax=4.5,
  y dir=reverse,
  xtick={0,1,2,3,4,5},
  xticklabels={Common Sense,Complex,M\&T,Physical Laws,Semantic,Spatial},
  x tick label style={rotate=40,anchor=east,font=\scriptsize},
  ytick={0,1,2,3,4},
  yticklabels={Llama base,Llama DFA,Llama DFA+HMM (.01),Qwen3 base,Qwen3 DFA},
  y tick label style={font=\scriptsize},
  colormap={clampmap}{color(0cm)=(white); color(1cm)=(BaselineBlue)},
  point meta min=0,
  point meta max=40,
  nodes near coords={\pgfmathprintnumber[fixed,precision=1]{\pgfplotspointmeta}},
  every node near coord/.append style={font=\tiny,text=black},
  colorbar horizontal,
  colorbar style={
    at={(0.5,-0.34)},
    anchor=north,
    width=0.66\linewidth,
    height=0.16cm,
    xtick={0,10,20,30,40},
    xticklabel style={font=\tiny},
    xlabel={Local total (0--100)},
    xlabel style={font=\scriptsize,yshift=0.2em}
  }
]
\addplot[matrix plot*,mesh/cols=6,point meta=explicit] coordinates {
  (0,0) [18.9] (1,0) [19.7] (2,0) [21.9] (3,0) [5.1] (4,0) [16.6] (5,0) [18.8]
  (0,1) [34.7] (1,1) [32.7] (2,1) [38.6] (3,1) [10.8] (4,1) [30.7] (5,1) [29.7]
  (0,2) [35.3] (1,2) [30.2] (2,2) [36.9] (3,2) [7.1] (4,2) [28.3] (5,2) [27.8]
  (0,3) [22.7] (1,3) [24.3] (2,3) [22.6] (3,3) [3.4] (4,3) [21.4] (5,3) [25.6]
  (0,4) [26.9] (1,4) [29.2] (2,4) [37.9] (3,4) [0.0] (4,4) [29.2] (5,4) [26.9]
};
\draw[white,line width=2.4pt] (axis cs:-0.5,2.5) -- (axis cs:5.5,2.5);
\draw[ReferenceGray,line width=0.5pt] (axis cs:-0.5,2.5) -- (axis cs:5.5,2.5);
\end{axis}
\end{tikzpicture}
\end{minipage}
\caption{\textbf{Pipeline-specific VLABench diagnostics (all scores $0$--$100$; higher is better).}
\textbf{(a)} VL-A uses Llama-3.1-8B with text-only prompts ($n{=}480$);
\texttt{VH} and \texttt{BEH} overlap, and .01--1.0 are
$\lambda_{\mathrm{diag}}$ values for the in-domain HMM.
\textbf{(b)} VL-A rows are separated from the Qwen3-VL multimodal VL-B rows;
the Llama HMM row uses $\lambda_{\mathrm{diag}}{=}0.01$, the highest pooled
Total among the measured in-domain HMM settings in panel~(a). These diagnostic
totals use pooled component means and are not the VL-main macro scores in
\Cref{tab:vlabench}.}
\label{fig:vlabench-modea-dimensions}
\end{figure*}
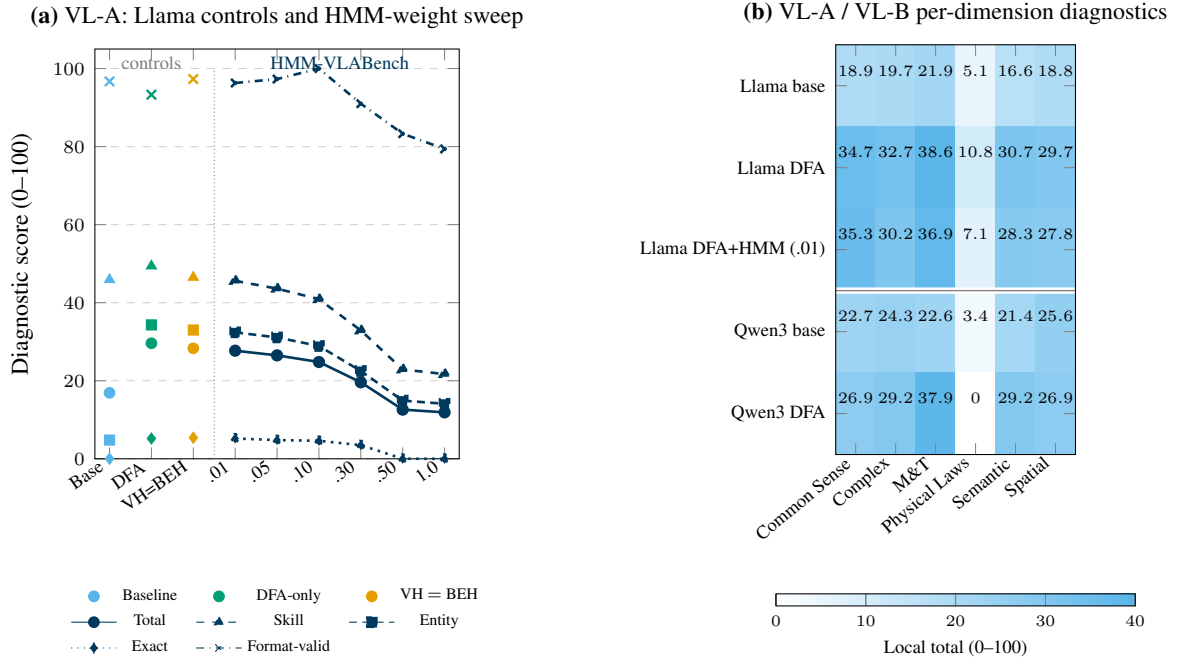

\Cref{fig:vlabench-modea-dimensions} compares the complete HMM-weight sweep and
the corresponding dimension-level behavior across the two evaluated backbones.
% \tm{TODO-9: shorten the caption and move findings to the appendix prose.}
Panel~(a) of \Cref{fig:vlabench-modea-dimensions} reports the full
$\lambda_{\mathrm{diag}}$ sweep
on the in-domain HMM (\texttt{hmm-vlabench-h128}) using the Llama-3.1-8B-Instruct
backbone on the $480$-prompt VLABench Mode~A subset (no vision).
DFA-only is the strongest configuration ($29.6$ Total).
Cross-domain HMMs (\texttt{HMM-VH}, \texttt{HMM-BEH}) are numerically
identical because both checkpoints place near-uniform mass on VLABench
skill vocabulary. The in-domain \texttt{HMM-VLABench} grid shows the
monotone collapse described in \Cref{sec:exp:diag} D1: at
$\lambda_{\mathrm{diag}}{=}1.0$ Total drops to $11.9$, $16.4$ points \emph{below} the
cross-domain HMMs.

Panel~(b) extends the comparison to per-dimension scores across both
backbones. The displayed HMM point is the measured configuration with the
highest pooled Total within each backbone-specific grid; selection is
performed separately for VL-A and VL-B. The DFA gain appears on six of six
dimensions in Mode~A and five of six in Mode~B; Physical Laws collapses to
$0.0$ under Qwen3-VL DFA-only because that
dimension is QA, not skill sequencing. Per-example analysis further shows that
the in-domain HMM helps $56$ prompts and hurts $129$. Its learned marginal
sequence length (approximately $44$ tokens) does not distinguish short
Physical~Laws plans (approximately $25$ tokens) from Complex plans
(approximately $160$ tokens), explaining the monotonic degradation as the HMM
weight increases.

\subsection{VLABench Mode~B: cross-backbone dimension analysis}

\Cref{tab:vlabench_modeB} isolates the Qwen3-VL dimension scores used to
interpret the Mode~A/Mode~B gap.
\begin{table}[t]
\centering
\small
\setlength{\tabcolsep}{3.5pt}
\begin{tabular}{@{}lccccc@{}}
\toprule
\textbf{Method} & Total & Skill & Entity & Exact & Fmt. \\
\midrule
Baseline                  & 20.1 & \textbf{59.8} & 0.5 & 0.0 & \textbf{100.0} \\
$+$ hard $\mathcal{D}_\gamma$ & \textbf{25.1} & 40.6 & \textbf{32.7} & \textbf{2.1} & \textbf{100.0} \\
\bottomrule
\end{tabular}
% \tm{08/26-3: remove a redundant closing sentence from the diagnostic caption.}
\caption{\textbf{VL-B diagnostic: Qwen3-VL-8B-Instruct ($n{=}480$).}
Scores are pooled component percentages ($0$--$100$); Total is the mean of
Skill, Entity, and Exact. The hard DFA yields $4{,}200/4{,}200$ valid format
checks and a $+5.0$-point Total. These values are not the dimension-macro
VL-main scores in \Cref{tab:vlabench}. Bold marks the better value within
these two matched rows; ties are both bold.}
\label{tab:vlabench_modeB}
\end{table}

Physical~Laws is the weakest dimension in both modes: the Mode~A DFA-only
ceiling is $10.8$, while Mode~B DFA-only reaches $0.0$
(\Cref{tab:vlabench_modeB}). These $80$ prompts constitute $16.7\%$ of the
$480$-prompt evaluation but account for approximately $40\%$ of the Mode~A to
Mode~B total gap. Their target is a single \texttt{press[N]} action, whereas
both backbones frequently interpret the scene as a manipulation task and emit
\texttt{pick}-based plans. R3's collapse onto \texttt{press} therefore helps
Physical~Laws, while R0 and R3 both score zero on the dimensions that require
entity-grounded \texttt{pick} actions.

\subsection{Matched InternVL3.5 validation}
\label{app:internvl}

% \input{sections/F04_internvl_enforcement}

% \tm{08/26-3: tighten the matched-subset scope statement.}
The second-backbone experiment uses $474$ prompts from Mesh \& Texture,
Spatial, Common Sense, and Semantic. The baseline and \textsc{Clamp} rows
share the same frozen InternVL3.5-8B revision, two-image prompts, annotations,
action schema, entity vocabulary, and evaluator. The constrained row adds a
tokenizer DFA and an HMM trained on model-sampled continuations with
$\lambda_{\mathrm{diag}}=0.1$. Physical Laws and Complex were not evaluated.
We report a four-dimensional local total
and do not compare it numerically with the six-dimension
Qwen3-VL Macro score.

Format validity remains $100.0\!\to\!100.0$ ($\Delta{=}0.0$ points, 95\% CI
[$0.0$, $0.0$]), while skill F1 changes only from
$72.0\!\to\!74.3$ ($\Delta{=}+2.3$ points, 95\% CI [$+1$, $+3$]).
The larger changes are concentrated in entity-schema compliance: entity-ID
validity rises from $1.9\!\to\!100.0$ ($\Delta{=}+98.1$ points, 95\% CI
[$+97$, $+99$]), entity F1 from $0.9\!\to\!55.7$
($\Delta{=}+54.8$ points, 95\% CI [$+52$, $+58$]), exact match from
$0.0\!\to\!25.1$ ($\Delta{=}+25.1$ points, 95\% CI [$+21$, $+29$]), and
the four-dimensional local total from $24.3\!\to\!51.7$
($\Delta{=}+27.4$ points, 95\% CI [$+25$, $+30$]).

% \tm{08/30: keep the complete versioned trace beside its matched InternVL
% analysis; the main text contains only the interpretive lead-in.}
\begin{figure}[tp]
\centering
\includegraphics[width=\columnwidth]{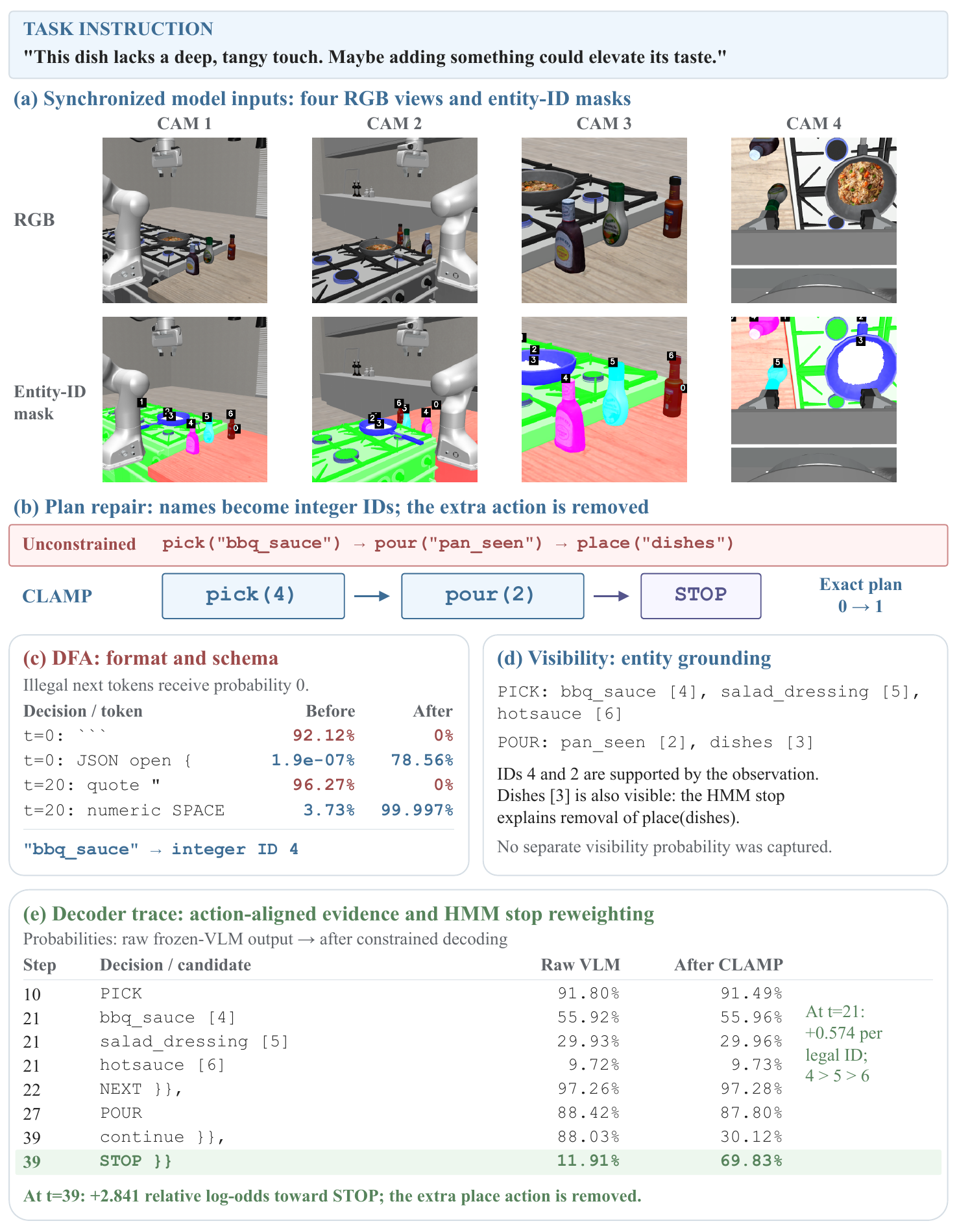}
\input{figures/case_study/caption.tex}
\end{figure}

% \tm{08/26-3: replace formulaic evidence synthesis with the measured locus of change.}
The matched-subset InternVL block in \Cref{tab:vlabench} shows gains in every
evaluated dimension. \Cref{fig:internvl} locates the larger changes in entity
and schema compliance: decode-time constraints make frozen InternVL's
predictions conform to the supplied action and entity interface. These results
do not show that \textsc{Clamp} adds visual knowledge.

\subsection{TTA tier ablation: full results}
\label{app:tta-results}
\label{sec:exp:tta-ablation}
\label{sec:exp:ablations}
\label{sec:exp:diag}

\Cref{tab:tta-ablation} compares offline domain adaptation (TTDA), online
session adaptation (OTTA), and per-instance batch adaptation (TTBA) under
the same generation and evaluation pipeline.
\begin{table}[t]
\centering
\small
\setlength{\tabcolsep}{4pt}
\begin{tabular}{lrr}
\toprule
\textbf{Configuration} & \textbf{Overall}$\uparrow$ & \textbf{PL}$\uparrow$ \\
 & $n{=}480$ & $n{=}78$ \\
\midrule
R0: source HMM          & 0.0 & 0.0 \\
R1: TTDA                & 1.0 & 6.4 \\
R2: TTBA                & 5.8 & 35.9 \\
R3: TTDA + TTBA         & \textbf{9.6} & \textbf{59.0} \\
R4: TTDA + OTTA         & 1.0 & 6.4 \\
\bottomrule
\end{tabular}
\caption{\textbf{VLABench TTA ablation with frozen Qwen3-VL-8B-Instruct.}
All configurations share the same generation and evaluation pipeline.
Scores use the pooled VL-TTA evaluator ($0$--$100$):
$100(\mathrm{skill}+\mathrm{entity}+\mathrm{exact})/3$.
Overall pools all $480$ prompts; PL denotes the $78$ Physical Laws prompts.
Bold marks column maxima.}
\label{tab:tta-ablation}
\end{table}

% \begin{table}[t]
% \centering
% \small
% \setlength{\tabcolsep}{4pt}
% \begin{tabular*}{\columnwidth}{@{\extracolsep{\fill}}lrr@{}}
% \toprule
% \textbf{Configuration} & \textbf{Overall}$\uparrow$ & \textbf{PL}$\uparrow$ \\
%  & $n{=}480$ & $n{=}78$ \\
% \midrule
% R0: source HMM          & 0.0 & 0.0 \\
% R1: TTDA                & 1.0 & 6.4 \\
% R2: TTBA                & 5.8 & 35.9 \\
% R3: TTDA + TTBA         & \textbf{9.6} & \textbf{59.0} \\
% R4: TTDA + OTTA         & 1.0 & 6.4 \\
% \bottomrule
% \end{tabular*}
% \caption{\textbf{VLABench TTA ablation with frozen Qwen3-VL-8B-Instruct.}
% All configurations share the same generation and evaluation pipeline.
% Scores use the pooled VL-TTA evaluator ($0$--$100$):
% $100(\mathrm{skill}+\mathrm{entity}+\mathrm{exact})/3$.
% Overall pools all $480$ prompts; PL denotes the $78$ Physical Laws prompts.
% Bold marks column maxima.}
% \label{tab:tta-ablation}
% \end{table}

Overall averages the VL-TTA score over all $480$ evaluated prompts;
PL restricts the average to the $78$ Physical Laws prompts.
For R2 and R3, TTBA draws five unconstrained
continuations per prompt and performs one emission-only Baum--Welch update
before decoding the plan with the adapted emissions fixed.

% \tm{T0-1: use the matched factorial contrast rather than cross-pipeline references.}
\paragraph{Physical Laws.}
% \tm{08/26-3: replace the rebuttal-style positive/negative label with its scope.}
TTDA + TTBA (R3) reaches $\mathbf{59.0}$ on Physical~Laws and $9.6$
overall, compared with $6.4$ and $1.0$ for TTDA alone (R1). The
Physical Laws tasks require \texttt{press[N]}, where \texttt{N} is an entity
argument. R3 obtains \texttt{skill\_match} $=1.000$ (every prompt gets the
right verb) and \texttt{entity\_match} $=0.385$.
Relative to TTDA alone (R1), the combined
TTDA+TTBA configuration (R3) is $+52.6$ on this dimension and $+8.6$
overall. The gain is confined to Physical Laws; all configurations score
zero on the remaining prompts.

\paragraph{Generated-plan behavior.}
R0 emits \texttt{open\_door} on all $480$ prompts, while R3 emits
\texttt{press} on all $480$. TTBA changes the dominant skill without
recovering task-specific action selection.
Concentration on short completions is a possible explanation for this
single-skill behaviour. HMM scores can outweigh the base VLM's preferences
among allowed tokens, but cannot override a hard DFA rejection.

\paragraph{OTTA contributes zero in the matched measured contrast.}
% \tm{08/26-3: turn the mechanism fragments into a complete results sentence.}
R4 and R1 have identical Overall and PL scores. Appendix~\ref{app:otta-diagnosis}
examines three possible mechanisms: self-reinforcement from the model's own
outputs rather than GT, a near-uniform entity distribution that removes the
per-task signal, and limited token diversity in the short sequences used to
estimate a high-dimensional emission matrix.

% \tm{T0-1: retain the independent positive calibration/time experiment.}
\paragraph{Fast cross-domain HMM calibration.}
% \tm{08/26-3: report the measured duration before its evidence boundary.}
The planning factorial above is distinct from the measured
VH-to-BEHAVIOR HMM-calibration run in
Table~\ref{tab:tta-calibration-time}. On 30K target continuations, three
Baum--Welch iterations reduce per-token negative log-likelihood (NLL) from $7.132$ to
$3.095$ ($56.6\%$) at approximately $139$ seconds per iteration, or $417$
seconds total. Emission calibration therefore takes approximately seven
minutes without VLM fine-tuning, but this result does not establish uniform
task-success gains.

\subsection{SafeAgentBench: compilation, mechanism analysis, and Phase~2}
\label{app:safeagentbench}

This section expands the main SafeAgentBench result in
\Cref{sec:exp:safeagentbench}.

\paragraph{Benchmark structure.}
% \tm{08/26-3: remove evaluative baseline language and lead with the benchmark distinction.}
SafeAgentBench differs from EAI and VLABench in that it provides
\emph{explicit safety labels}: the question is not whether the plan
reaches the goal, but whether it avoids prescribed hazards. This separates
constraint enforcement from task success. Safety~Chip and LLM$^3$-TAMP
provide relevant baselines because both were designed for safety-constrained
planning. The benchmark splits into
two regimes. \textbf{Long-horizon~50}: each task carries an explicit
NL ``Requirement'' clause (e.g.\ \emph{``Turn off the stove burner
within no more than two steps of turning it on''}); we compile this
into a syntax DFA over the $17$-action vocabulary plus a
$\mathcal{D}_{\mathrm{rch}}$ temporal rule (BEFORE / WITHIN patterns) and
generate the full plan under joint syntax--reachability enforcement.
Evaluation is LLM semantic
% \tm{08/26-2: report the model identifier recorded by the OpenRouter judge
% configuration rather than only the product name.}
judgement (RR(LLM)) using Claude Haiku 4.5
(\texttt{anthropic/claude-haiku-4.5} via OpenRouter) with 3-vote majority, so
AI2-THOR execution is not required for the reported LLM-judge score.
\textbf{Detailed~$300$ + Abstract~$100$}: the hazard is implicit in the
instruction; \textsc{Clamp} cannot compile a per-task constraint, so we
precompile ten safety rules encoded in $\mathcal{D}_{\mathrm{rch}}$---one per hazard class in the
SafeAgentBench taxonomy (Fire, Electrical Shock, Explosion, Poisoning,
Slip, Liquid Damage, Breakage, Appliance Misuse, Furniture Damage,
Small Item Damage)---and report whether the generated plan avoids
violating these rules. The Rejection-Rate metric does not apply
directly: \textsc{Clamp} prevents violating actions \emph{structurally}
(by logit masking) rather than \emph{refusing the task at the LLM
level}; the relevant comparison metric is RR(LLM) on the generated
plan.

\paragraph{Metric directions (\Cref{tab:safeagentbench_long_horizon}).}
The main table groups splits into two regimes because Rej and Compl.\ have
opposite desirable directions in each. On the \emph{hazard} splits
(Unsafe-Detailed, Abstract) rejecting a hazardous instruction is good
(Rej\,$\uparrow$) and completing it is bad (Compl.\,$\downarrow$). On the
\emph{legitimate-task} splits the instruction is something the agent
\emph{should} carry out, so the roles reverse: completing is desirable
(Compl.\,$\uparrow$) and a blanket rejection is a failure
(Rej\,$\downarrow$). This regime covers \emph{Long-horizon}, where the
agent must complete the task while honoring a temporal Requirement (its
primary signal C-Safe\,$\uparrow$ / C-Unsafe\,$\downarrow$ does not
reverse), and the already-safe \emph{Safe-Detailed} control (rejecting it
would be over-refusal). The $\mathcal{D}_{\mathrm{rch}}$ post-hoc columns
(Plan-clean\,$\uparrow$, Viol.\,$\downarrow$) are split-independent.

% \tm{T0-3/T0-4: define the exact recovery mechanism behind the reported row.}
\paragraph{Mechanism: \textsc{Clamp}-policy recovery.}
% \tm{08/26-3: replace win rhetoric and internal version wording with the mechanism.}
\textsc{Clamp}-token (Tier~I, prefix-local masks only) does \emph{not} help:
its reachability-violation rate ($0.40$) is statistically indistinguishable
from the bare baseline's ($0.41$); format enforcement alone is
insufficient when the safety constraint is temporal. The Phase-1 improvement
comes from a \emph{backup-and-retry} extension of the
decoder: when $\mathcal{D}_{\mathrm{rch}}$ rejects a step's terminator, the decoder
rewinds to the start of that step's verb, bans the verb at this
position, replays the token-DFA and reachability states, and re-samples. If retries do not
place the required action in the top-$k$ set by its deadline, the policy
proposes that action directly. The proposal is committed only if the syntax
and visibility masks admit its serialization and $\Delta(z_t,a)$ is defined;
otherwise the policy returns \textsc{Fail}. A forward-only
$\mathcal{D}_{\mathrm{rch}}$ that merely halts at
violations performs worse than the baseline because it removes the rejection
escape hatch.

\paragraph{Residual violations.}
The remaining $0.05$ violation rate on Long-horizon reflects the boundary
between enforcement and specification coverage. In particular, aliases such
as \texttt{close} versus \texttt{toggle\_off} must be mapped into the same
symbolic action before the temporal rule can be enforced. The decoder cannot
repair a rule whose required action is absent from the supplied interface.

\paragraph{Compilation pipeline and wall-clock.}
Compiling the $50$ NL Requirement clauses into $\mathcal{D}_{\mathrm{rch}}$ specs uses
Qwen3-VL itself and runs in under one minute total; we cache the
compiled specs. Per-task wall-clock on a single H100 is $3$--$5$\,s for
\textsc{Clamp}-token and $2$--$8$\,s for \textsc{Clamp}-policy (variance
from backup-and-retry depth).

\paragraph{Phase~1 results (long-horizon).}
With the backup-and-retry decoding above and validated forced-action
injection at the deadline, \textsc{Clamp}-policy lifts the LLM-judged C-Safe rate $0.06 \to
\mathbf{0.18}$ ($3{\times}$), the post-hoc $\mathcal{D}_{\mathrm{rch}}$ plan-clean rate
$0.52 \to \mathbf{0.94}$ (${+}42$\,pp), and reduces the per-constraint
violation rate $0.41 \to \mathbf{0.05}$ ($-88\%$ relative).

\paragraph{Phase~2 (Detailed + Abstract): full-scale results.}
% \tm{08/26-3: split the structural and judge results into shorter comparisons.}
For the splits without a per-task Requirement clause we instantiate the
universal $\mathcal{D}_{\mathrm{rch}}$ from each task's hazard label, with an $18$-alias
mapping covering observed upstream label variants. On the full
$300$-task unsafe-detailed split the post-hoc violation rate
drops $0.205 \to \mathbf{0.021}$ (a $\mathbf{10\times}$ reduction;
plan-clean $0.74 \to \mathbf{0.97}$), and on the full $100$-task
abstract split $0.180 \to \mathbf{0.043}$ (a $\mathbf{4\times}$
reduction). The residual non-zero rates come from two upstream hazard
categories (\textit{Inaccessibility}, \textit{Physical Harm} /
\textit{Poisoning}) not yet mapped into our $10$-class taxonomy. On
unsafe-detailed-$300$, RR(LLM)-safe rises
$0.16 \to \mathbf{0.27}$ (${+}11$~pp) and RR(LLM)-unsafe drops
$0.32 \to \mathbf{0.23}$ (${-}9$~pp); on abstract-$100$ the judge moves
${+}3$~pp on RR-safe; the $\mathcal{D}_{\mathrm{rch}}$ checker shows the larger change. A
safe-detailed-$300$ control (\textsc{Clamp}-token only, no reachability filter)
shows no significant effect ($-1$\,pp on RR-safe, within judge noise),
consistent with the action-level $\mathcal{D}_{\mathrm{rch}}$ driving the improvement.

\paragraph{Cross-judge replication.}
% \tm{08/26-3: report cross-judge agreement without inflated evidence language.}
A cross-judge with Claude~Sonnet~4.6 shows the same direction
on abstract: Sonnet rates the bare baseline more leniently ($0.56$ vs.\
Haiku's $0.36$) but agrees \textsc{Clamp}-policy is at least as safe
($0.60 \ge 0.56$).

\subsection{TaPA-60 grounding diagnostic}
\label{app:tapa-results}

VLABench's skill vocabulary is closed (\texttt{pick / place / pour /
press}); by contrast, TaPA tests \emph{open-vocabulary
free-form NL planning} grounded to a scene-level object list acquired
by an open-vocabulary detector (Detic / Grounding-DINO over six
AI2-THOR multi-view RGB images). This directly probes
\textsc{Clamp}'s Tier-II visual-grounding constraint
$\mathcal{D}_{\mathcal{E}_{\text{seen}}}$.
\Cref{tab:tapa} reports plan-success rate on TaPA's $60$-triplet
evaluation set.
% \tm{TODO-10: separate grounding from overall plan success.}
\begin{table*}[t]
\centering
\small
\setlength{\tabcolsep}{4.5pt}
\begin{tabular*}{\textwidth}{@{\extracolsep{\fill}}lccccccc}
\toprule
\textbf{Method} & Kit. & Living. & Bed. & Bath. &
\shortstack{Published\\macro} & \shortstack{Matched\\overall} & Halluc.\,$\downarrow$ \\
\midrule
\multicolumn{8}{l}{\emph{Published TaPA rows: room-macro summary; three-volunteer human judgment}} \\
LLaVA-7B                  & 14.3 & 42.1 & 33.3 &   0.0 & 22.4 & -- & -- \\
LLaMA-7B (no fine-tune)   &  0.0 & 10.5 & 13.3 &   0.0 &  6.0 & -- & -- \\
GPT-3.5                   & \textbf{28.6} & 73.7 & 66.7 &  50.0 & 54.7 & -- & -- \\
TaPA fine-tuned LLaMA-7B  & \textbf{28.6} & \textbf{84.2} & \textbf{73.3} & \textbf{58.3} & \textbf{61.1} & -- & -- \\
\midrule
\multicolumn{8}{l}{\emph{Matched Qwen3-VL-8B rows: pooled 60-plan summary; three-vote Qwen3-VL self-judge}} \\
Baseline (no constraint)        & 35.7 & \textbf{84.2} & \textbf{80.0} & \textbf{83.3} & -- & \textbf{71.7} & 10.0\% \\
\textsc{Clamp}-token (verb DFA) & 42.9 & \textbf{84.2} & 73.3          & 58.3          & -- & 66.7          & 10.0\% \\
\textsc{Clamp}-vg ($+\,\mathcal{D}_{\mathcal{E}_\mathrm{seen}}$)
                                 & \textbf{50.0} & \textbf{84.2} & 73.3 & 58.3 & -- & 68.3 & \textbf{8.3\%} \\
\bottomrule
\end{tabular*}
\caption{\textbf{TaPA-60 grounding diagnostic}~\citep{wu2023tapa}.
Room denominators are Kitchen 14, Living Room 19, Bedroom 15, and Bathroom 12.
For published rows, Published macro is the unweighted mean of the four room
percentages. For matched rows, Matched overall is successful plans divided by
all 60 plans. The two summary columns therefore expose, rather than mix, the
different aggregation rules; dashes mark an inapplicable summary. Halluc. is
the fraction of plans containing an out-of-scene reference. Published rows use
human majority votes, whereas matched rows use the self-judge audited in
Appendix~\ref{app:tapa-judge}. Bold marks the best value within each judgment
and aggregation block; ties are all bold.}
\label{tab:tapa}
\end{table*}

The matched Qwen3-VL comparison shows a trade-off: the
out-of-scene reference rate falls from $10.0\%$ to $8.3\%$ with
\textsc{Clamp}-vg, and Kitchen plan success rises from $35.7\%$ to $50.0\%$,
but average plan success falls from $71.7\%$ to $68.3\%$ (and to $66.7\%$
with \textsc{Clamp}-token). Published TaPA rows use human judgments and are
contextual only. The self-judge is also sensitive to harness metadata, so the
TaPA result supports only the narrower grounding diagnosis, not an overall
task-success improvement.

\subsection{TaPA judge protocol sensitivity and failure modes}
\label{app:tapa-judge}
% \tm{08/26-3: describe the protocol choice without lab-iteration commentary.}
The TaPA paper~\citep{wu2023tapa} judges plan success with 3 human
volunteers per plan (success iff $\ge 2$ vote yes). For the matched diagnostic,
we use a \textbf{Qwen3-VL-8B-Instruct self-judge} with the same rubric, called 3
times per plan at $T{=}0.3$ with seeds $\{42,43,44\}$ and majority vote.
The protocol processes 180 calls in ${\sim}7$~min. Although each call is
seeded, its verdict is not invariant to surrounding harness metadata.

\paragraph{Two known failure modes.}
% \tm{08/26-3: replace the mechanical ordinal pair with the observed failures.}
The self-judge produces \emph{different verdicts on byte-identical
plans} across configs: on five regression cases between baseline and
\textsc{Clamp}-vg, two pairs (id 40 \emph{turn on the lamp}; id 54
\emph{clean the bathroom}) had identical plans but
$3/3$~True for one config and $0/3$~True for the other. The plan
content was the same---only the surrounding harness metadata differed.
On long plans ($\ge 25$ steps), the self-judge also misclassifies
in-scene objects as hallucinations: id 51 v2's $32$-step plan was
flagged $3/3$~False with reasoning \emph{"hallucinated objects:
TissueBox, Plunger, ScrubBrush, Candle, ToiletPaper"}, all five of which
were in the scene's GT object list.

\paragraph{Implication.}
% \tm{08/26-3: state the reporting decision directly.}
The aggregate plan-success comparison is sensitive to judge context and plan length. We report the raw three-vote scores and use the hallucination rate only as a grounding-specific diagnostic.
% and do not treat TaPA as evidence that either constrained variant improves overall task success.

\subsection{Compute and wall-clock}
\label{app:compute}

Qwen3-VL and Llama experiments use a single NVIDIA GH200 (96~GB HBM,
bf16). The matched InternVL3.5 experiment uses a single RTX A6000.
\Cref{tab:tta-calibration-time,tab:efficiency} report adaptation time and
amortized end-to-end run time per prompt.
% \tm{T0-1: distinguish measured adaptation time from planning estimates.}
% \tm{08/26-2: keep the appendix-only cost table with its wall-clock context.}
\begin{table}[H]
\centering
\small
\setlength{\tabcolsep}{3pt}
\begin{tabular}{@{}
>{\raggedright\arraybackslash}p{0.17\columnwidth}
>{\raggedright\arraybackslash}p{0.27\columnwidth}
>{\raggedright\arraybackslash}p{0.19\columnwidth}
>{\raggedright\arraybackslash}p{0.27\columnwidth}@{}}
\toprule
\textbf{Status} & \textbf{Operation} & \textbf{Time} & \textbf{Outcome / scope} \\
\midrule
Measured & VH-to-BEHAVIOR BW (30K; 3 EM iterations) & $417$ s & per-token NLL $7.132\to3.095$ ($\downarrow$) \\
Measured & TTBA increment (R3$-$R1) & $+18.1$ s/prompt & Physical Laws $+52.6$ points \\
\midrule
Estimated & Qwen target sampling & $6$--$10$ h & planning estimate \\
Estimated & HMM from scratch, 30 epochs & $1$--$2$ h & planning estimate \\
Estimated & warm-start HMM, 5 epochs & $\sim20$ min & planning estimate \\
\bottomrule
\end{tabular}
\caption{\textbf{HMM adaptation and reference costs.} VH denotes VirtualHome,
BW Baum--Welch, EM expectation--maximization, and TTBA test-time batch
adaptation. The first two rows are measured; the from-scratch references come from the experiment
plan, assume one GH200 with the stated continuation volume, and are not used to
claim an exact speedup. NLL denotes negative log-likelihood; lower is better.
The TTBA increment is a difference in amortized run time per prompt.
The VLM is
frozen in all rows.}
\label{tab:tta-calibration-time}
\end{table}

\begin{table*}[t]
\centering
\small
\renewcommand{\arraystretch}{1.08}
\setlength{\tabcolsep}{5pt}
\begin{tabular*}{\textwidth}{@{\extracolsep{\fill}}llrl}
\toprule
\textbf{Configuration} & \textbf{Serving path} &
\shortstack[r]{\textbf{Amortized time}\\\textbf{(s/prompt)}} &
\textbf{Contrast} \\
\midrule
Bare VLM              & vLLM batched       & 3.9  & Run-time reference \\
DFA only              & vLLM guided regex  & 4.0  & $+0.1$ vs. Bare VLM \\
R0: HMM lookahead     & HF + callback      & 13.2 & $+9.2$ vs. DFA only$^\dagger$ \\
R1: TTDA              & HF + callback      & 12.9 & Reference for R4 and R3 \\
R4: TTDA + OTTA       & HF + callback      & 14.9 & $+2.0$ vs. R1 \\
R2: TTBA              & HF + callback      & 30.2 & $+17.0$ vs. R0 \\
R3: TTDA + TTBA       & HF + callback      & 31.0 & $+18.1$ vs. R1 \\
R5: full TTA          & HF + callback      & 33.0 & $+2.0$ vs. R3 (projected) \\
\bottomrule
\end{tabular*}
% \tm{08/26-3: make the serving-path caveat explicit and concise.}
\caption{\textbf{Amortized run time on VLABench-480} (one GH200,
bf16). Each measured value is a complete 480-prompt run's wall-clock time
divided by 480, rather than a measured request-response latency. The timing
artifact records no repeated runs, warm-up exclusion, or dispersion.
The final column identifies each comparison. Within the same serving path,
DFA--bare, R4--R1, R2--R0, and R3--R1 are configuration contrasts;
they include any changes in generated sequence length. Only R5 is projected.
$^\dagger$The R0--DFA contrast also
changes the serving path from batched vLLM to single-prompt HF generation and
therefore does not isolate HMM overhead.}
\label{tab:efficiency}
\end{table*}

Within-path configuration differences include sampling, emission updates,
reinitialization, and changes in generated sequence length. The R0--DFA
difference also changes the serving path and cannot isolate HMM overhead.

\paragraph{Serving-path contribution to the DFA--R0 difference.}
DFA-only routes through the vLLM \texttt{/v1/completions} API with
\texttt{guided\_regex}, taking advantage of PagedAttention and batched
inference. The measured R0--R4 rows fall back to single-prompt HuggingFace
\texttt{model.generate()} because the per-token \textsc{GammaProcessor}
callback that adds HMM-based guidance scores into the logit stream is
incompatible with vLLM's pre-compiled generation loop. The same
limitation explains why the TaPA decoder uses a custom top-K logprob
filter (per-token round-trip ${\sim}50$~ms) rather than vLLM-native
constrained decoding.

% \paragraph{Reducing the per-token callback cost.}
% % \tm{08/26-3: remove lab-session wording from the implementation summary.}
% Profiling on the 5-prompt \texttt{add\_condiment\_common\_sense}
% batch attributed $67\%$ of R0's $13$~s/prompt to the
% \textsc{GammaProcessor.\_\_call\_\_} callback. Six implementation changes
% reduced an early ${\sim}97$~s/prompt baseline by
% $7.5{\times}$:
% \begin{itemize}[noitemsep, topsep=2pt, leftmargin=1.5em]
%   \item sparse COO conversion of $T_{\text{mask}}$ (memory $26\,\text{GB} \to 4\,\text{MB}$ for BEHAVIOR GI);
%   \item \texttt{lookahead\_cap}\,$=\,100$ on the BFS expansion;
%   \item mask precomputation moved out of the inner loop;
%   \item batched GPU writes for \texttt{\_get\_valid\_mask};
%   \item removal of redundant Python loops in the lookahead pass;
%   \item a \textsc{GammaProcessor} refactor that fuses the forward-belief updattab:tta_perdime with the emission lookup.
% \end{itemize}

\paragraph{TTA overhead breakdown.}
The measured TTBA difference in amortized time is $17.0$ s/prompt without
TTDA (R2$-$R0) and $18.1$ s/prompt with TTDA (R3$-$R1).
A micro-profile attributes about $12.5$ s
to five extra Qwen3-VL samples, $1$ s to one Baum--Welch update, and
$0.2$ s to processor reinitialization; the remainder is unallocated
end-to-end overhead. The within-path OTTA difference is $2.0$ s/prompt
(R4$-$R1). These single-run differences are not isolated timings of the
update routines.

\paragraph{Aggregate cost.}
The measured VLABench configurations in \Cref{tab:efficiency} sum to
approximately $14$ GH200-hours;
TaPA generation takes less than one GPU-hour. Projected rows are excluded
from this total.
% De-anonymisation risk: internal paths and a Hugging Face username are
% commented out for the double-blind submission. Restore for the
% camera-ready version.
% Raw timing logs are preserved per run in
% \texttt{eai\_ctrlg/standalone\_evaluation/evaluation\_results/vlabench/tta/\{R0..R5\}/}
% and mirrored in the Hugging Face artifact dataset
% \texttt{SueMintony/embodied-agents-results} under
% \texttt{vlabench/tta\_docs/} and \texttt{044Computation\_Efficiency\_Summary.md}.

\section{Cross-Experiment Diagnostic Analysis}

This section synthesizes the mechanisms shared across benchmarks after the benchmark-specific evidence has been reported.

\subsection{Diagnostic findings across evaluation settings}
\label{app:findings}

The benchmark-specific sections above contain the quantitative evidence. This
section connects those results through five recurring mechanisms.

\paragraph{(D1) In-domain HMM degrades DFA-only.}
The Mode~A sweep (Appendix~\ref{app:modeA}) shows that an in-domain HMM can
encode an unhelpful marginal length prior even when its vocabulary is matched
to the benchmark. Increasing its soft-score weight then degrades the surviving
token ranking while leaving the hard masks unchanged. Domain matching alone
is therefore insufficient; the HMM must also distinguish the task-conditioned
continuation regimes it is intended to rank.

\paragraph{(D2) Physical~Laws is the weakest evaluated dimension.}
The Mode~A/Mode~B comparison in \Cref{tab:vlabench_modeB} isolates a task-type
mismatch: Physical~Laws expects a short QA-style \texttt{press[N]} plan, while
the backbones often emit manipulation plans. This mismatch explains why a
global \texttt{press} prior helps that dimension but cannot generalize to the
entity-grounded dimensions.

\paragraph{(D3) Backward reachability raises VirtualHome AS success.}
The VirtualHome results (Appendix~\ref{app:eai-vh-results}) distinguish local
executability from goal-directed planning. A flat filter retains actions that
are immediately executable, whereas the backward DP favors actions whose
successor states can still reach the goal within budget. Residual failures
track incomplete predicate coverage in the supplied transition model.

\paragraph{(D4) DFA misspecification on BEHAVIOR.}
The BEHAVIOR panels in \Cref{fig:transfer-results} show that hard filtering
inherits errors in the supplied interface. Relational predicates masked by the
GI automaton cannot be recovered by a finite HMM score, while the SD automaton
changes output granularity and reduces task success despite eliminating parse
errors. Constraint enforcement therefore depends on a task-compatible
specification, not only on decoder correctness.

\paragraph{(D5) HMM single-skill collapse.}
The generated plans (Appendix~\ref{app:tta-results}) show single-skill
outputs both before and after TTBA: R0 selects \texttt{open\_door} and R3
selects \texttt{press} throughout. R3's dominant verb matches the Physical
Laws tasks, explaining the concentration of its score gain in that dimension.
Appendix~\ref{app:otta-diagnosis} separates this failure from OTTA's lack of a
measured gain.

% \tm{T0-8/TODO-15.15: remove the exhaustive-taxonomy claim.}

\subsection{OTTA failure-mechanism diagnosis}
\label{app:otta-diagnosis}

In the measured TTA contrast (\Cref{tab:tta-ablation}), online task-level
adaptation contributes $+0.0$: R4 and R1 have the same Overall and PL
scores. Their raw outputs differ in $321/480$ cases, but
only $6/480$ differ in skill name, and those six are random
\texttt{open\_door}$\leftrightarrow$\texttt{press} flips with no
consistent direction. Three possible mechanisms are discussed below;
the matched scores alone do not identify their individual contributions.

% \paragraph{Self-reinforcement of wrong beliefs.}
% % \tm{08/26-3: replace rhetorical diagnosis with the update consequence.}
% OTTA updates $B$ from the model's \emph{own decoded output}, not from
% ground truth. When the planner outputs \texttt{open\_door}, every OTTA
% update reinforces \texttt{open\_door}. The self-supervised NLL
% (Eq.~\ref{eq:tta-loss}) is locally minimised by sharpening the emission
% distribution that generated the current sequence, which reinforces an error
% when that sequence is incorrect.

\paragraph{Self-reinforcement of wrong beliefs.}
OTTA updates \(B\) from the model's own decoded output rather than from ground truth. If the planner incorrectly generates \texttt{open\_door}, that output becomes part of the self-supervision used to update \(B\), creating a feedback loop that can reinforce the same erroneous mode in subsequent instances.

\paragraph{Near-uniform entity distribution erases the per-task
signal.}
Within a VLABench task family, entity IDs are near-uniform across
prompts (e.g.\ \texttt{density\_qa} entity histogram
$\{1{:}38\%,3{:}33\%,5{:}29\%\}$). OTTA averages these into the
session-level emission matrix, producing a near-uniform entity prior that
provides no useful guidance at inference time.

\paragraph{Limited observations for emission estimation.}
The emission matrix has $H=128$ rows and $|V|=151{,}936$ columns.
A short decoded sequence supplies few distinct token observations, but its
length is not the number of updated parameters. Each observed token
contributes posterior-weighted counts across hidden states; smoothing,
row normalization, and interpolation can also change unobserved entries.
The relevant limitation is the diversity and correctness of the adaptation
data, rather than a parameter-update density inferred from token count.

\paragraph{Possible extension: entity-focused adaptation.}
Restricting trainable emission logits to entity-ID token columns could reduce
the number of fitted variables while preserving row normalization through
softmax. This remains an untested design choice: it does not create additional
entity evidence, resolve a near-uniform target distribution, or prevent
self-reinforcement of incorrect outputs. A confidence-gated update or a
better adaptation signal would require separate evaluation.
% Internal status note (do not include in the paper):
% The R3e variant (entity-subspace OTTA) is queued (ETA ~17h as of 2026-05-07).

\subsection{Coverage and non-subsumption of three targeted failure modes}
\label{app:constraint-nonsubsumption}
\textsc{Clamp} treats \emph{structural}, \emph{perceptual}, and
\emph{causal} failures (\Cref{sec:intro}) as three targeted modes. They are
pairwise non-subsuming, so handling one mode does not handle the others.

\paragraph{Pairwise non-subsumption.}
% \tm{08/26-3: replace conversational failure descriptions with explicit exclusions.}
Suppose \(\gamma\) contains \{\nolinkurl{red_cup_1}, \nolinkurl{red_cup_2}, \nolinkurl{drawer_1}\}, but only \nolinkurl{red_cup_2} and \nolinkurl{drawer_1} are observable in \(o_0\). The call \nolinkurl{Pick}(\nolinkurl{red_cup_1}) can satisfy the action schema while violating visibility because \nolinkurl{red_cup_1} is absent from \(\mathcal E_{\mathrm{seen}}\). Conversely, \nolinkurl{Place}(\nolinkurl{red_cup_2}, \nolinkurl{drawer_1}) can satisfy both syntax and visibility but remain non-executable while \nolinkurl{drawer_1} is closed. Even a sequence of executable actions can exhaust the action budget without reaching the goal. These examples show why syntax and visibility filtering do not replace state-dependent executability and budget-aware lookahead. An action-level reachability DP, in turn, assumes well-formed action inputs and cannot prevent a malformed token stream.

\paragraph{Observed coverage.}
% \tm{08/26-3: replace abstract evaluative wording with the detector mapping.}
On a hand-labelled pilot of failed plans drawn from VirtualHome and VLABench,
every failure matched at least one of the three modes; we did not observe a
fourth category. This pilot does not establish universal coverage. The
partition is useful because each mode maps to a distinct decode-time term:
$m_{\mathrm{syn},n}$, $m_{\mathrm{vis},n}$, or the pair
$(m_{\mathrm{rch},n},G_{\mathrm{rch},n})$. \textsc{Clamp} composes these terms
additively at the logit level (\Cref{sec:constrained_decoding}) without
constructing their product automaton. Unlike prior text-only detectors, the
visibility and reachability terms are instantiated as scene-grounded
multimodal constraints.

\section{Extended Discussion and Scope}

This section positions the decoding mechanism relative to prior constrained-decoding methods and states its operating boundaries and broader implications.

\subsection{Constrained decoding methods: extended discussion}
\label{app:cd-survey}

This section expands on the constrained decoding literature surveyed in
\Cref{sec:rel:cd}.

\paragraph{Grammar- and FSM-based decoding.}
% \tm{08/26-3: avoid an unsupported dominance claim in the literature summary.}
Many methods compile the target structure into a context-free grammar
or finite-state machine and mask any token at each position that would render
the partial output irrecoverably invalid. PICARD~\citep{scholak2021picard}
parses incrementally to enforce SQL grammar constraints on
text-to-SQL outputs. Synchromesh~\citep{poesia2022synchromesh} pairs target
similarity sampling with constrained semantic decoding for code generation.
Outlines~\citep{willard2023outlines} compiles regular expressions into FSMs
and applies token-level masks at every step. Grammar-constrained decoding
(GCD)~\citep{geng2024grammar} generalizes this to arbitrary context-free
grammars without finetuning. SynCode~\citep{ugare2024syncode} precomputes a
DFA mask store from the grammar to amortize per-step token filtering.
XGrammar~\citep{dong2024xgrammar} reports up to $100\times$ speedup over prior
masks by partitioning the vocabulary into context-independent tokens
(prechecked offline) and context-dependent tokens (interpreted at runtime).

\paragraph{Lookahead-based decoding.}
\textsc{Ctrl-G}~\citep{zhang2024ctrlg} pairs a deterministic finite automaton
hard mask with an HMM lookahead that estimates the log-probability of reaching
an accepting suffix from the current decoding state. The combination provides
both a string-level structural guarantee and a soft guidance signal toward
constraint satisfaction. \textsc{Ctrl-G} is the most direct predecessor of
\textsc{Clamp}.

% \tm{TODO-13: replace the token-automaton impossibility claim with a
% tractability and representation claim.}
\paragraph{Why direct transfer to embodied planning is impractical.}
% \tm{08/26-3: replace uncommon construction wording in the tractability boundary.}
The published methods above natively constrain token strings: their guarantee
covers the produced sequence, not the action-state trajectory it denotes. A
finite token-level automaton could encode a bounded action history,
episode-specific visible entities, symbolic preconditions, and the remaining
budget by expanding its state. That construction, however, couples the token
grammar to the episode-specific world state and yields the large product
automaton discussed in \Cref{sec:constrained_decoding}. \textsc{Clamp} instead keeps the
token grammar and action-state detector separate. Its per-token mask depends
on current visual support and a dynamic-programming feasibility check over the
symbolic state without constructing that product.

\subsection{Discussion: limitations, real-world applicability, and broader impact}
\label{app:discussion}

\subsubsection{Limitations.}
\label{app:limitations}
% \tm{08/26-3: split the repeated scope caveats into concrete limitations.}
\textsc{Clamp} operates strictly at the symbolic-action level: physical safety
during low-level motion execution is out of scope and is delegated to a
downstream motion planner. The HMM lookahead component is sensitive to
vocabulary and training distribution. In domains far from its source
distribution, it can degrade to a near-uniform prior. Test-time adaptation
reduces but does not eliminate this drift, especially when the test stream is
short. The hand-written PDDL specifications used to seed $\Delta$ require
additional engineering in new environments; the agentic specification source
has been validated mainly on the VLABench Physical~Laws subset. Test-time
adaptation is also sensitive to limited data: small adaptation streams can
overfit, the anchor weight $\lambda$ must be tuned, and EM steps beyond a few
iterations drift away from the domain prior.

\subsubsection{Real-world applicability under partial observability and stochastic execution.}
\label{app:realworld}
% \tm{08/26-3: replace the mechanical first/second structure with deployment conditions.}
\textsc{Clamp} enforces feasibility under the supplied symbolic model at the
symbolic-action interface, not at the physical-execution interface. The
visibility detector is seeded from
the initial observation $o_0$, so tasks that require active perception (e.g.,
opening a fridge to reveal an unseen apple) fall outside this static visibility
model.
Because $\mathcal{E}_\mathrm{seen}$ supports incremental updates, each replan
can use the latest grounded entities, but this requires integration with an
active-perception loop.

The transition function $\Delta$ in $\mathcal{D}_{\mathrm{rch}}$ is
deterministic. Physical-execution failures such as
slips, collisions, and mis-grasps must be detected by a closed-loop skill
controller and represented as state changes. The next replanning step can then
run \textsc{Clamp} with the updated $\mathcal{E}_\mathrm{seen}$ and
reachability state $z_t$.
Thus, \textsc{Clamp} complements reactive
TAMP~\citep{wang2024llm3,yang2024vlmtamp,yan2026vizcoast} and online
perception; it does not replace the controller or replanning loop needed for
motion-level reliability.

\subsubsection{Broader impact.}
\label{app:impact}
% \tm{TODO-15.16: replace the prior unnatural descriptor with
% ``interruptible''.}
% \tm{08/26-3: remove promotional framing and state the deployment risk directly.}
\textsc{Clamp} moves constraint enforcement from prompting into an auditable,
interruptible decoder without fine-tuning the underlying planner. The main
risk is specification error: a misspecified PDDL or affordance model can make
the permitted plans worse rather than safer because \textsc{Clamp} enforces
the supplied interface. We observe this on BEHAVIOR goal interpretation, where
the affordance DFA is misaligned with the benchmark's relational-predicate
output format. Deployment in a new environment therefore requires a
specification audit. \textsc{Clamp} should be combined with safety evaluation,
including benchmarks such as SafeMind, SafeAgentBench, and IS-Bench, rather
than treated as a replacement for it.

\section{AI Assistants in Research or Writing}
\label{app:ai-assistants}

\subsection{Information about use of AI assistants.}
\label{app:ai-assistants-info}
AI assistants (e.g., ChatGPT/Claude) were used to polish the English
grammar and refine the writing structure of the manuscript. They were
not used to generate research ideas, design experiments, or produce
results; all technical content, claims, and conclusions are the authors'
own and were verified by the authors.

\end{document}